\documentclass[acmtog,nonacm]{acmart}

\usepackage{booktabs} %
\usepackage{diagbox}
\usepackage{multirow}
\usepackage{mathrsfs}
\usepackage{graphicx}
\usepackage{graphbox}
\usepackage{enumitem}
\usepackage{soul}
\usepackage{url}
\usepackage[normalem]{ulem}
\usepackage[skip=3pt]{caption}
\usepackage{subfigure}
\usepackage{rotating}
\usepackage{makecell}
\usepackage{tikz}

\setcitestyle{square} %

\usepackage[ruled]{algorithm2e} %

\SetAlFnt{\small}
\SetAlCapFnt{\small}
\SetAlCapNameFnt{\small}
\SetAlCapHSkip{0pt}

\acmJournal{TOG}

\newcommand{\figframe}[2][scale=0.5]{
        \setlength{\fboxrule}{0.25pt}
        \setlength{\fboxsep}{0pt}
        \hspace{-45\fboxrule}
        \framebox{\includegraphics[#1]{#2}}
}

\definecolor{MarziaColor}{RGB}{255,0,127}
\definecolor{GiuseppeColor}{RGB}{85, 150, 223}
\definecolor{FabioColor}{rgb}{119, 0, 200}

\begin{document}

\author{Marzia Riso}
\affiliation{%
	\institution{Sapienza University of Rome}
	\country{Italy}
}
\email{riso@di.uniroma1.it}

\author{Giuseppe Vecchio}
\affiliation{%
	\institution{Independent Researcher}
	\country{Italy}
}
\email{giuseppevecchio@hotmail.com}

\author{Fabio Pellacini}
\affiliation{%
	\institution{University of Modena and Reggio Emilia}
	\country{Italy}
}
\email{fabio.pellacini@unimore.it}

\newcommand{\methodName}{Paff\xspace}
\title{Structured Pattern Expansion with Diffusion Models}

\begin{abstract}
Recent advances in diffusion models have significantly improved the synthesis of materials, textures, and 3D shapes. By conditioning these models via text or images, users can guide the generation, reducing the time required to create digital assets. In this paper, we address the synthesis of structured, stationary patterns, where diffusion models are generally less reliable and, more importantly, less controllable.

Our approach leverages the generative capabilities of diffusion models specifically adapted for the pattern domain. It enables users to exercise direct control over the synthesis by expanding a partially hand-drawn pattern into a larger design while preserving the structure and details of the input. To enhance pattern quality, we fine-tune an image-pretrained diffusion model on structured patterns using Low-Rank Adaptation (LoRA), apply a noise rolling technique to ensure tileability, and utilize a patch-based approach to facilitate the generation of large-scale assets.

We demonstrate the effectiveness of our method through a comprehensive set of experiments, showing that it outperforms existing models in generating diverse, consistent patterns that respond directly to user input.
\end{abstract}

\ccsdesc[500]{Computing methodologies~Computer graphics}

\begin{teaserfigure}
    \begin{tabular}{ccc}
        \figframe[width=0.33\textwidth]{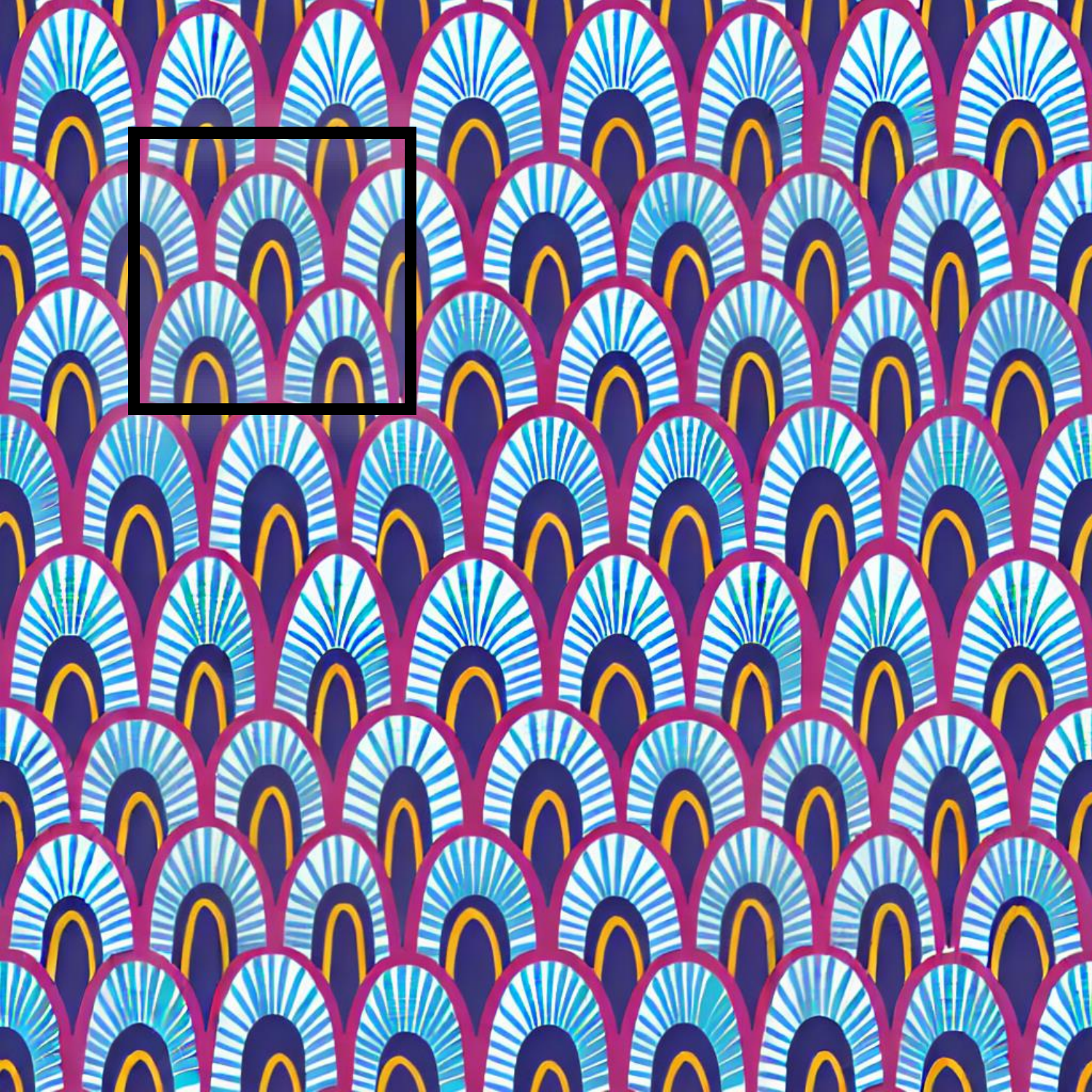}&
        \hspace{-3mm}
        \figframe[width=0.33\textwidth]{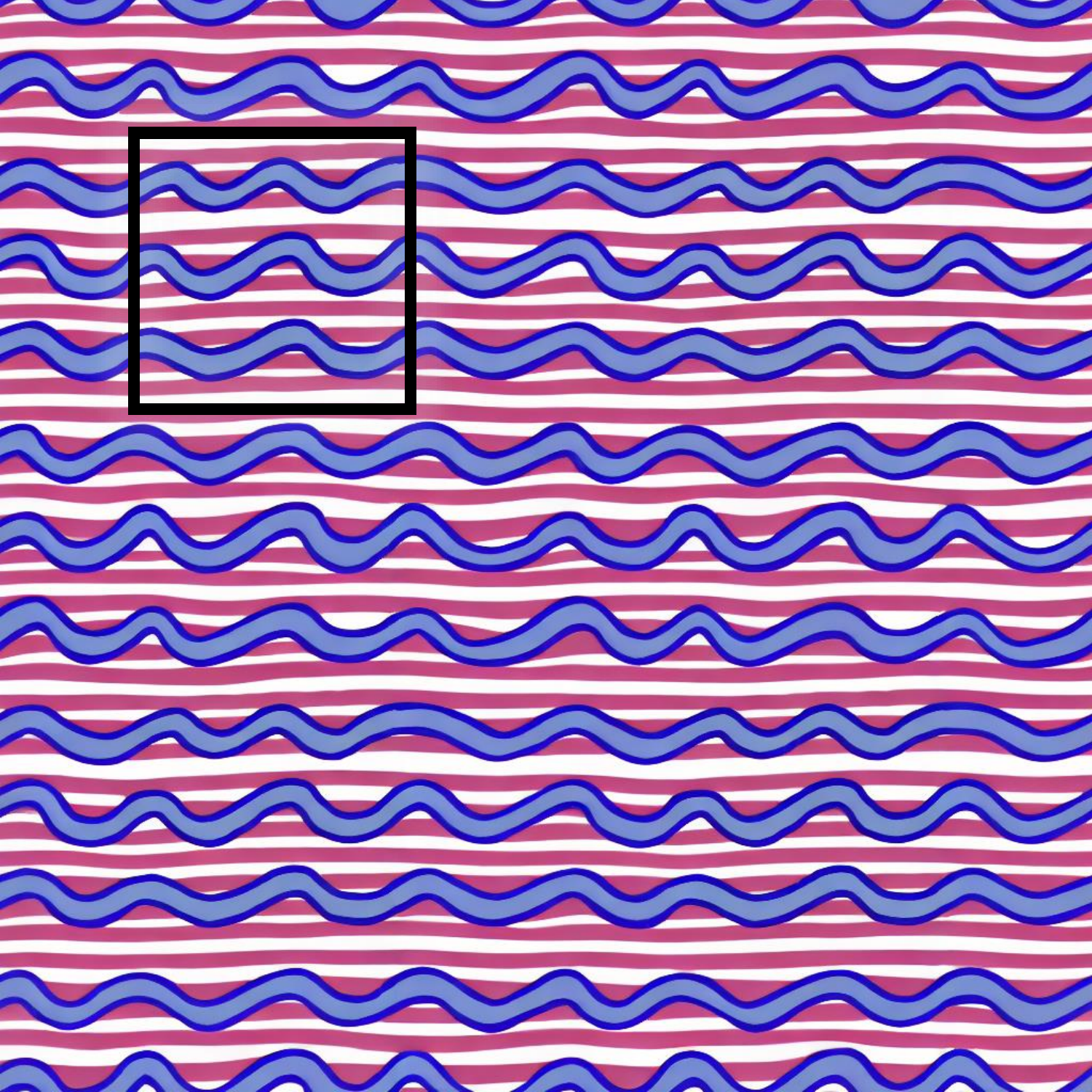}&
        \hspace{-3mm}
        \figframe[width=0.33\textwidth]{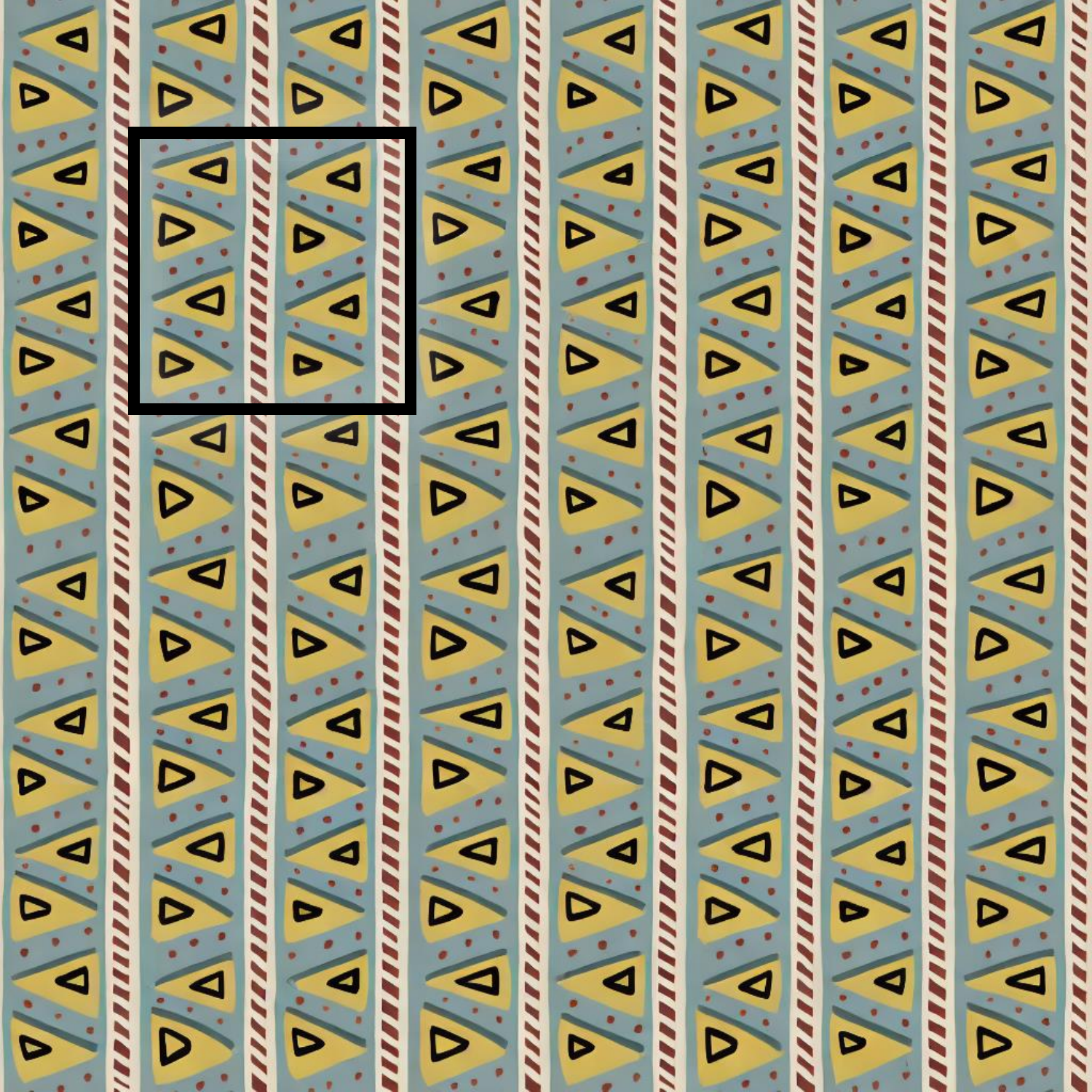} \\
    \end{tabular}
        \caption{\textbf{Overview.} We present a diffusion-based model for structured pattern expansion. Our approach enables the generation of large-scale, high-quality tileable patterns by extending a user-drawn input, shown within the black boxes, to an arbitrarily sized canvas. Our method extends the input pattern while faithfully following the user input and producing coherently structured and yet non-repetitive images.}
    \label{fig:teaser}
\end{teaserfigure}

\maketitle

\section{Introduction}
\label{sec:introduction}

Hand-drawn structured patterns are central to computer graphics, with applications spanning various domains in design and digital art. Creating these patterns remains a complex and time-consuming task that requires specialized expertise. AI-assisted content creation offers the potential to simplify this process. For instance, learning-based image synthesis methods have shown impressive generation capabilities for natural images \cite{rombach2022high,brock2018large,karras2017progressive,karras2020analyzing,podell2023sdxl}. However, the application of these methods to pattern-like synthesis has primarily focused on unstructured, realistic materials~\cite{zhou2018, zhou2023neural, zhou2024generating, guo2023text2mat, vecchio2023matfuse, vecchio2023controlmat, vecchio2023matsynth}, leaving the creation of structured patterns an underexplored task. In contrast, the synthesis of highly structured vector patterns has been assessed by automatically discovering and exploiting their structure, geometry and topology \cite{tu2020, reddy2020} or by optimizing the procedural parameters of differentiable vector pattern to match a sketch or a viewport edit \cite{riso2023, riso2022}.

\begin{figure}[t]
    \centering
    \setlength{\tabcolsep}{.5pt}
    \begin{tabular}{cccc}
        \multicolumn{2}{c}{\textbf{In-domain samples}} & \multicolumn{2}{c}{\textbf{Out-of-domain samples}}\\
        
        \vspace{-0.5mm}\hspace{-1mm}
        \figframe[width=0.115\textwidth, height=0.115\textwidth]{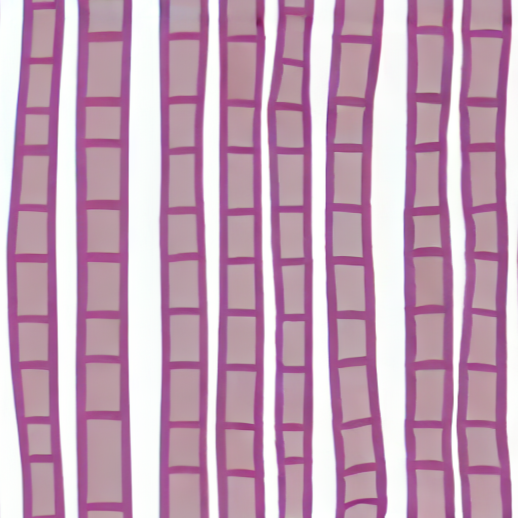} &
        \figframe[width=0.115\textwidth, height=0.115\textwidth]{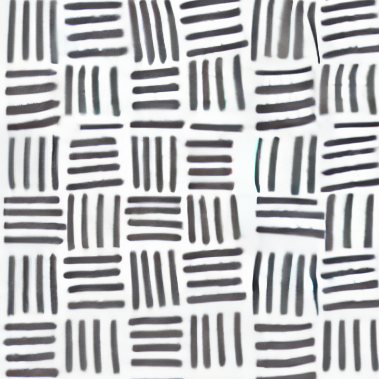} &
        \hspace{2pt}\figframe[width=0.115\textwidth, height=0.115\textwidth]{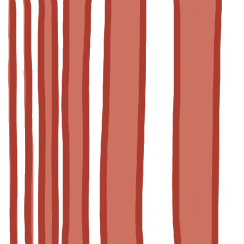} &
        \figframe[width=0.115\textwidth, height=0.115\textwidth]{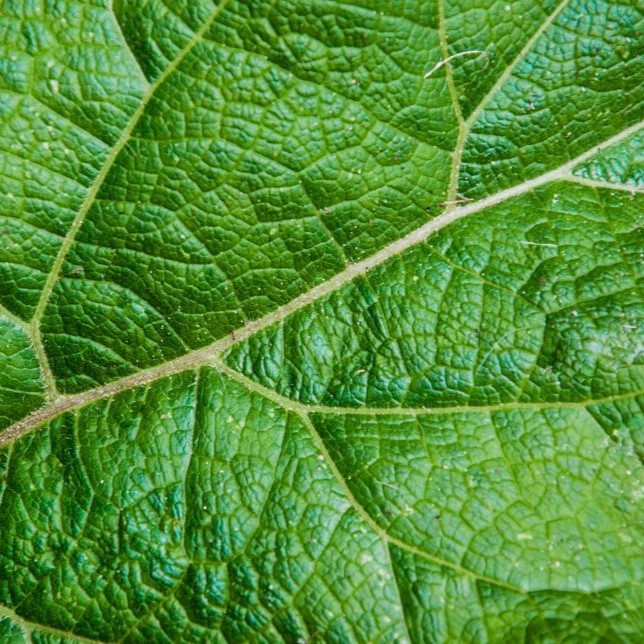} \\
        
        \vspace{-0.5mm}\hspace{-1mm}
        \figframe[width=0.115\textwidth, height=0.115\textwidth]{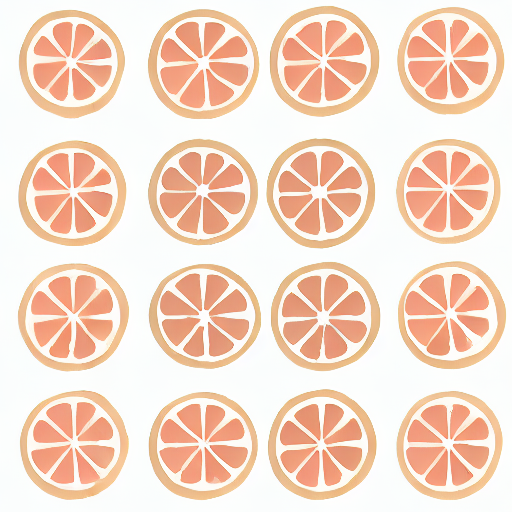} &
        \figframe[width=0.115\textwidth, height=0.115\textwidth]{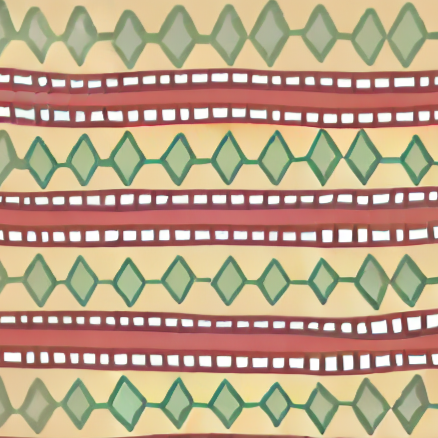} &
        \hspace{2pt}\figframe[width=0.115\textwidth, height=0.115\textwidth]{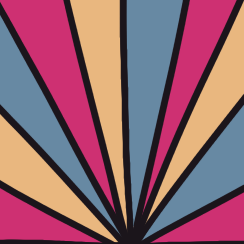} &
        \figframe[width=0.115\textwidth, height=0.115\textwidth]{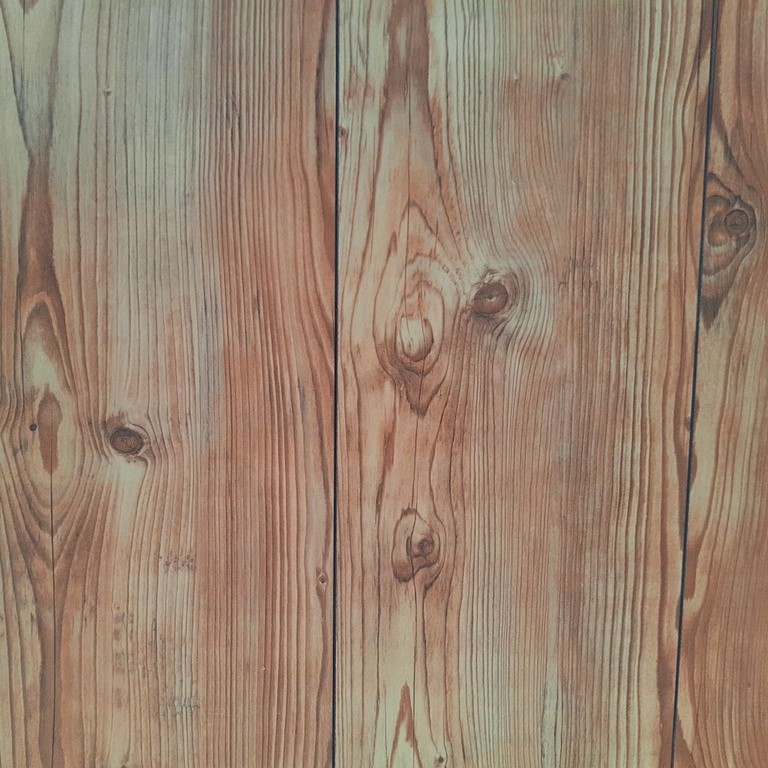} \\
    \end{tabular}
    \caption{\textbf{Pattern Category}. In this work, we focus on structured, stationary, patterns in a hand-drawn style, characterized by repeated recognizable shapes drawn in flat colors (left). Unstructured or aperiodic patterns, as well as photorealistic textures, fall outside the scope of this paper (right).}  
    \label{fig:pattern_cat}
\end{figure}

Our work focuses on structured patterns with a hand-drawn appearance, characterized by the repetitions of sketch-like shapes filled with solid colors and defined by sharp, crisp edges. Formally, these structured patterns consist of stationary repetitions of recognizable shapes, each with individual variations, and are drawn with piecewise-constant colors.
Examples of these patterns are shown throughout the paper, with Fig.~\ref{fig:pattern_cat} also highlighting examples of textures outside of our scope.
We focus on this type of pattern for their importance in design applications and due to the general lack of learning-based methods addressesing their synthesis.

Our approach leverages Latent Diffusion Models~\cite{rombach2022high} as a foundation for the synthesis.
Although these models have achieved significant advancements in natural image generation, they are not optimized to generate structured patterns. One primary limitation is that the synthesized patterns often lack quality, as these models are typically trained to generate photorealistic images with unstructured, chaotic textures and high-frequency, stochastic color variations. When applied to structured patterns, these methods often fail to maintain the inherent structure, sharpness, and cohesive visual style of the patterns.

Furthermore, design applications often require precise pattern controll by the users, often lacking in generative approaches, generally focusing on text-to-image synthesis. While high-level conditioning may be sufficient for natural image synthesis, specifying the exact structure and appearance of a pattern is far more challenging. Even when using images as conditioning inputs, existing methods perform inconsistently in producing structured patterns in our domain.

To address this gap in the literature and offer artists a more accessible yet controllable tool for content creation, we propose a diffusion-based model specifically designed for the synthesis and expansion of structured, stationary patterns.
In particular, we leverage the extensive knowledge already available in large-scale models, such as Stable Diffusion~\cite{rombach2022high,podell2023sdxl}, and adapt it to the patterns domain by training a ``lightweight'' LoRA~\cite{hu2022lora}. Doing so we limit the computational and data requirements of training a diffusion model from scratch, but helps to retain the expressivity of models trained on large-scale datasets like LAION~\cite{schuhmann2021laion}, while adapting it to our specific domain.
To that end, we collect a dataset of procedurally designed patterns that we use to train our LoRA. 

We base our architecture on an inpainting pipeline, which supports the expansion of a partial, hand-drawn input sketch into a larger pattern while preserving its structural integrity and details.
During inference, we leverage noise rolling and patch-based synthesis to produce large-scale, tileable patterns, at high quality in a reliable way. These design choices allow us to generate large-scale, tileable patterns that accurately follow the input sketch, while adding a limited degree of variation avoiding visible repetitions.

We qualitatively evaluate the effectiveness of our approach across a diverse range of input patterns, demonstrating significant improvements over previous state-of-the-art methods in texture synthesis. To assess user satisfaction with generation quality, we also conduct a user study that captures preferences and perceived fidelity in the synthesized outputs.
Additionally, we analyze our architecture through a comprehensive set of experiments and ablation studies to highlight the benefits of our design choices.
The results show that our method consistently generates a wide variety of structure patterns, effectively preserving the structure and visual coherence of the input sketches.
In summary, the contributions of our work are as follows: 
\begin{itemize}
    \item we present a new diffusion-based approach for structured pattern synthesis and expansion;
    \item we introduce a new medium-scale dataset for fine-tuning generative models on the pattern domain;
    \item we demonstrate the generation capabilities of our model for different types of structured patterns and show its ability to control the generation precisely from input sketches; 
    \item we validate the improvements over other generative methods, non-specifically trained for patterns, underlying the need for a specifically trained model.
\end{itemize}

\section{Related Work}
\label{sec:related}

\paragraph{Generative models.} Image generation is a long-standing challenge in computer vision due to the complexity of visual data and the diversity of real-world scenes. With the advent of deep learning, the generation task has been increasingly posed as a learning problem, with Generative Adversarial Networks (GAN)~\cite{goodfellow2014generative} enabling the generation of high-quality images~\cite{karras2017progressive,brock2018large,karras2020analyzing}. However, GANs are characterized by an unstable adversarial training~\cite{arjovsky2017wasserstein,gulrajani2017improved,mescheder2018convergence}, and unable to model complex data distributions~\cite{metz2016unrolled}, exhibiting a \textit{mode collapse} behavior, thus producing a limited variety of outputs.

Diffusion Models (DMs)~\cite{sohl2015deep,ho2020denoising,rombach2022high} have recently emerged as an alternative to GANs, achieving state-of-the-art results in image generation tasks~\cite{dhariwal2021diffusion} also thanks to their stable supervised training approach. Furthermore, DMs have enabled a whole new level of classifier-free conditioning~\cite{ho2022classifier} through cross-attention between latent image representations and conditioning data. More recently, ControlNet~\cite{zhang2023adding} has been proposed to extend generation controllability beyond the typical global-conditioning (e.g.: text prompts) for a fine control over the generation structure. Moreover, approaches like DreemBooth~\cite{ruiz2023dreambooth} and LoRA~\cite{hu2022lora}, allow users to customize large-scale pre-trained models, adapting them to particular tasks or domains or domains, without requiring to finetune them and requiring only a limited set of training samples. 

\paragraph{Generative models for textures synthesis.}
Several works have assessed the synthesis of patterns in the form of natural textures or BRDF materials, with few explicitly focusing on structured patterns.
\citet{heitz2021sliced} address the problem of texture synthesis via optimization by introducing a textural loss based on the statistics extracted from the feature activations of a convolutional neural network optimized for object recognition (e.g. VGG-19).
\citet{vecchio2023controlmat} recently introduced ControlMat to perform SVBRDF estimation from input images, and generation when conditioning via text or image prompts. It employs a novel \textit{noise rolling} technique in combination with patched diffusion to achieve tileable high-resolution generation. MatFuse~\cite{vecchio2023matfuse}, on the other hand, focuses on extending generation control via multimodal conditioning and editing of existing materials via \textit{volumetric inpainting}, to independently edit different material properties. 

Focusing on non-stationary textures, \citet{zhou2018} proposes an example-based synthesis GAN that is trained to double the spatial extent of crops extracted from an arbitrary texture, using a combination of Style and $L_1$ losses. After the GAN is trained, its generator can recurrently being applied to expand texture samples while coherently maintainings its non-stationary features.
\citet{zhou2023neural} introduces a new Guided Correspondence Distance metric that can be employed as a loss function to optimize the texture synthesis process, improving the similarity measurement of output textures to examples. \citet{zhou2024generating}, in contrast, leverages a diffusion model backbone combined with a two-step approach and a \textit{"self-rectification"} technique, to generate seamless texture, faithfully preserving the distinct visual characteristics of a reference example.

Our method synthesizes a large texture from a small seed provided as a user-input sketch, offering versatility without requiring further tuning to expand the pattern while preserving the structure and design properties of the input. In particular, it differs from ControlNet-based methods, which require an additional network and therefore increase computational overhead without significant improvements in performance, in our setting.

\begin{figure*}[!h]
    \centering
    \includegraphics[width=0.95\textwidth]{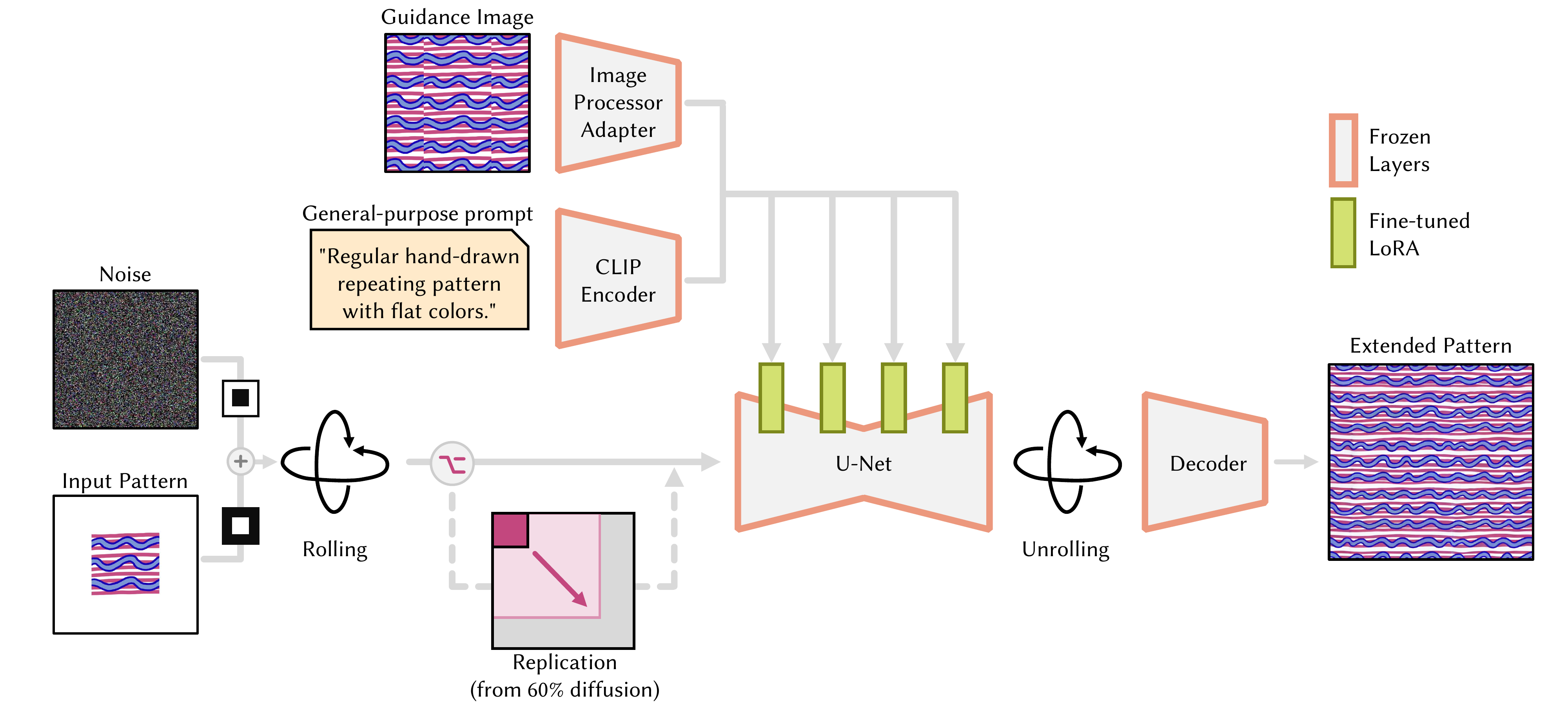}
\caption{\textbf{Architecture.} Given a hand-drawn input pattern, we extend it to an arbitrary-sized canvas, introducing variations while keeping the overall structure and appearance unchanged. Our approach combines text and image conditioning to guide a fine-tuned Latent Diffusion Model~\cite{rombach2022high} to generate high-quality, consistent patterns. At inference, the input pattern is centrally placed within a larger canvas, with our model extending the design outward in a process similar to "outpainting" and effectively filling the entire frame. To ensure tileability of the output, we \textit{roll} the input tensor at each diffusion step and \textit{unroll} afterward. To further extend the pattern beyond the initial canvas scale, we replicate the latent after 60\% diffusion steps.}
    \label{fig:architecture}
\end{figure*}

\section{Method}
\label{sec:method}
Our work is based on the Latent Diffusion (LDM) architecture~\cite{rombach2022high} adapted to synthesize high-quality stationary, structured patterns with a sketched vector-like appearance.
Given a hand-drawn sample, which serves as a seed for expansion, we extend it to an arbitrary-sized canvas, introducing subtle variations while preserving the overall structure and visual consistency. In particular, we leverage the inpainting capabilities of diffusion models, via latent masking, by centrally placing the input pattern in a larger canvas and generating the outer border. Our model extends the design outwards in a process similar to the "outpainting", thus effectively filling the entire frame.

To achieve this, we fine-tune a pre-trained LDM for image generation by training a Low-Rank Adaptation (LoRA) on a dataset of procedurally generated patterns. To further enhance generation fidelity, we employ an IP-Adapter~\cite{ye2023ip} for image prompting, ensuring that the extended design remains visually consistent with the original input, loosely replicated to be provided as a \textit{guidance image}. We additionally use text prompting to constrain the generation to the structural regularity and solid-color look characteristic of our target domain.
To enable seamless extension of patterns to arbitrarily large sizes, we adopt a latent replication strategy, which introduces controlled variations while preserving structural integrity. We also apply the noise rolling technique~\cite{vecchio2023controlmat}, to achieve tileable pattern generation. Specifically, latent replication occurs after every $N$ iterations, while noise rolling and unrolling are applied before and after each diffusion step, respectively.

An overview of our model architecture is shown in Fig.~\ref{fig:architecture}.
In the following, we first provide an overview of the latent diffusion architecture for image generation and the approaches to combine text and image conditioning, to then detail our approach and architectural choices specific to the structured pattern domain. We then ablate our design choices and architectural component in Sec.~\ref{sec:ablation}, demonstrating the benefits of our approach.

\subsection{Guided Image Generation}

\subsubsection{Latent Diffusion Model}
\label{sec:ldm}

We leverage the Latent Diffusion architecture, consisting of a Variational Autoencoder (VAE)~\cite{kingma2013auto} and a diffusion U-Net~\cite{rombach2022high}. The encoder $\mathcal{E}$, compresses an image $\textbf{x} \in \mathbb{R}^{H \times W \times 3}$ into a latent representation $z = \mathcal{E}(\textbf{x})$, where $z \in \mathbb{R}^{h \times w \times c}$, and $c$ is the dimensionality of the encoded image, capturing the essential features in a lower-dimensional space. The decoder $\mathcal{D}$, reconstructs the image from this latent space, effectively projecting it back to the pixel space.

The diffusion process involves a series of transformations that gradually denoise a latent vector, guided by a time-conditional U-Net.
During training, noised latent vectors are generated, following the strategy defined in~\cite{ho2020denoising}, through a deterministic forward diffusion process $q \left( z_t | z_{t-1} \right)$, transforming the encoding of an input image into an isotropic Gaussian distribution. The diffusion network $\epsilon_{\theta}$ is then trained to perform the backward diffusion process $q \left( z_{t-1} | z_t \right)$, effectively learning to ``denoise'' the latent vector and reconstruct its original content.

\subsubsection{Text conditioning}
\label{sec:text_conditioning}

Latent Diffusion models can typically be globally conditioned with high-level text prompts via cross-attention \cite{vaswani2017attention} between each convolutional block of the denoising U-Net and the embedding of the condition $y$, extracted by an encoder $\tau_{\theta}$, with the attention defined as:
\begin{equation}
    \text{Attention}(Q, K, V) = \text{softmax}\left(\frac{Q K^T}{\sqrt{d}} \right) V,
    \label{eq:attn}
\end{equation}
where $Q = W^{i}_{Q} \cdot \varphi_{i}(z_{t})$,  $K = W^{i}_{K} \cdot \tau_{\theta}(y)$, $V = W^{i}_{V}\tau_{\theta}(y)$. Here, $\varphi_{i}(z_{t}) \in \mathbb{R}^{N\times d^{i}_{\epsilon}}$ is the flattened output of the previous convolution block of $\epsilon_{\theta}$, and $W^{i}_{Q} \in \mathbb{R}^{d\times d^{i}_{\tau}}$, $W^{i}_{K} \in \mathbb{R}^{d\times d^{i}_{\epsilon}}$, $W^{i}_{V} \in \mathbb{R}^{d\times d^{i}_{\epsilon}}$, are learnable projection matrices.

The training objective in the conditional setting becomes 
\begin{equation}
    L_{LDM} := \mathbb{E}_{\mathcal{E}(M),y,\epsilon \backsim \mathcal{N}(0, 1), t}\left[\lVert\epsilon - \epsilon_{\theta}(z_t, t, \tau(y))\lVert^2_2\right].
    \label{eq:ldm_objective}
\end{equation}

Openly available LDM implementations use a pre-trained CLIP~\cite{clip} model as feature extractor $\tau$ to encode the text condition. In particular, we use the same CLIP encoder as Stable Diffusion v1.5, relying on a ViT model with a patch size of $14\times14$.

\subsubsection{Image conditioning}
\label{sec:image_conditioning}

Despite the expressive capabilities of text, which has shown remarkable results in the context of natural image synthesis, accurately describing a pattern structure with text is challenging since it would require precise definitions of the pattern shapes, their positions, and symmetries in relation to the other, and their appearance features.

To provide better control of the synthesized pattern, we propose to combine a high-level text prompt, generally valid for all our patterns, with image conditioning via an IP-Adapter~\cite{ye2023ip} model. This lightweight adapter achieves image prompting capability, for pre-trained text-to-image diffusion models, through a decoupled cross-attention mechanism that separates cross-attention layers for text features and image features. 
In particular, the adapter computes separate attention for the text and image embeddings, which are then summed before being fed to the next U-Net layer. The output of the new cross-attention is computed as:
\begin{equation}
    \text{Atten.}(Q, K_t, V_t, K_i, V_i) = \text{softmax}\left(\frac{Q K_t^T}{\sqrt{d}} \right) V_t + \text{softmax}\left(\frac{Q K_i^T}{\sqrt{d}} \right) V_i,
    \label{eq:cross_attn}
\end{equation}
with $K_t, V_t, K_i, V_i$ being respectively the keys and values for the text and image embeddings. During the training of the IP-Adapter, only the image cross-attention layers are trained, while the rest of the diffusion model is kept frozen.

This approach has shown remarkable performances in controlling the generation process with image prompts, allowing it to closely follow the reference image.

\begin{figure}
   \centering
   \setlength{\tabcolsep}{.5pt}
   \begin{tabular}{ccccc}
        & \hspace{-1mm}\small{Input} & \small{Generation} & \hspace{2pt} \small{Input} & \small{Generation} \\
        
        \vspace{-0.5mm} \begin{sideways} \hspace{6.5mm} \small{Text} \end{sideways}\hspace{1mm} & 
        \makecell[b]{\\Purple\\zebra \\ stripes \vspace{3.5mm}} &
        \figframe[width=0.23\linewidth, height=0.23\linewidth]{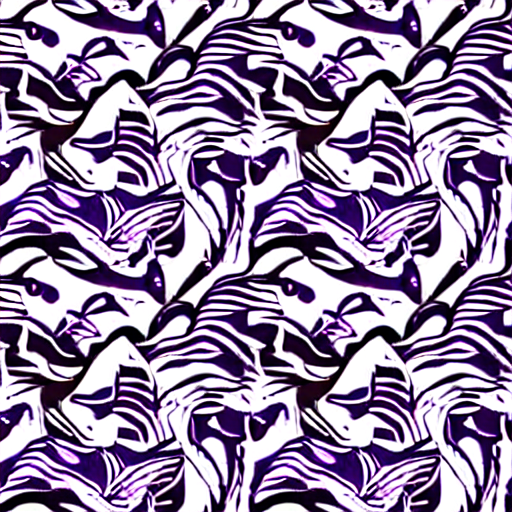} &
        \hspace{2pt} \makecell[b]{\\Cartoon\\brick\\wall \vspace{3.5mm}} &
        \figframe[width=0.23\linewidth, height=0.23\linewidth]{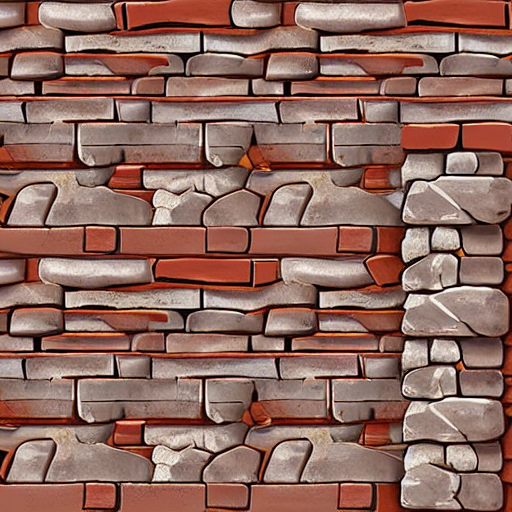} \\
    
        \vspace{-0.5mm} \begin{sideways} \hspace{5.5mm} \small{Image} \end{sideways}\hspace{1mm} & 
        \figframe[width=0.23\linewidth, height=0.23\linewidth]{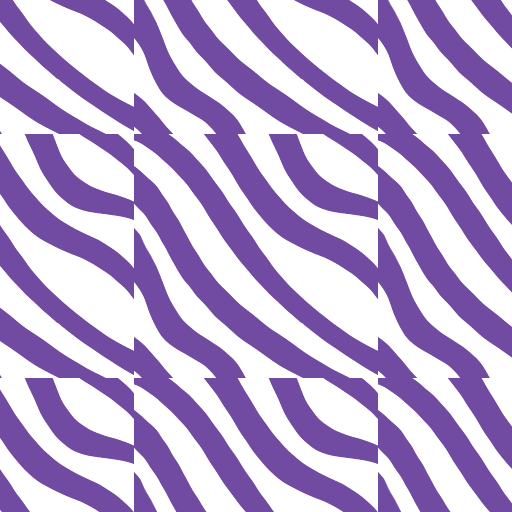} &
        \figframe[width=0.23\linewidth, height=0.23\linewidth]{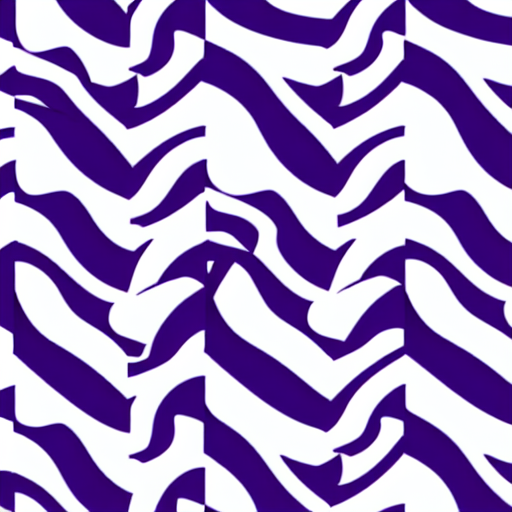} &
        \hspace{2pt} \figframe[width=0.23\linewidth, height=0.23\linewidth]{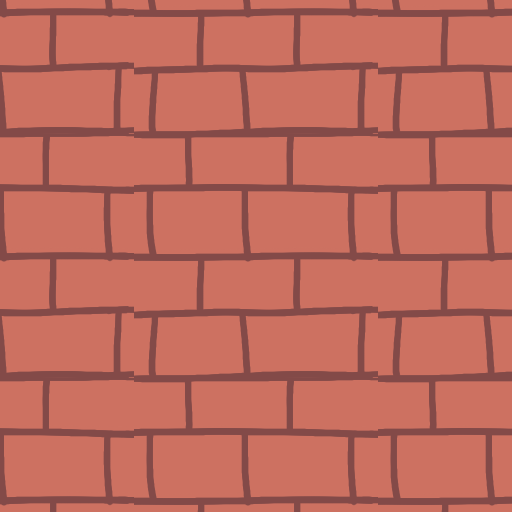} &
        \figframe[width=0.23\linewidth, height=0.23\linewidth]{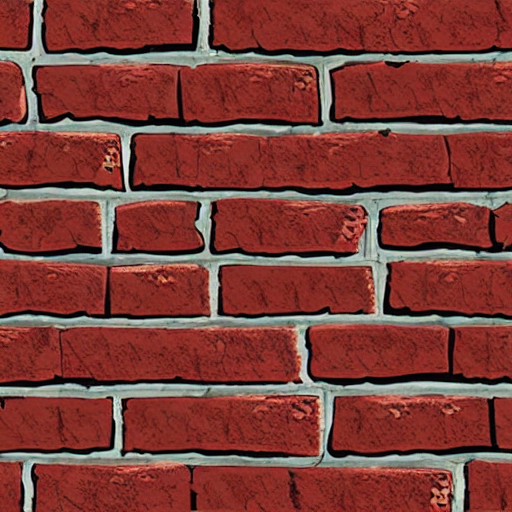} \\
    
        \vspace{-1mm} \begin{sideways} \hspace{3.0mm} \small{Expansion} \end{sideways}\hspace{1mm} &
        \figframe[width=0.23\linewidth, height=0.23\linewidth]{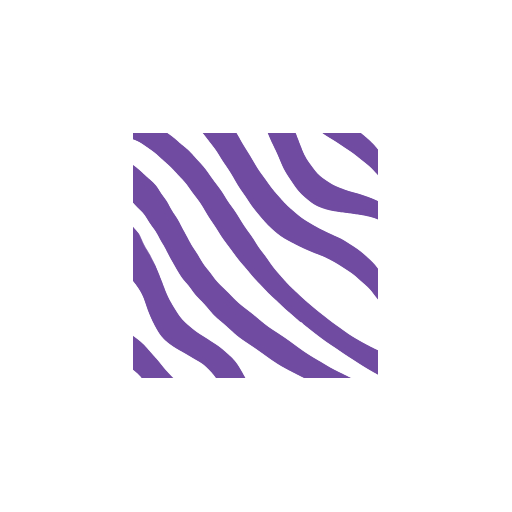} &
        \figframe[width=0.23\linewidth, height=0.23\linewidth]{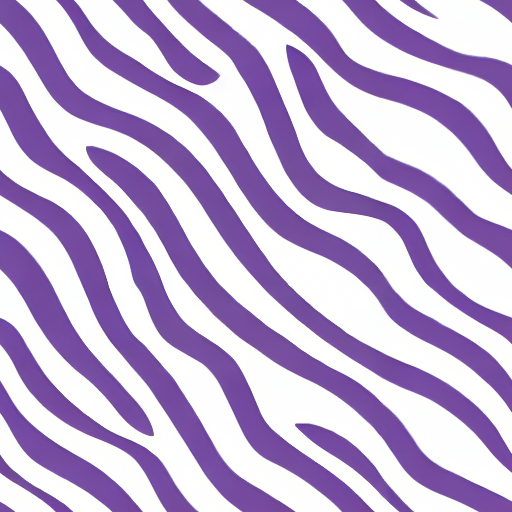} &
       \hspace{2pt} \figframe[width=0.23\linewidth, height=0.23\linewidth]{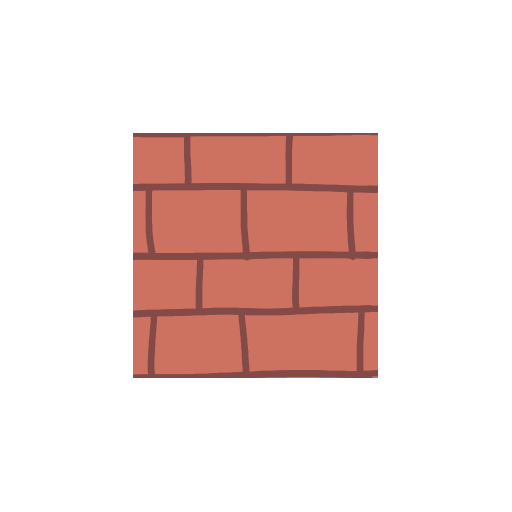} &
        \figframe[width=0.23\linewidth, height=0.23\linewidth]{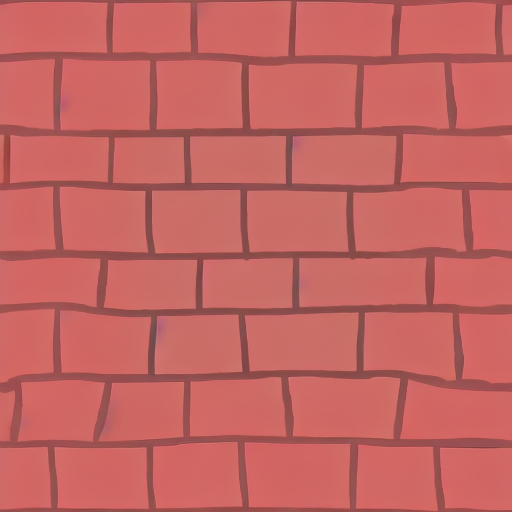} \\
    \end{tabular}
    
\caption{\textbf{Comparison of the generation modalities}. We show the different generation modalities that our base architecture supports. While global conditioning via text or image only (first two rows) is an option, it struggles to generate a consistent pattern and lacks fine-grained user control. In contrast, our expansion approach (last row) enables the generation of arbitrarly large patterns, ensuring consistency with the provided input.}
    \label{fig:generation}
\end{figure}

However, global conditioning through text or image prompts alone lacks the level of detail necessary to capture and reproduce the characteristics of our class of patterns, as shown in Fig.~\ref{fig:generation}. We address these limitations by employing the expansion strategy described in Sec.~\ref{sec:pattern_generation}.

\begin{figure*}[!h]
    \centering
    \setlength{\tabcolsep}{.5pt}
    \begin{tabular}{ccc}        
        \vspace{-0.5mm}\hspace{-1mm}
        \figframe[width=0.32\textwidth, height=0.32\textwidth]{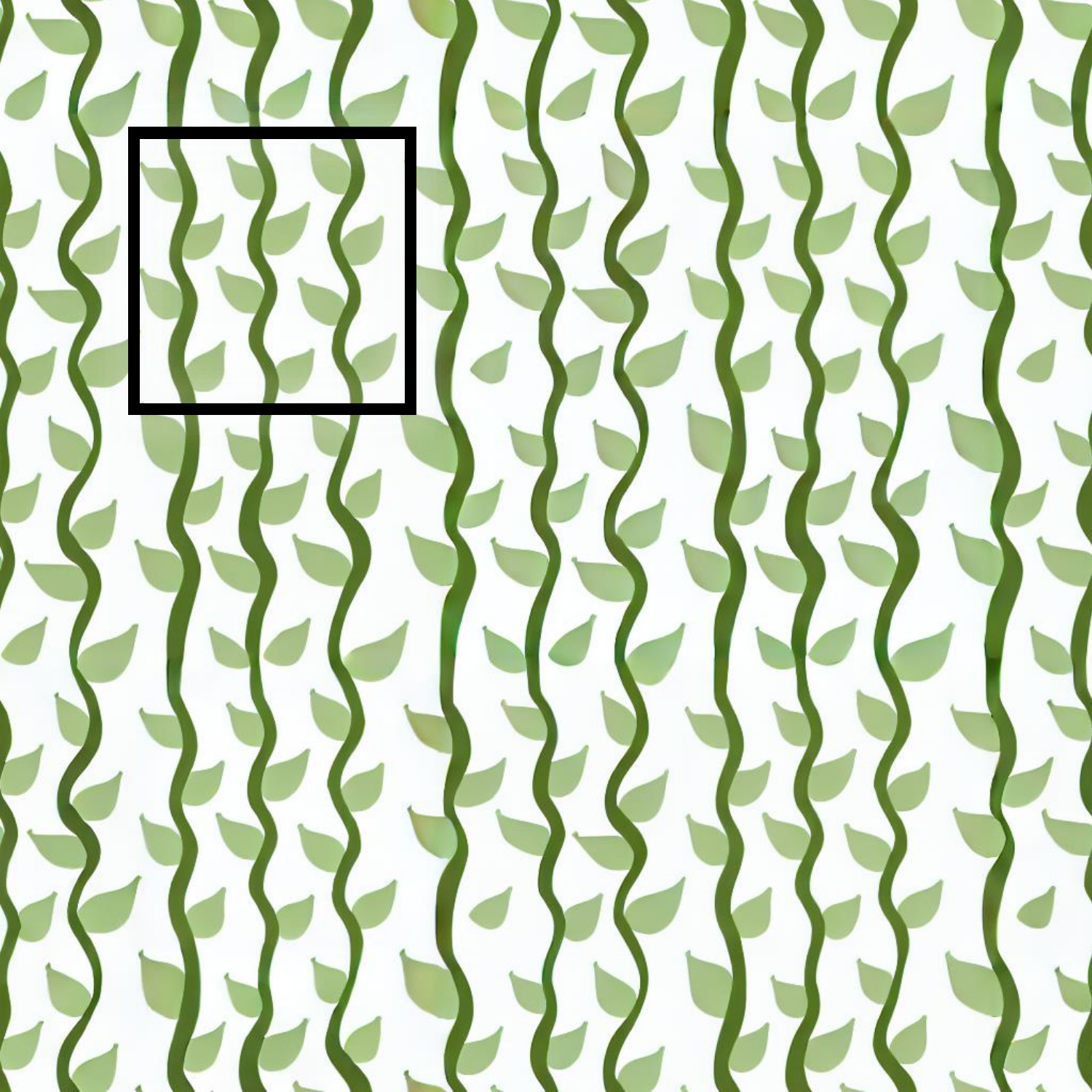}
        \figframe[width=0.32\textwidth, height=0.32\textwidth]{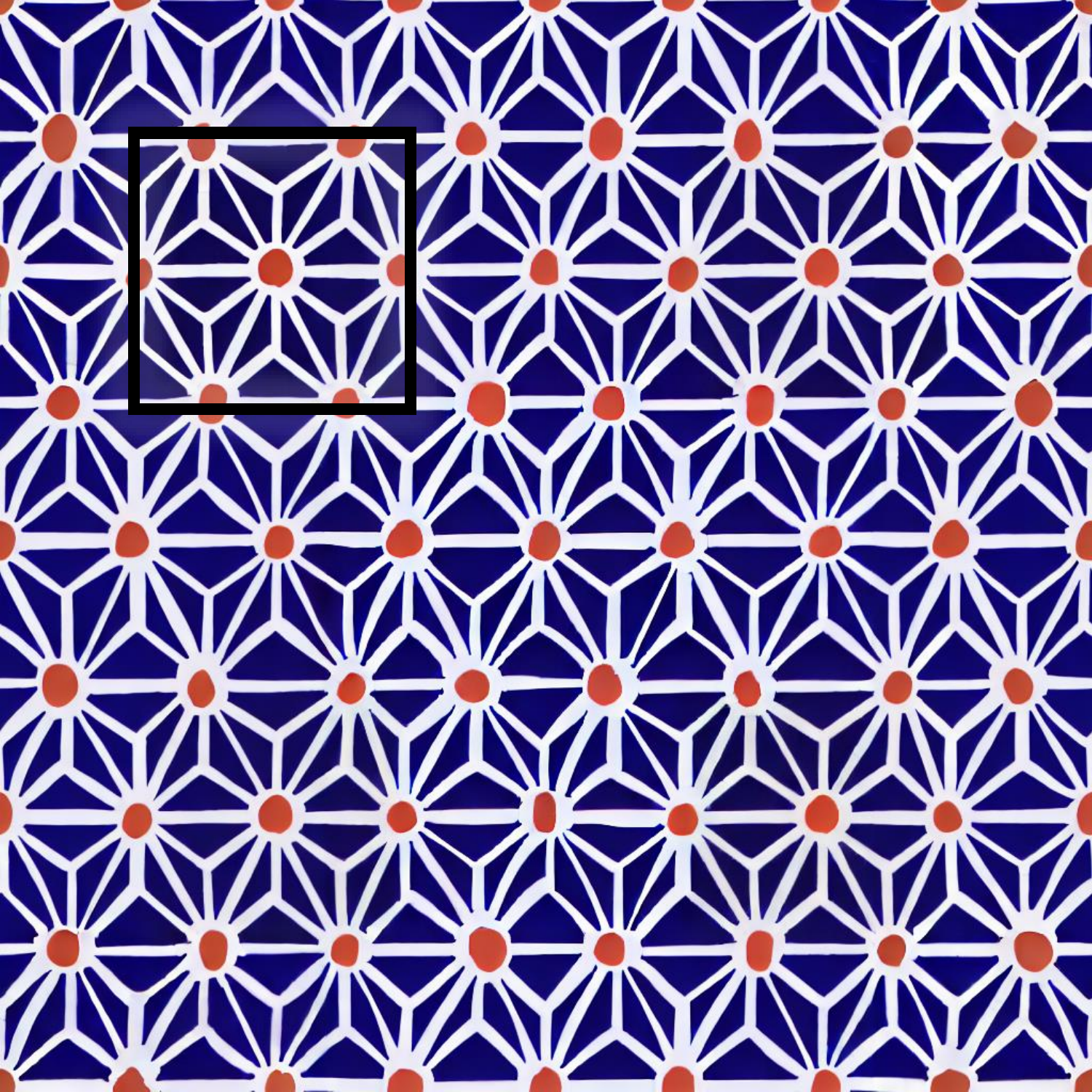} &
        \figframe[width=0.32\textwidth, height=0.32\textwidth]{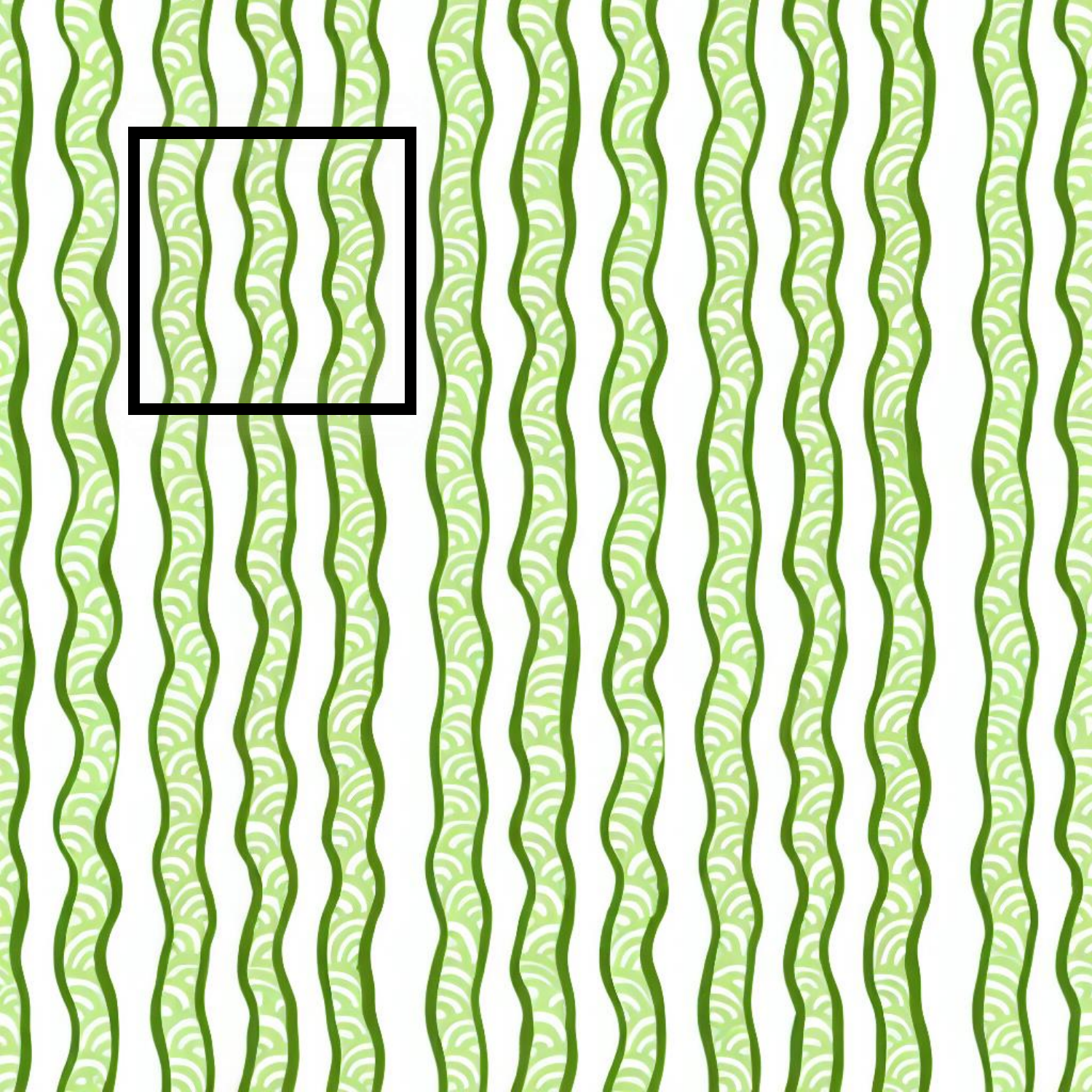} \\
        \end{tabular}
\caption{\textbf{Pattern expansion.} The figure illustrates the pattern expansion capabilities of our model, where the input patterns are contained within the black boxes. The left and right samples showcase organic patterns, demonstrating the model's ability to extrapolate intricate designs while maintaining a natural flow. The central panel presents a more structured geometric pattern, which is also successfully expanded to cover a larger area. This example highlights the model's versatility in handling both organic and geometric patterns.}
    \label{fig:pattern_expansion}
\end{figure*}

\subsection{Stable Diffusion Finetuning}
\label{sec:finetuning}
To achieve visually coherent pattern synthesis and expansion, we fine-tune the Stable Diffusion 1.5 model~\cite{rombach2022high} to our specific pattern domain. In particular, we leverage a Low-Rank Adaptation (LoRA) technique~\cite{hu2022lora} for efficient fine-tuning of large, pre-trained models, limiting the number of training parameters, while also avoiding catastrophic forgetting, which is the tendency of a model to lose previously learned knowledge when fine-tuned on new data.

In particular, we train a low-rank matrix and add it to the transformer layers of the base Latent Diffusion Model (LDM):
\begin{equation}
\theta' = \theta + \Delta\theta,
\end{equation}
where $\theta$ represents the original weights of the transformer in the LDM, and $\Delta\theta$ is the low-rank update, computed as:
\begin{equation}
\Delta\theta = U \cdot V^T,
\end{equation}
with $U\in\mathbb{R}^{r\times d}$ being the trainable matrices, and $r$ much smaller than $d$, the dimensionality of the layer's parameters.

This fine-tuning step focuses the generation on the pattern domain and is mostly responsible for the model's ability to maintain stylistic consistency and detailed coherence specific to the target patterns. By introducing these low-rank updates, we ensure that the model adapts efficiently to the specific feature of the pattern domain, without losing its expressive capabilities from the training on the image domain.

\subsection{Pattern expansion}
\label{sec:pattern_generation}
To expand patterns to arbitrary sizes while ensuring aesthetic coherence and tileability, we rely on the Stable Diffusion 1.5 model trained for image inpainting~\cite{rombach2022high}. By leveraging its capability to interpret partial images and generate coherent completions, we combine inpainting with latent replication and noise rolling~\cite{vecchio2023controlmat} to produce high-quality, tileable expansions. This ensures that the extended patterns retain consistency with the original input, preserving the overall visual appearance, as demonstrated in Fig.~\ref{fig:generation} and Fig.~\ref{fig:pattern_expansion}.

In particular, we start the denoising process at the model's native resolution of $512\times512$, placing the input at the center of the canvas and leveraging the inpainting capabilities of Stable Diffusion to fill the masked area. However, inpainting can effectively reconstruct missing parts, it tends to lose long-term dependencies inside the image, leading to inconsistencies in the global structure. This limitation arises from the local receptive field of the model, which makes it challenging to maintain consistency in regions farther from the fixed input pattern.
As a result, border artifacts can appear, disrupting the overall structure on the expanded canvas. To address this, we incorporate noise rolling, which cyclically shifts the latent representation $z_t$ at each diffusion step. This technique effectively refocuses the model’s receptive field, allowing it to capture a broader spatial context and better preserve structural coherence throughout the pattern. In particular, for each diffusion step, we compute:

\begin{equation}
z'_{t} = \text{roll}(z_t, \Delta x, \Delta y),
\end{equation}

where $\text{roll}(\cdot, \Delta_x, \Delta_y)$ denotes the cyclic shift operation along the image's width ($\Delta_x$) and height ($\Delta_y$).
After rolling, the model estimates the noise component and performs a denoising step, computing $z'_{t-1}$. Subsequently, the latent space is unrolled back to its original configuration to maintain the integrity of the global pattern structure:
\begin{equation}
z_{t-1} = \text{roll}(z'_{t-1}, -\Delta x, -\Delta y).
\end{equation}
By manipulating the latent space in this manner, the model effectively treats the pattern's edges as interconnected, thus intrinsically minimizing the presence of visible seams.

Finally, to be able to cover an arbitrarily large canvas, we replicate the latent up to the target canvas size after $N$ iterations and perform the remaining denoising steps using patched diffusion.
In our experiments, we set $N$ to the 60\% of the inference steps, resulting in the best compromise between pattern consistency and variation. By combining latent replication and noise rolling, we support a larger expansion while guaranteeing the quality of the generated patterns, as demonstrated in Sec.~\ref{sec:results} and in the ablation study in Sec.~\ref{sec:ablation}.

\begin{figure}
   \centering
   \setlength{\tabcolsep}{.5pt}
   \begin{tabular}{cccc}
        \figframe[width=0.23\linewidth, height=0.23\linewidth]{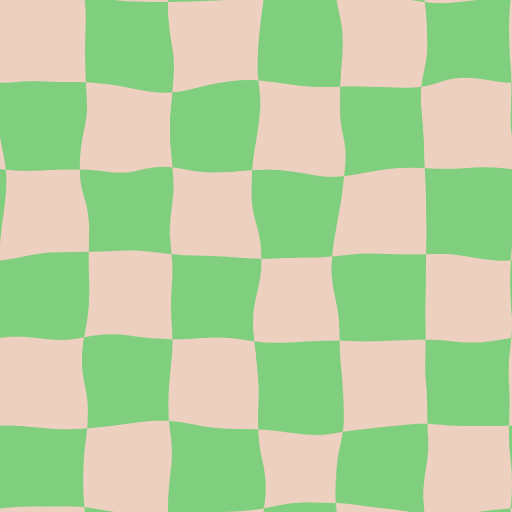} &
        \figframe[width=0.23\linewidth, height=0.23\linewidth]{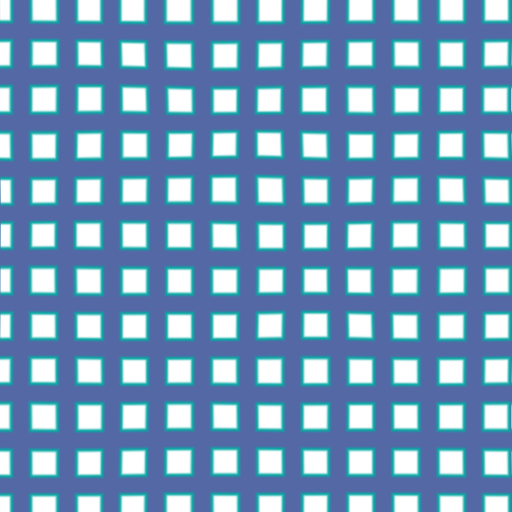} &
        \figframe[width=0.23\linewidth, height=0.23\linewidth]{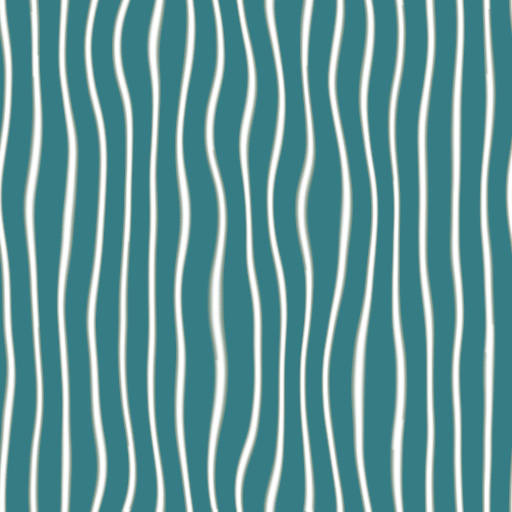} &
        \figframe[width=0.23\linewidth, height=0.23\linewidth]{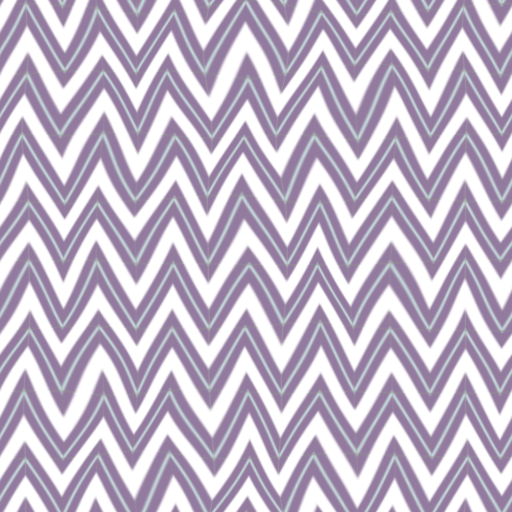} \\
        \figframe[width=0.23\linewidth, height=0.23\linewidth]{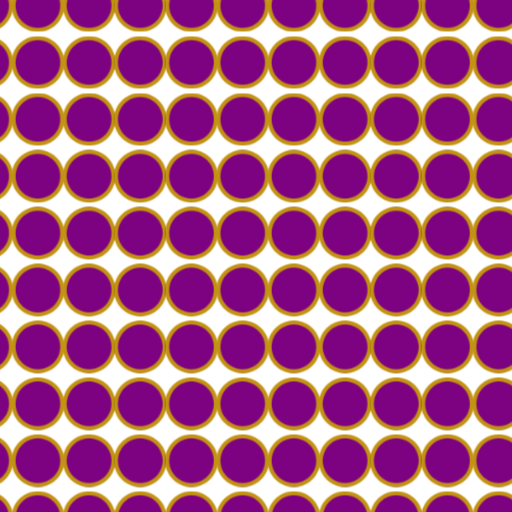} &
        \figframe[width=0.23\linewidth, height=0.23\linewidth]{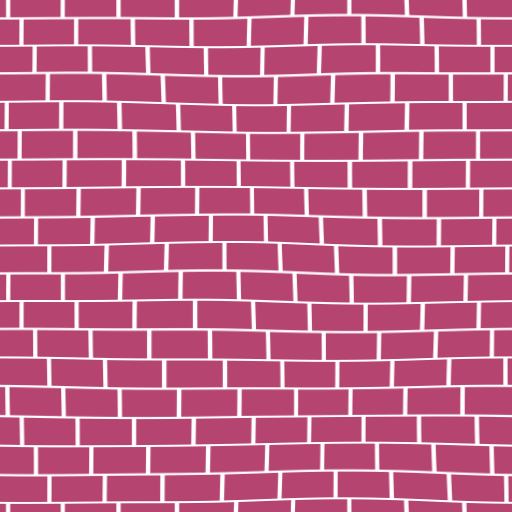} &
        \figframe[width=0.23\linewidth, height=0.23\linewidth]{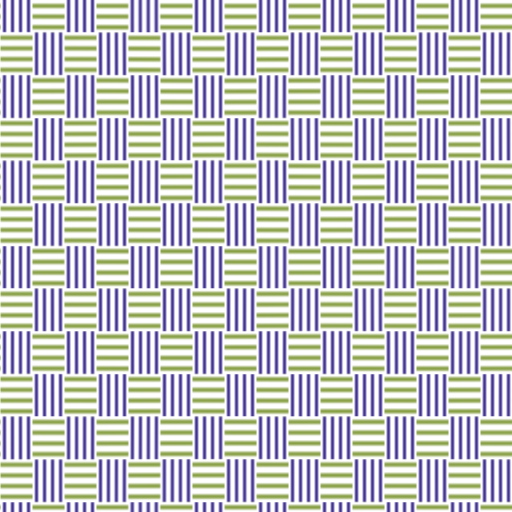} &
        \figframe[width=0.23\linewidth, height=0.23\linewidth]{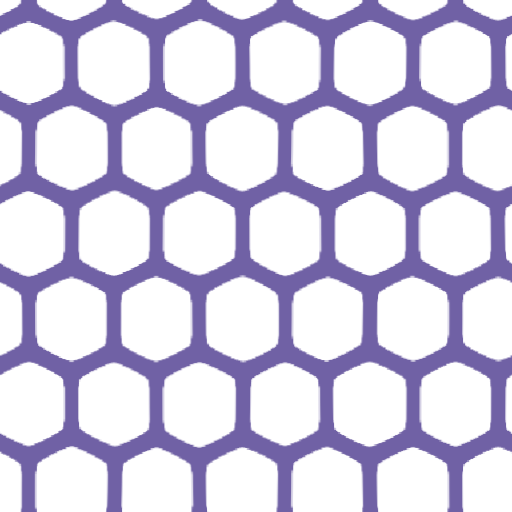} \\     
    \end{tabular}
    \caption{\textbf{Pattern samples from the dataset}. Samples from each class in our custom pattern dataset, namely \textit{grid}, \textit{checker}, \textit{stripes}, \textit{zigzag}, \textit{dots}, \textit{bricks}, \textit{metal} and \textit{hexagons}. For each pattern example, we use the generating procedural parameters to define a caption matching its design details, thus creating pattern-text pairs that are used while training the LoRA.}
    \label{fig:dataset}
\end{figure}

\begin{figure*}
    \centering
    \setlength{\tabcolsep}{.5pt}
        \begin{tabular}{cccc|}
            \hspace{-1mm}\small{Init} & \hspace{-1mm}\small{SD base (a)} & \hspace{-1.75mm}\small{+ IP-Adapter (b)} & \hspace{-1.75mm}\small{+ LoRA (c)} \\
            
            \vspace{-0.5mm}
            \figframe[width=0.093\textwidth, height=0.093\textwidth]{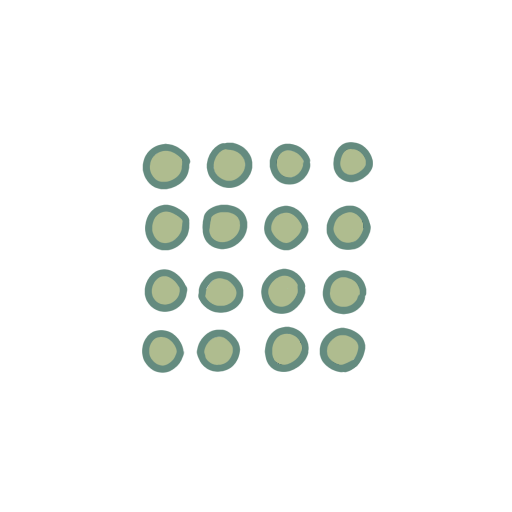} &
            \figframe[width=0.093\textwidth, height=0.093\textwidth]{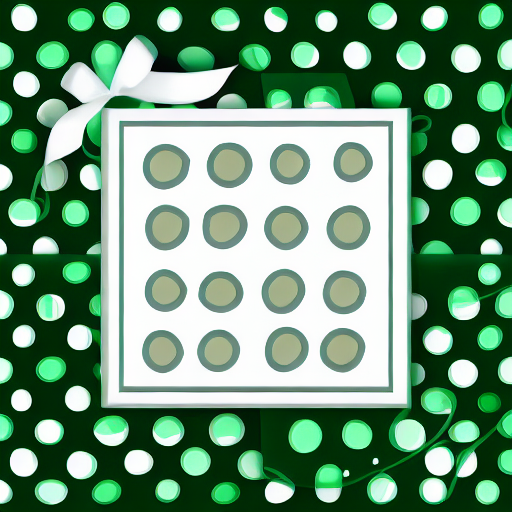} &
            \figframe[width=0.093\textwidth, height=0.093\textwidth]{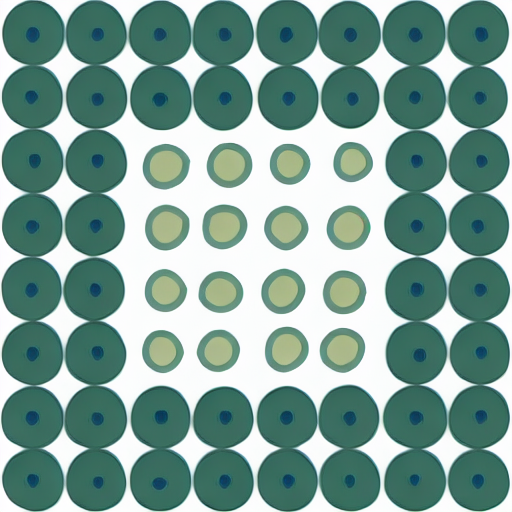} & 
            \figframe[width=0.093\textwidth, height=0.093\textwidth]{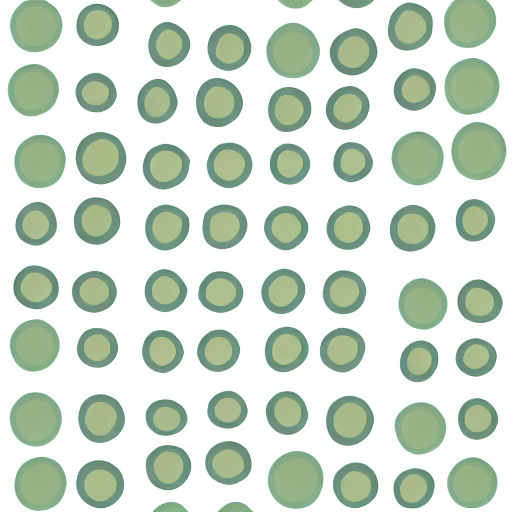} \\

            \vspace{-0.5mm}
            \figframe[width=0.093\textwidth, height=0.093\textwidth]{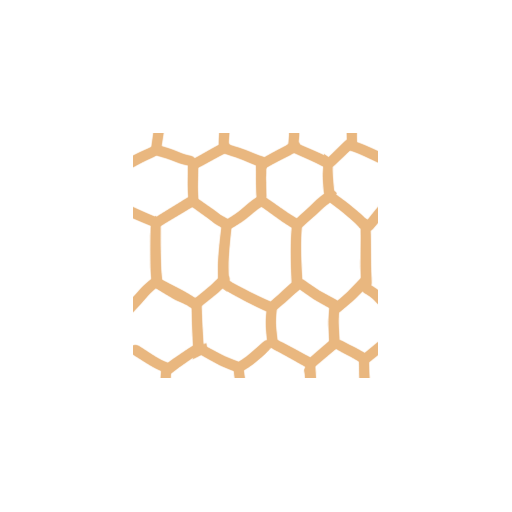} &
            \figframe[width=0.093\textwidth, height=0.093\textwidth]{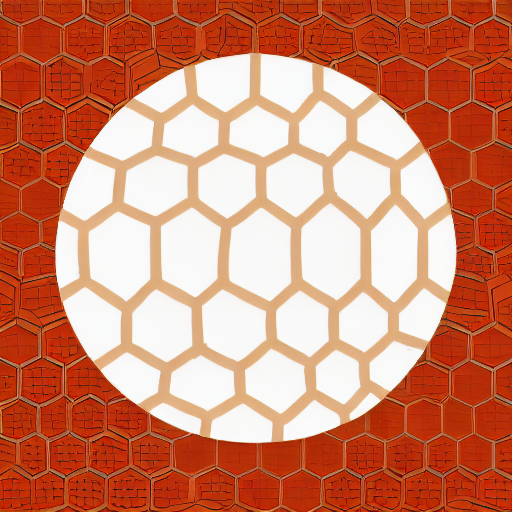} &
            \figframe[width=0.093\textwidth, height=0.093\textwidth]{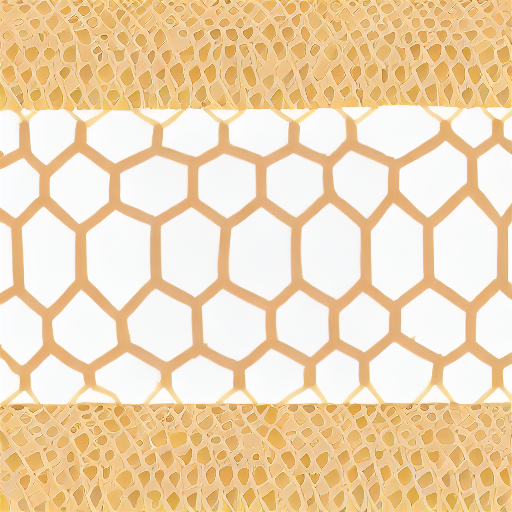} & 
            \figframe[width=0.093\textwidth, height=0.093\textwidth]{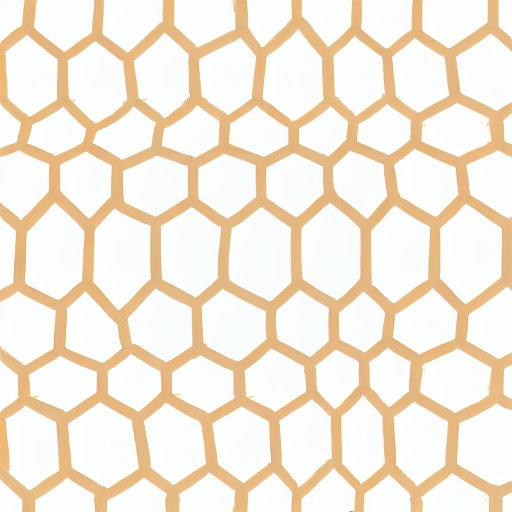} \\
            
            \vspace{-0.5mm}
            \figframe[width=0.093\textwidth, height=0.093\textwidth]{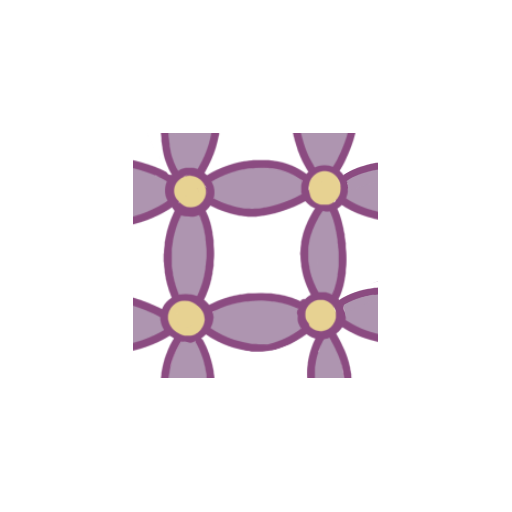} &
            \figframe[width=0.093\textwidth, height=0.093\textwidth]{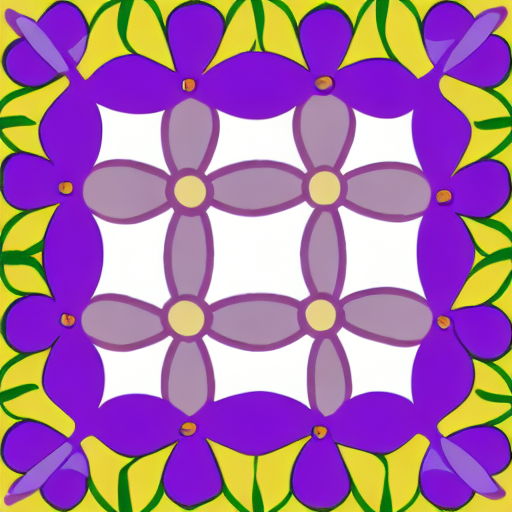} &
            \figframe[width=0.093\textwidth, height=0.093\textwidth]{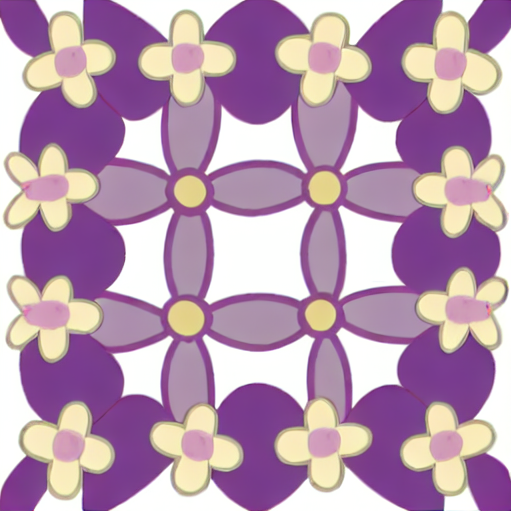} & 
            \figframe[width=0.093\textwidth, height=0.093\textwidth]{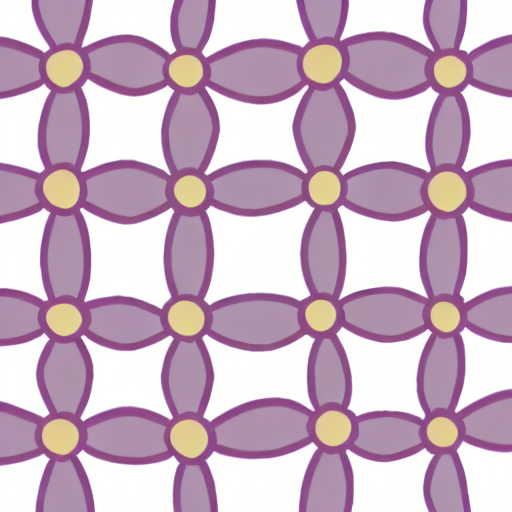} \\
        \end{tabular}
        \begin{tabular}{cc}
            \small{Expansion (d)} & \hspace{-1mm}\small{Expansion + Noise Rolling (e)} \\
            \hspace{0.25mm}
            \figframe[width=0.29\textwidth, height=0.29\textwidth]{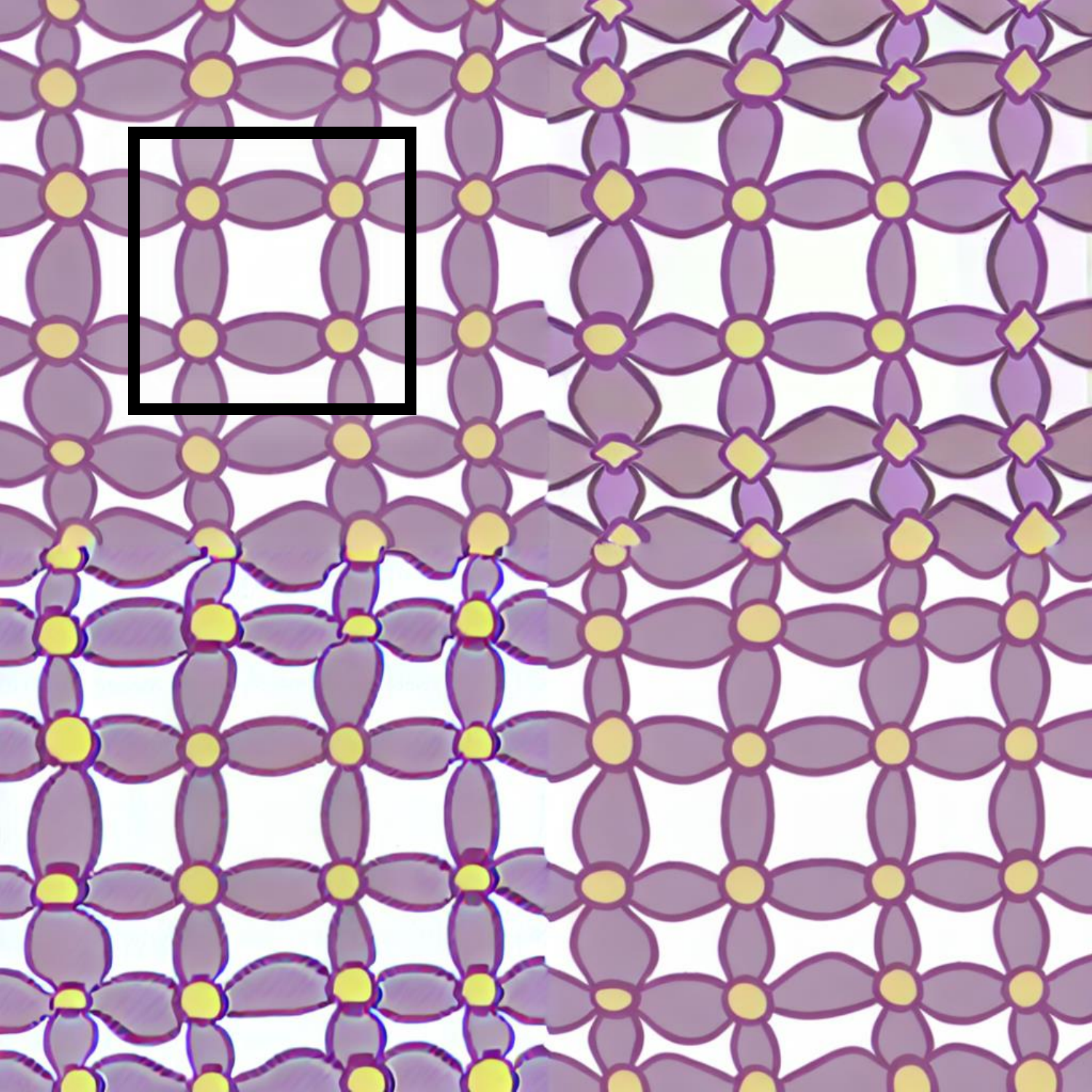} &
            \figframe[width=0.29\textwidth, height=0.29\textwidth]{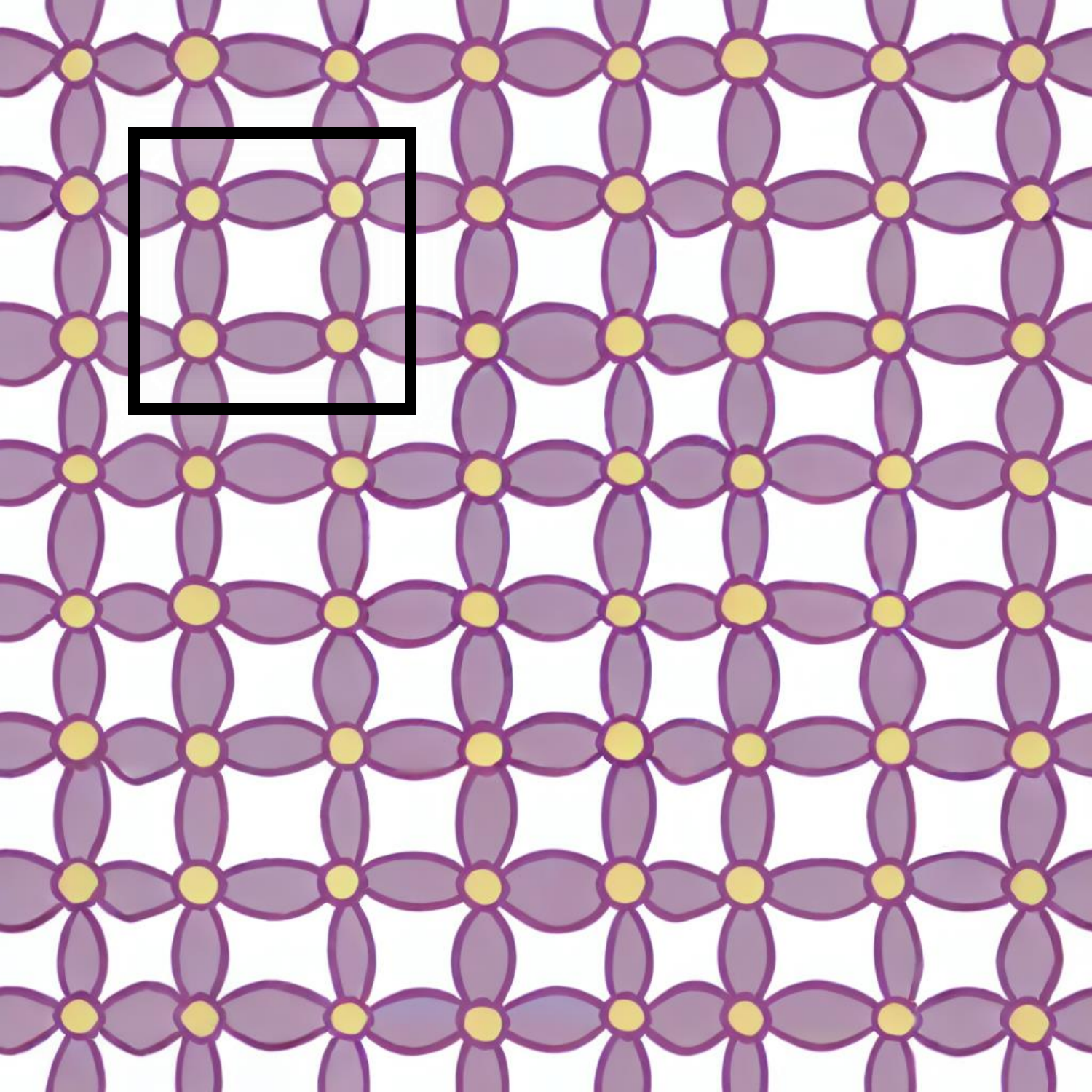}\\
        \end{tabular}
\caption{\textbf{Ablation study.} We show the performance improvements with the introduction of each design choice. The base inpainting model is unable to expand the pattern while maintaining visual coherence. The introduction of image conditioning via the IP-Adapter improves generation consistency with the prompt. The LoRA finetuning, on a small dataset, greatly enhances generation quality, ensuring visual consistency over the entire generation. Finally, the introduction of noise rolling enables tileable generation and removes repetition seams and visual artifacts.}
    \label{fig:ablation}
\end{figure*}

\begin{figure}[h]
    \centering
    \includegraphics[width=0.95\linewidth]{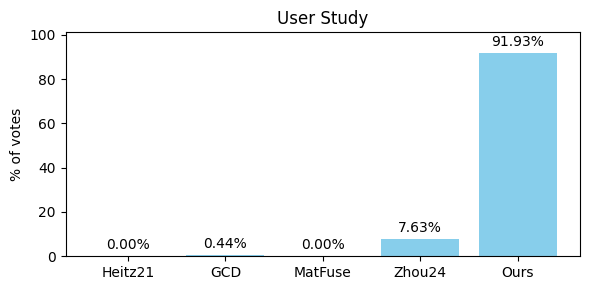}
    \caption{Each of the 80 participants in our user study had to indicate their preference for pattern expansion on a subset of 20 generations that were chosen at random. According to the results, our approach is favored in the vast majority of cases (91.03\%), with \cite{zhou2024generating} coming far in second (7.63\%) and other methods being negligible. This demonstrates the excellent quality achieved by our pattern expansions, which far surpass existing methods. }
    \label{fig:user-study}
\end{figure}

\section{Experimental results}
\label{sec:results}

\subsection{Datasets}
\label{sec:datasets}
Due to the lack of publicly available pattern datasets, we created a custom dataset consisting of 4000 patterns of 8 classes, namely \textit{grids}, \textit{checkers}, \textit{stripes}, \textit{zigzag}, \textit{dots}, \textit{bricks}, \textit{metal} and \textit{hexagons}, each of which is showcased in Fig.~\ref{fig:dataset}. Such classes were specifically designed to expose strong geometrical structures and shape arrangements, to help the LoRA learn the key features of structured pattern domains. For each class, we defined an ad-hoc procedural program capable of generating a diverse set of samples in both design and colors. To simulate a sketched style, we combined our patterns with varying scales of Perlin noise~\cite{perlin85}, introducing the irregularities commonly found in hand-drawn designs.

Our dataset consists of procedurally generated pattern-text caption pairs. We generate each pattern by randomly sampling a value in the proper range for each procedural parameter, including colors. We use these values to build the descriptive caption that highlights the main feature of the pattern. We use a caption template for each pattern class, that is filled with details drawn from the procedural parameter values. As an example, the caption matching the checkered pattern in Fig.~\ref{fig:dataset} (top left) is generated from the base caption of \textit{``A hand-drawn checkered pattern. Checkers are colored in \texttt{<even\_color>} and \texttt{<odd\_color>}, and their size is \texttt{<checker\_size>}. Checkers are surrounded on all four sides by a checker of a different color. Colors are flat and without shading.''}, where the free variables are completed by ``light green'', ``wheat'' and ``big'' respectively. 

For each class, we sample 500 different parameter sets and generate the corresponding pattern-text pairs for training.

For inference, we manually sketched $33$ small pattern samples on a graphics tablet, whose expansion is shown in Fig.~\ref{fig:teaser}, Fig.~\ref{fig:pattern_expansion}, Fig.~\ref{fig:results_highres} as well as in all the experiments reported in this paper.

\subsection{Technical details}
\label{sec:training}

\paragraph{Training}
We train our LoRA with a mini-batch gradient descent, using the Adam~\cite{adam} optimizer with a learning rate set to $10^{-4}$ and a batch size of 8. The training is carried out for 5000 iterations on a single NVIDIA RTX3090 GPU with 24GB of VRAM, using the pre-trained inpainting Stable Diffusion 1.5 checkpoint from~\cite{rombach2022high}.

\paragraph{Inference}
\label{sec:inference}
Generation is performed by denoising a latent random noise for 50 steps, using the DDIM sampler~\cite{song2020denoising} with a fixed seed.
Pattern expansion takes about 2 seconds at $512\times512$ and 4 seconds at $1024\times1024$ and 6GB of VRAM, about 12 seconds at $2048\times2048$ and 8GB VRAM. Memory and requirements can be further reduced by processing fewer patches in parallel, albeit at the cost of increased computation time.

\subsection{Results and comparisons}
\label{sec:comparison}

We evaluate our model generation capabilities when conditioning using either text or image. Despite not being the main focus of this work, we show that our architecture is able to generate pattern-like images when being globally conditioned. However, as shown in Fig.~\ref{fig:generation}, the generation style tends to diverge significantly from the guidance image, highlighting the need for stronger constraints for a specific pattern expansion that closely follows the input sample.

\paragraph{Expansion results}

We evaluate our model generation capabilities for pattern expansion (Fig.~\ref{fig:teaser}, Fig.~\ref{fig:pattern_expansion} and Fig.~\ref{fig:results_highres}). The results demonstrate that our method closely follows the input prompt, highlighting the pattern expansion capabilities of our model both in terms of quality and coherence of the expanded result.

All the results figures, present the original input pattern contained within a black box, while the surrounding part of the canvas is filled during inference time. Our pipeline successfully extends the input pattern, maintaining coherence and preserving the structural integrity of the original design. Each generated pattern flows naturally from the input, ensuring that there are no abrupt transitions or noticeable repetitions. The model keeps color consistent in the generated area, matching the original input. Due to the adoption of the \textit{noise rolling} technique, all results are tileable, thus allowing seamless repetition. All the provided examples use an expansion factor of 2 for both width and height dimensions.

\begin{figure*}
    \centering
    \setlength{\tabcolsep}{.5pt}
    \begin{tabular}{cccccc}
        \hspace{-1mm}\small{Init} & \small{(a) \citet{heitz2021sliced}} & \small{(b) GCD} & \small{(c) MatFuse} & \small{(d) \citet{zhou2024generating}} & \textbf{Our method}\\
        
        \vspace{-0.5mm}\hspace{-1mm}
        \figframe[width=0.166\textwidth, height=0.166\textwidth]{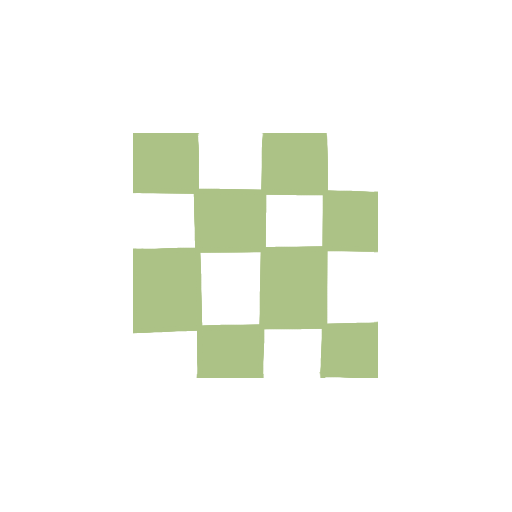} &
        \figframe[width=0.166\textwidth, height=0.166\textwidth]{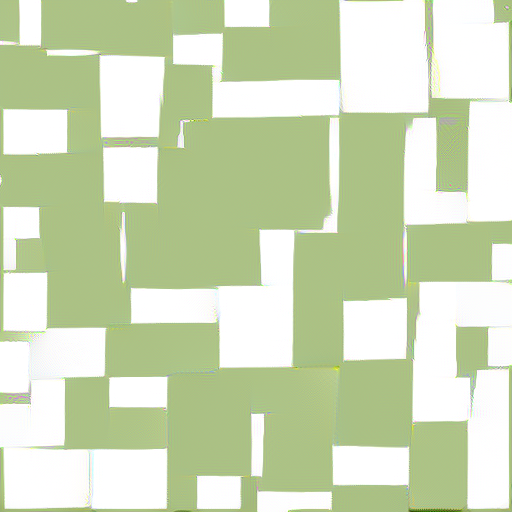} &
        \figframe[width=0.166\textwidth, height=0.166\textwidth]{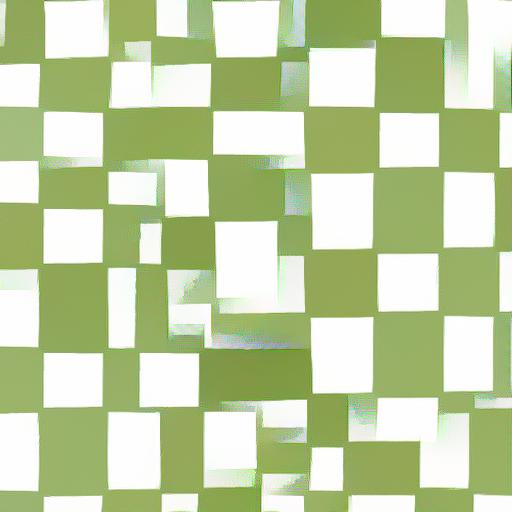} & 
        \figframe[width=0.166\textwidth, height=0.166\textwidth]{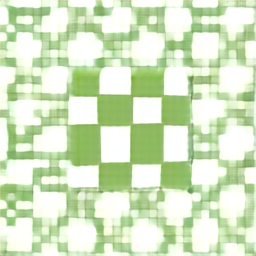} &
        \figframe[width=0.166\textwidth, height=0.166\textwidth]{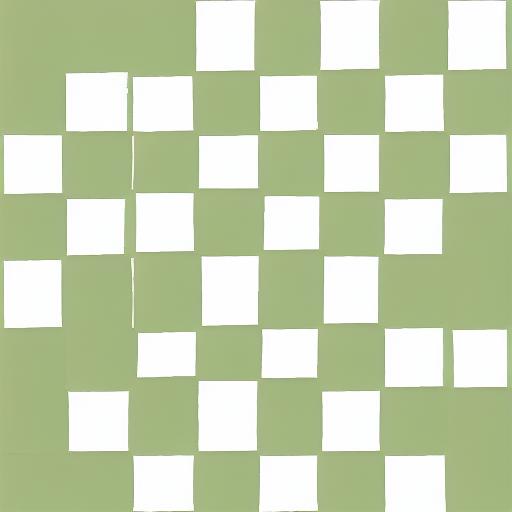} &
        \figframe[width=0.166\textwidth, height=0.166\textwidth]{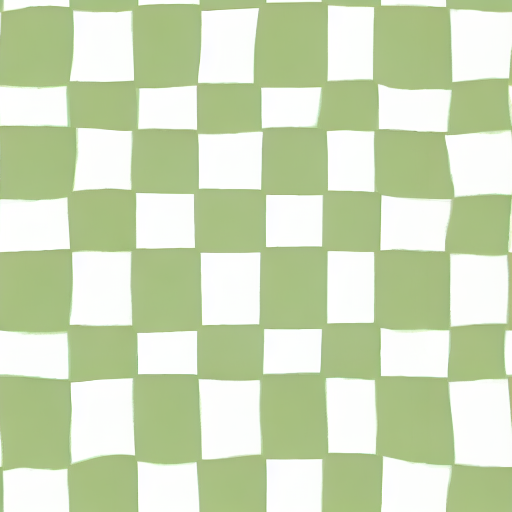} \\
    
        \vspace{-0.5mm}\hspace{-1mm}
        \figframe[width=0.166\textwidth, height=0.166\textwidth]{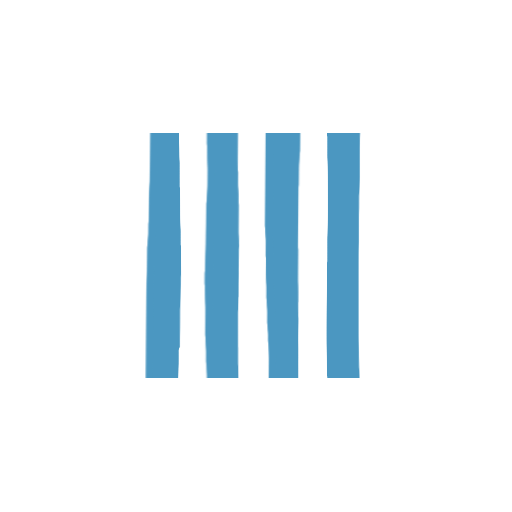} &
        \figframe[width=0.166\textwidth, height=0.166\textwidth]{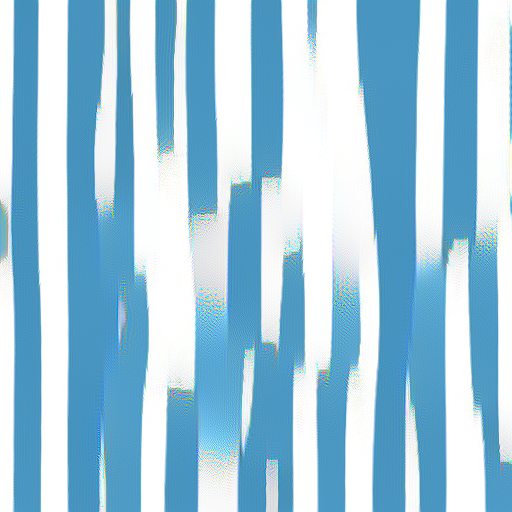} &
        \figframe[width=0.166\textwidth, height=0.166\textwidth]{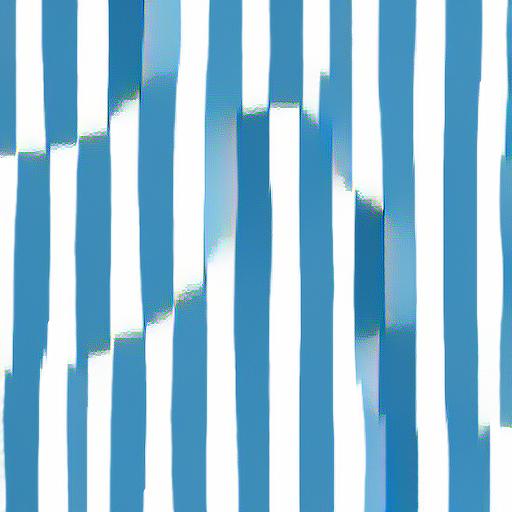} & 
        \figframe[width=0.166\textwidth, height=0.166\textwidth]{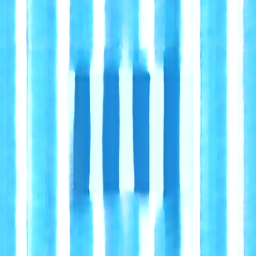} &
        \figframe[width=0.166\textwidth, height=0.166\textwidth]{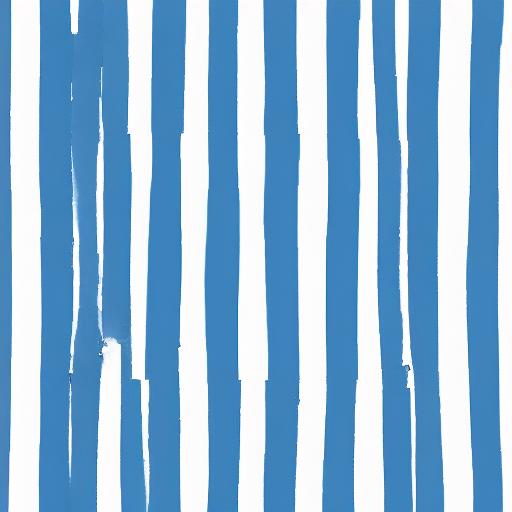} &
        \figframe[width=0.166\textwidth, height=0.166\textwidth]{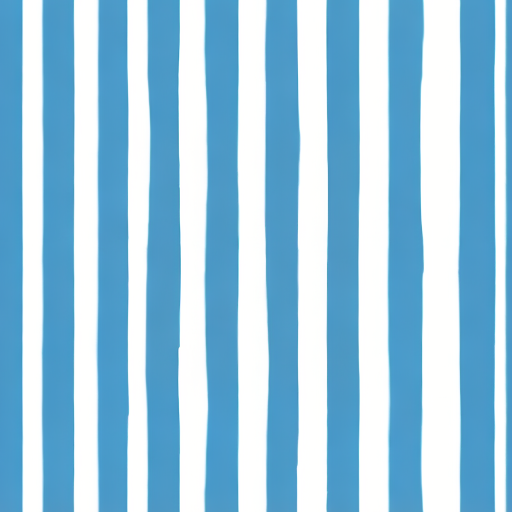} \\

        \vspace{-0.5mm}\hspace{-1mm}
        \figframe[width=0.166\textwidth, height=0.166\textwidth]{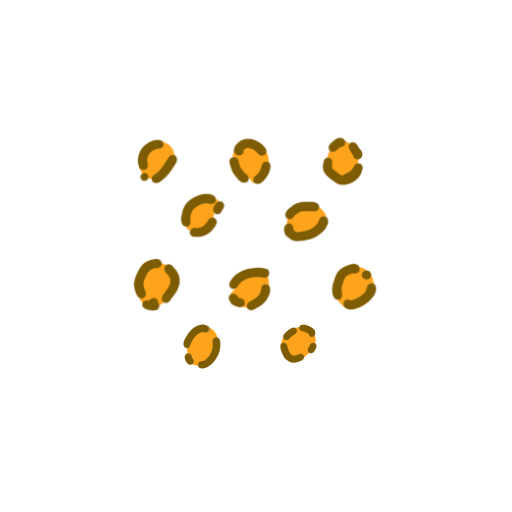} &
        \figframe[width=0.166\textwidth, height=0.166\textwidth]{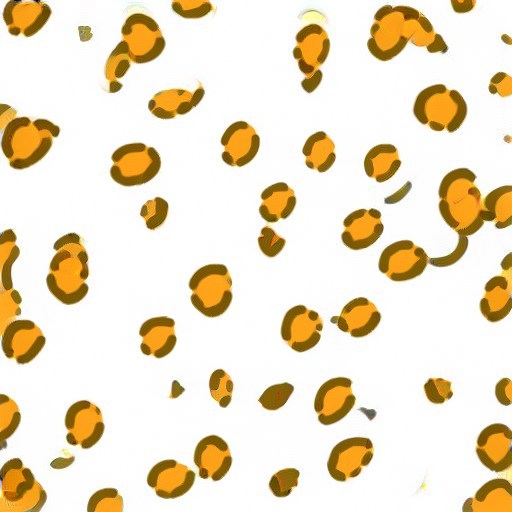} &
        \figframe[width=0.166\textwidth, height=0.166\textwidth]{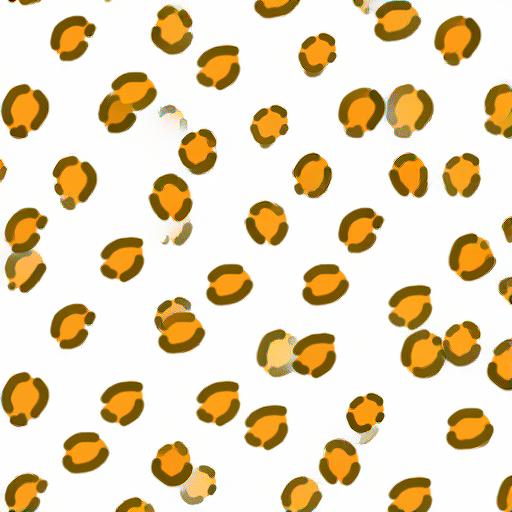} & 
        \figframe[width=0.166\textwidth, height=0.166\textwidth]{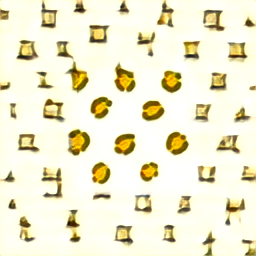} &
        \figframe[width=0.166\textwidth, height=0.166\textwidth]{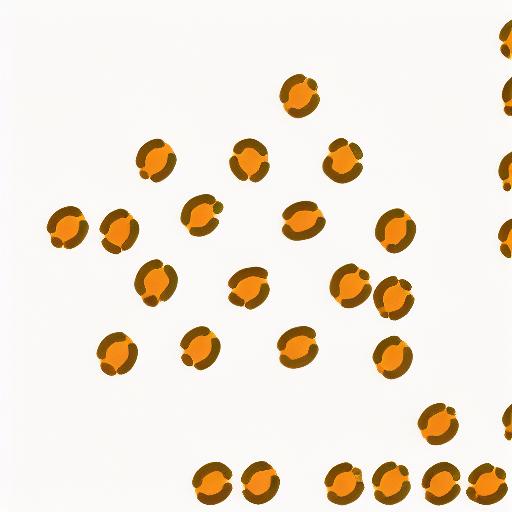} &
        \figframe[width=0.166\textwidth, height=0.166\textwidth]{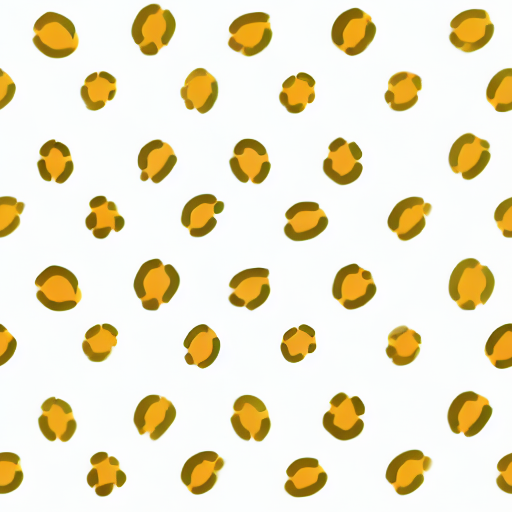} \\

        \vspace{-0.5mm}\hspace{-1mm}
        \figframe[width=0.166\textwidth, height=0.166\textwidth]{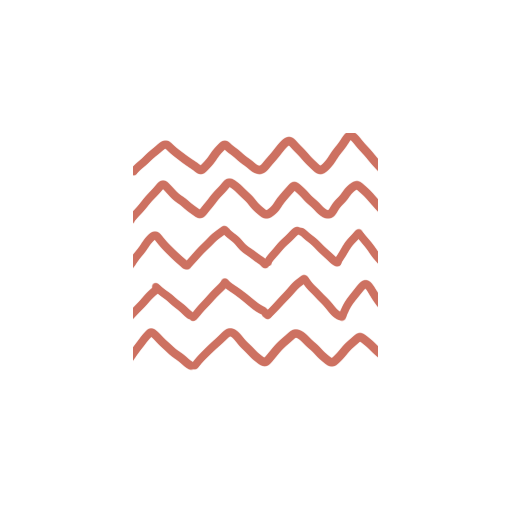} &
        \figframe[width=0.166\textwidth, height=0.166\textwidth]{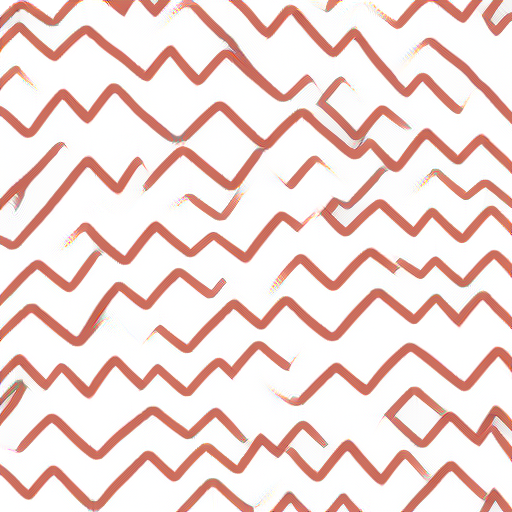} &
        \figframe[width=0.166\textwidth, height=0.166\textwidth]{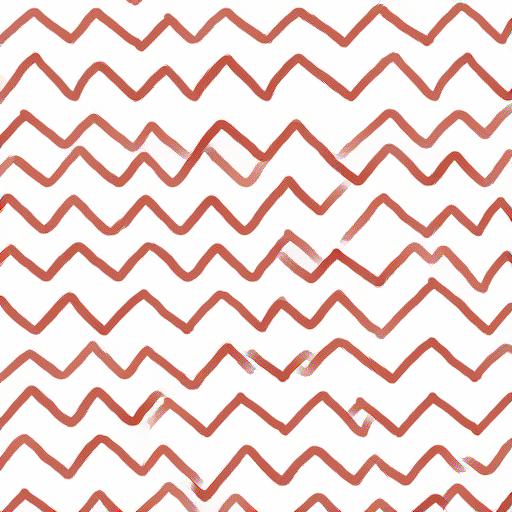} & 
        \figframe[width=0.166\textwidth, height=0.166\textwidth]{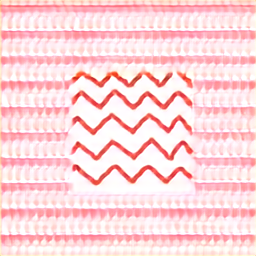} &
        \figframe[width=0.166\textwidth, height=0.166\textwidth]{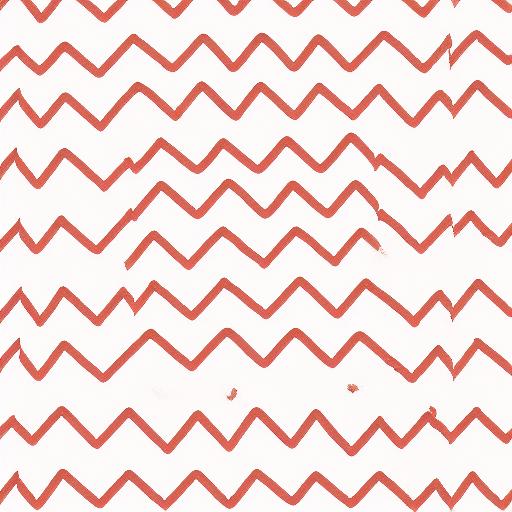} &
        \figframe[width=0.166\textwidth, height=0.166\textwidth]{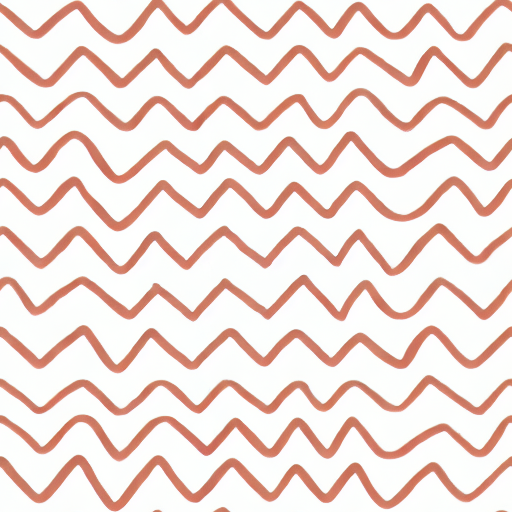} \\

        \vspace{-0.5mm}\hspace{-1mm}
        \figframe[width=0.166\textwidth, height=0.166\textwidth]{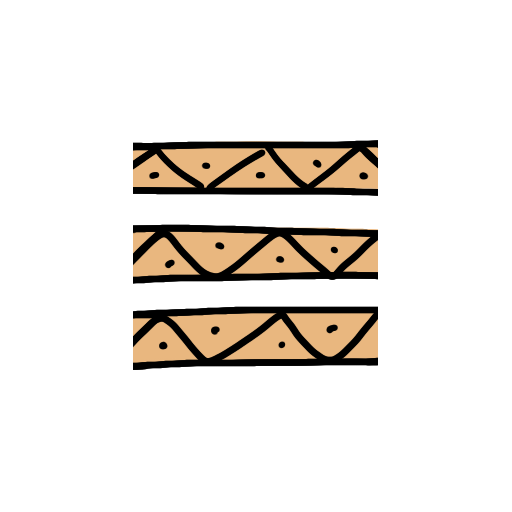} &
        \figframe[width=0.166\textwidth, height=0.166\textwidth]{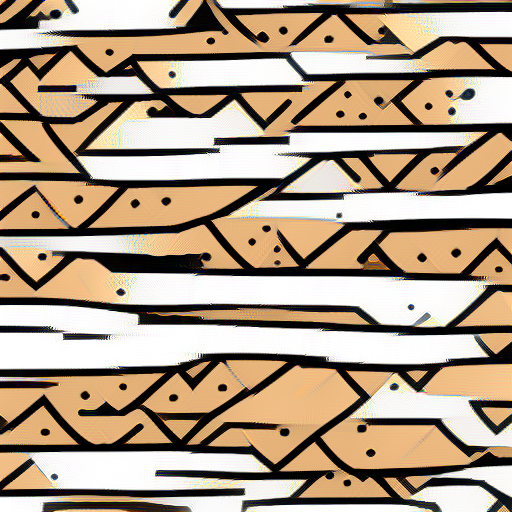} &
        \figframe[width=0.166\textwidth, height=0.166\textwidth]{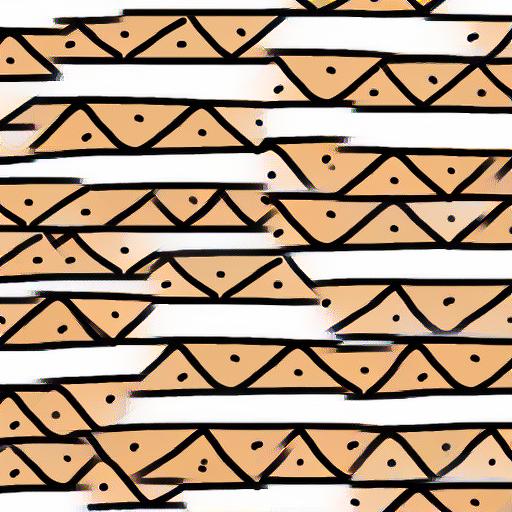} & 
        \figframe[width=0.166\textwidth, height=0.166\textwidth]{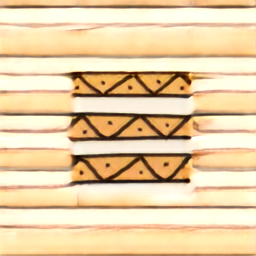} &
        \figframe[width=0.166\textwidth, height=0.166\textwidth]{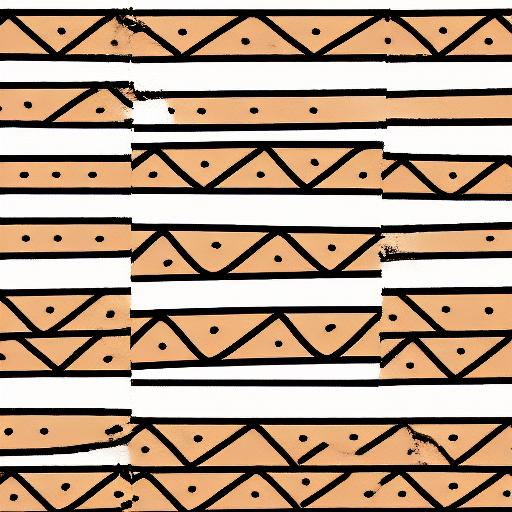} &
        \figframe[width=0.166\textwidth, height=0.166\textwidth]{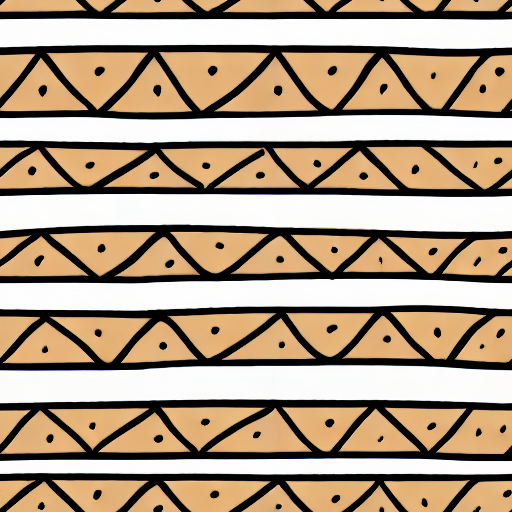} \\

        \vspace{-0.5mm}\hspace{-1mm}
        \figframe[width=0.166\textwidth, height=0.166\textwidth]{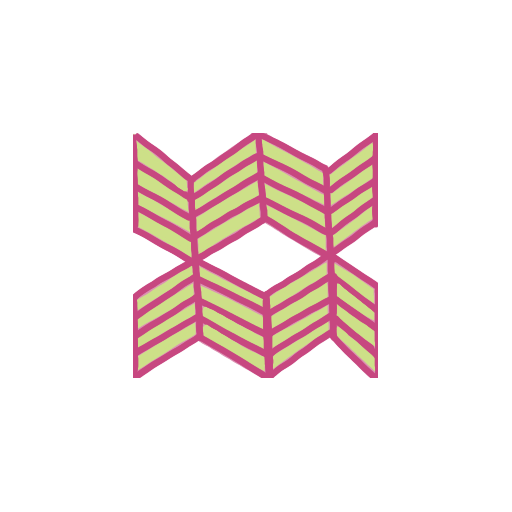} &
        \figframe[width=0.166\textwidth, height=0.166\textwidth]{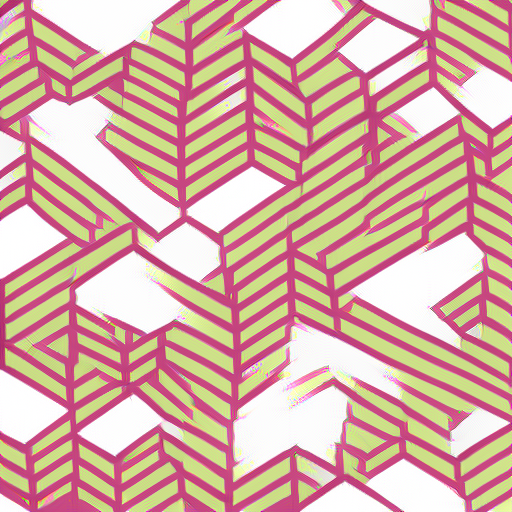} &
        \figframe[width=0.166\textwidth, height=0.166\textwidth]{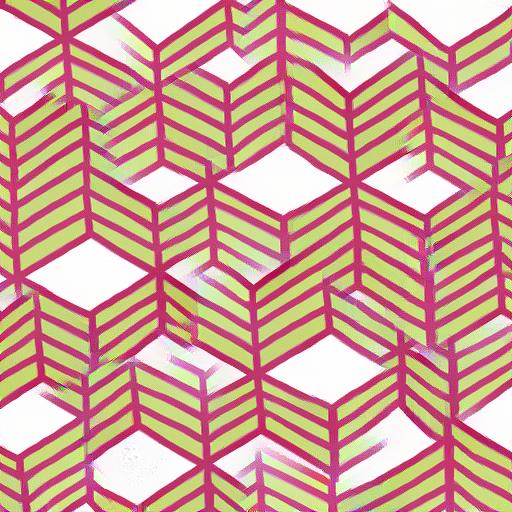} & 
        \figframe[width=0.166\textwidth, height=0.166\textwidth]{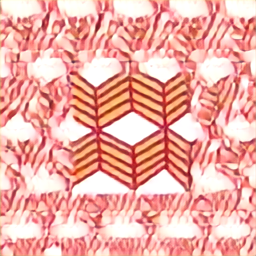} &
        \figframe[width=0.166\textwidth, height=0.166\textwidth]{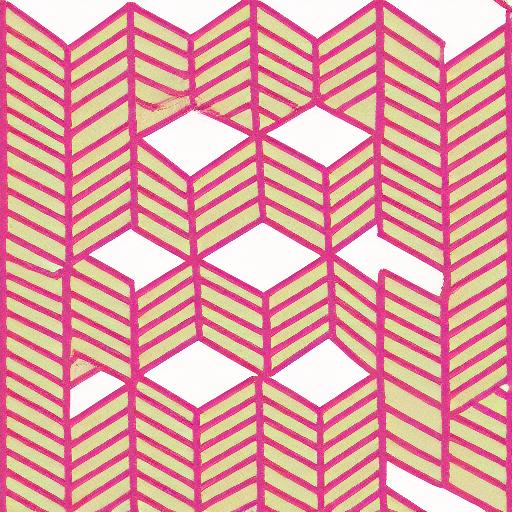} &
        \figframe[width=0.166\textwidth, height=0.166\textwidth]{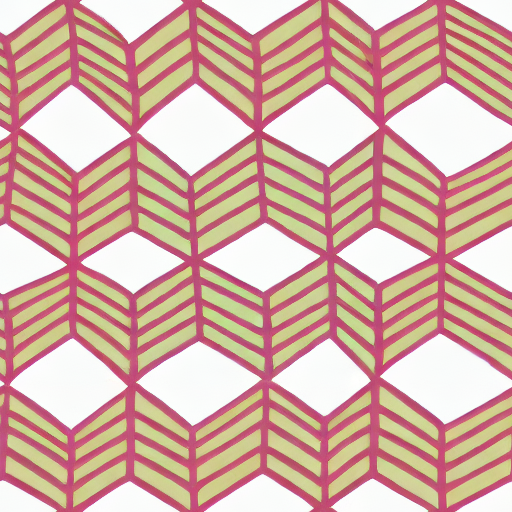} \\

        \vspace{-0.5mm}\hspace{-1mm}
        \figframe[width=0.166\textwidth, height=0.166\textwidth]{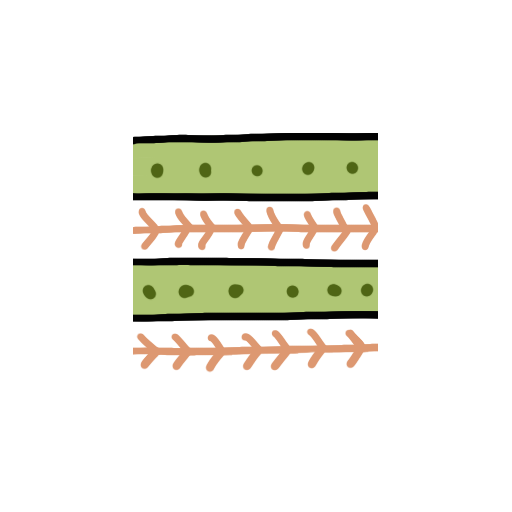} &
        \figframe[width=0.166\textwidth, height=0.166\textwidth]{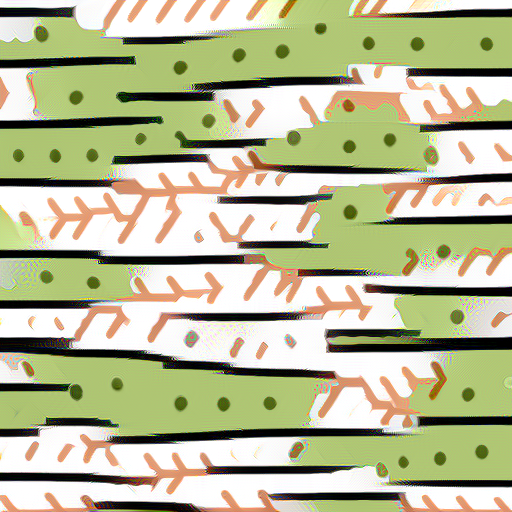} &
        \figframe[width=0.166\textwidth, height=0.166\textwidth]{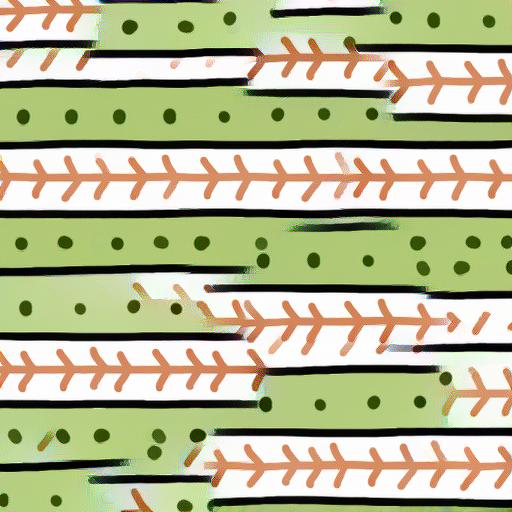} & 
        \figframe[width=0.166\textwidth, height=0.166\textwidth]{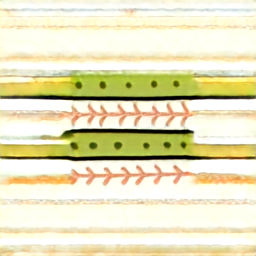} &
        \figframe[width=0.166\textwidth, height=0.166\textwidth]{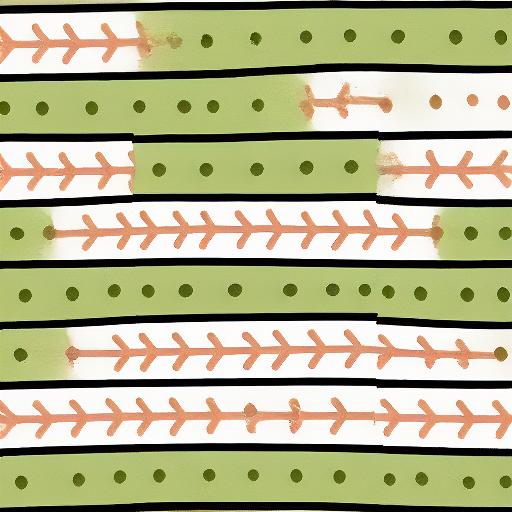} &
        \figframe[width=0.166\textwidth, height=0.166\textwidth]{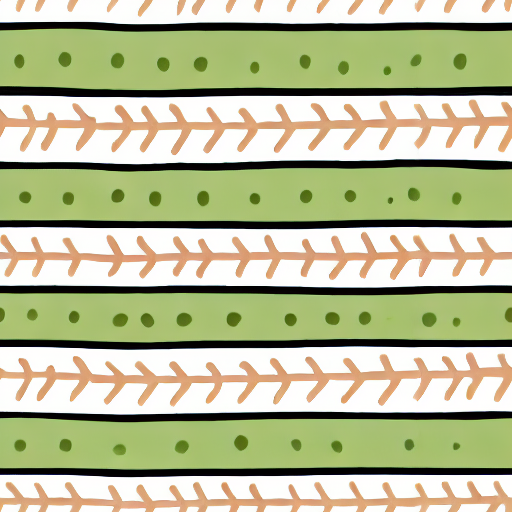} \\
    \end{tabular}
    \caption{\textbf{Comparisons.} We compared our method with established texture and material synthesis techniques. As previous work tends to break the overall structure (a, b, d) or fails at reconstructing the pattern appearance (c), our method consistently expands the input by preserving structural integrity and input coherency.}
    \label{fig:comparison}
\end{figure*}

\begin{figure}
    \centering
    \setlength{\tabcolsep}{.5pt}
    \begin{tabular}{ccc}
        \hspace{-1mm}\small{Init} & \small{\citet{zhou2018}} & \small{Our Method}\\
        \vspace{-0.75mm}\hspace{-1mm}
        \figframe[width=0.23\linewidth]{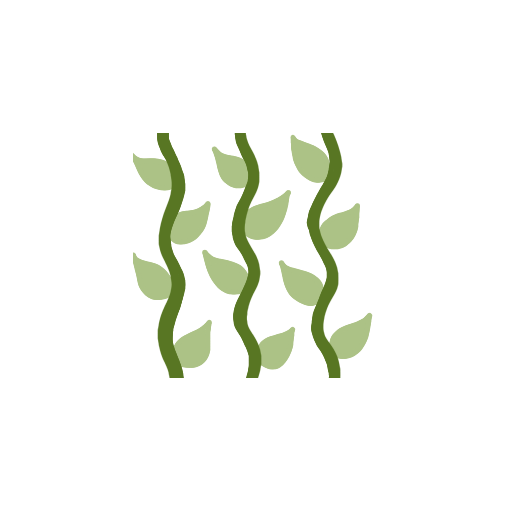} &
        \figframe[width=0.23\linewidth]{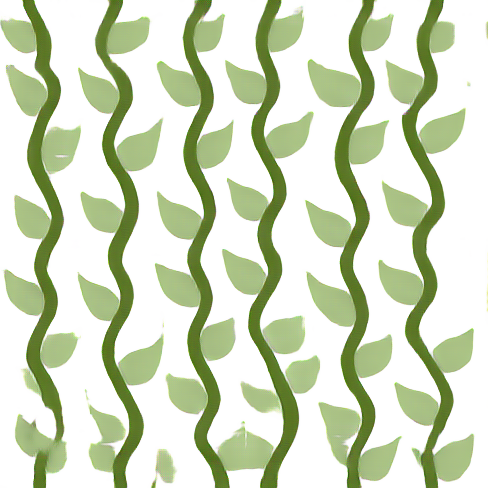} &
        \figframe[width=0.23\linewidth]{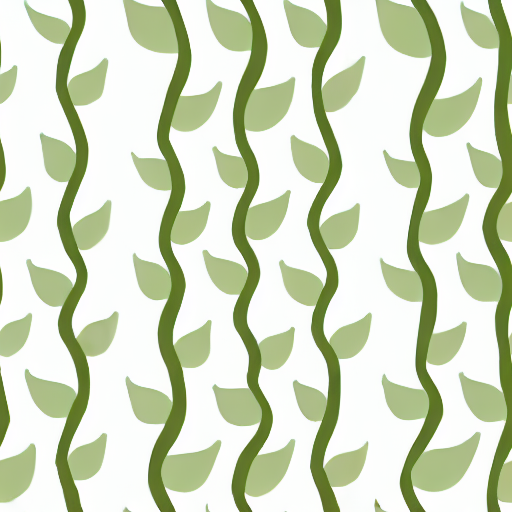} \\

        \vspace{-0.75mm}\hspace{-1mm}
        \figframe[width=0.23\linewidth]{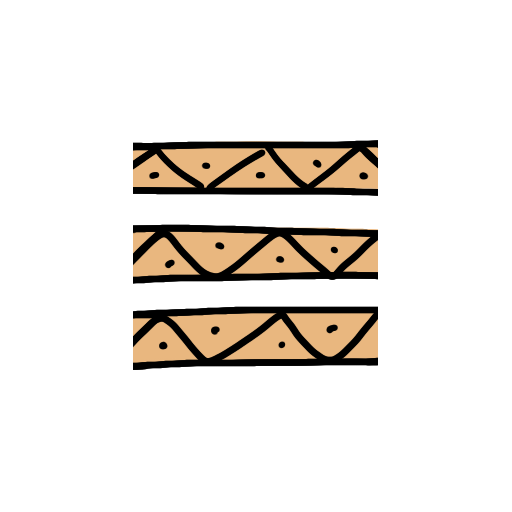} &
        \figframe[width=0.23\linewidth]{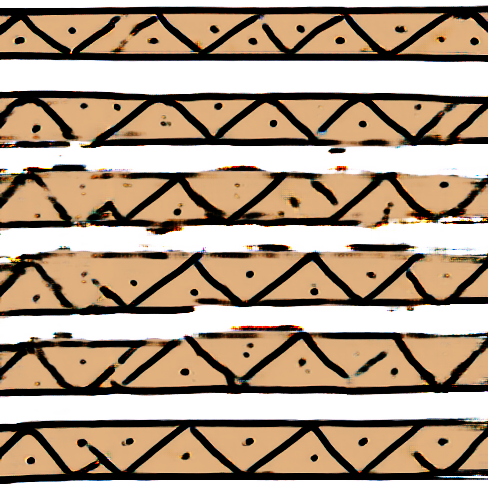} &
        \figframe[width=0.23\linewidth]{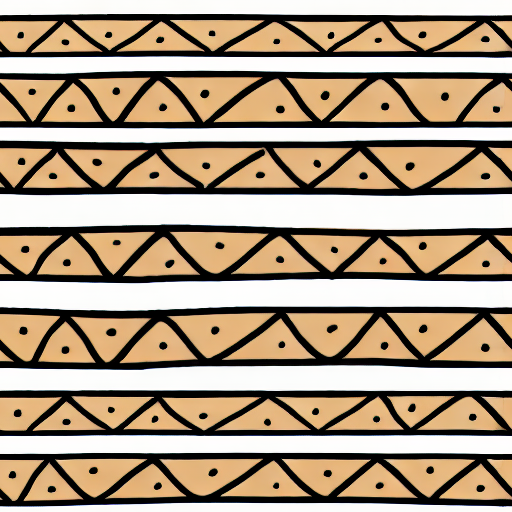} \\
    \end{tabular}
    \caption{\textbf{Comparison to \citet{zhou2018}.} We compared our diffusion-based pipeline for pattern expansion to the GAN-based extension strategy proposed by \citet{zhou2018}, which requires training a network for each input sketch. Our method better captures the overall pattern structure and at the same time shows less visual artifacts, which are frequent in GAN-based approaches.}
    \label{fig:non_stationary}
\end{figure}

\paragraph{Comparison}
We compare our approach against several state-of-the-art methods: \citet{heitz2021sliced}, GCD~\cite{zhou2023neural}, MatFuse~\cite{vecchio2023matfuse}, and \citet{zhou2024generating}. For each method, we use the official code and weights released by the authors and adapt our input to match the ones required by each method. As MatFuse~\cite{vecchio2023matfuse} is trained to generate PBR materials, we provide the pattern as the diffuse component of the material, initializing the other properties to the default values.

As shown in Fig.~\ref{fig:comparison}, our method significantly improves on previous approaches in preserving the structural integrity and visual fidelity of patterns. While methods like \citet{heitz2021sliced} and GCD capture the visual features of the patterns, they tend to break the overall structure, introducing unnatural distortions resulting quality degradation. MatFuse fails to capture the appearance of the pattern, mostly due to the training on natural textures, being only able to reproduce the colors and general shape of the pattern but lacking any fine details. \citet{zhou2024generating}, in contrast, is generally able to reproduce the sharp visual appearance of the pattern, and capture the main features; however, due to its main focus on non-stationary textures, it tends to break the overall structure, resulting in sharp discontinuity edges inside the image and transitioning between different parts of the pattern. Additionally, it struggles with very sparse patterns (e.g., third row in Fig.~\ref{fig:comparison}), and introduces a color shift on the original input. None of the other methods produces tileable results.
Compared to the other approaches, our work is able to capture the visual features of the pattern and extend it seamlessly, introducing slight variations without altering the overall structure. Moreover, all of our expansion results are tileable, thanks to the use of noise rolling at inference time.

It is important to highlight that our pipeline completes the diffusion step in around 2 seconds, whereas the execution times for \citet{heitz2021sliced}, GCD~\cite{zhou2023neural}, and \citet{zhou2024generating} range from at least 2 minutes up to 40 minutes for each example. MatFuse~\cite{vecchio2023matfuse} has similar timings to our method, due to the similar diffusion backbone, but shows significantly worse generation quality.

For the sake of completeness, we also compared our method to \citet{zhou2018}, which aims to double the spatial size of a texture by leveraging a GAN specifically trained to reconstruct a $2k$ x $2k$ texture from a $k$ x $k$ patch. We trained a GAN model for each of the input patterns shown in Fig.\ref{fig:non_stationary}, each of which required approximately 1.5 hours. As shown, \citet{zhou2018} is able to reproduce the overall pattern structure while spatially extending the input sketch, but it also introduces several artifacts and discontinuities that degrade the overall quality output. In contrast, our approach achieved a better result in terms of both structure preservation and image quality, without requiring ad-hoc training for each texture and preserving the input sketch.
\begin{figure}
    \centering
    \setlength{\tabcolsep}{.5pt}
    \begin{tabular}{cccc}
        \hspace{-1mm}\small{Init} & \small{No prompt} & \small{General prompt} & \small{Adhoc prompt}\\

        \vspace{-0.75mm}\hspace{-1mm}
        \figframe[width=0.23\linewidth, height=0.23\linewidth]{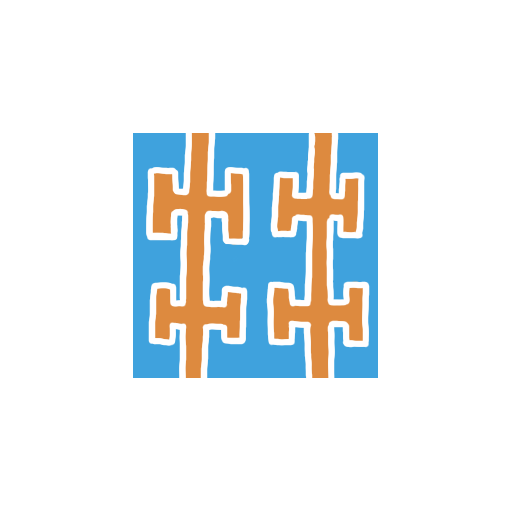} &
        \figframe[width=0.23\linewidth, height=0.23\linewidth]{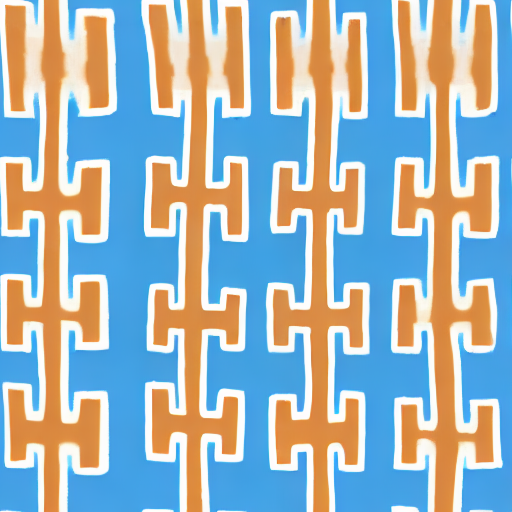} &
        \figframe[width=0.23\linewidth, height=0.23\linewidth]{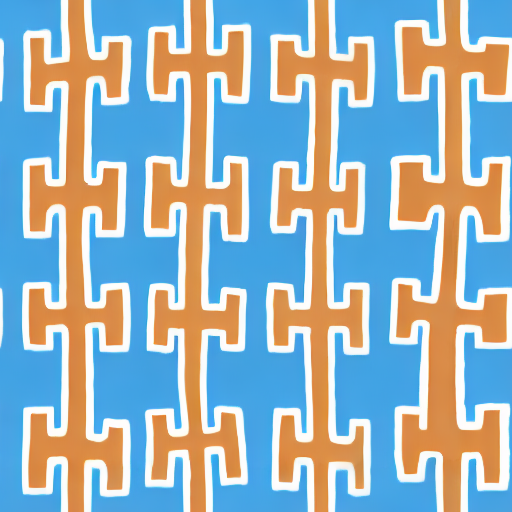} &
        \figframe[width=0.23\linewidth, height=0.23\linewidth]{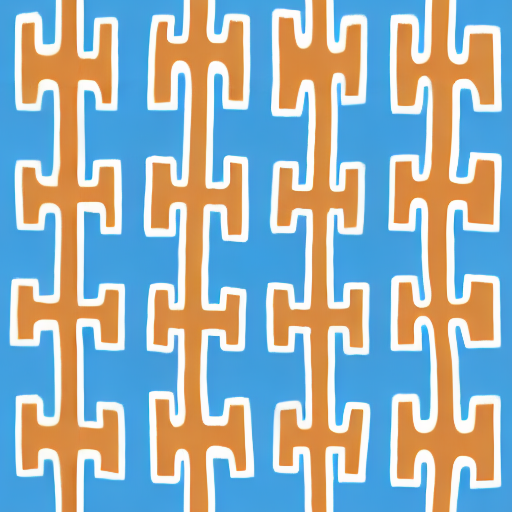} \\
        \vspace{-0.75mm}\hspace{-1mm}
        \figframe[width=0.23\linewidth, height=0.23\linewidth]{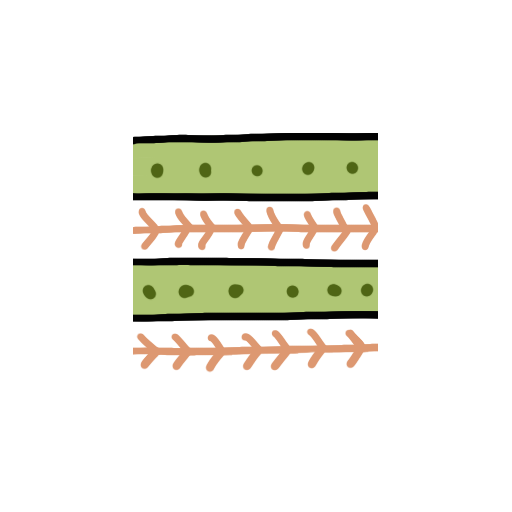} &
        \figframe[width=0.23\linewidth, height=0.23\linewidth]{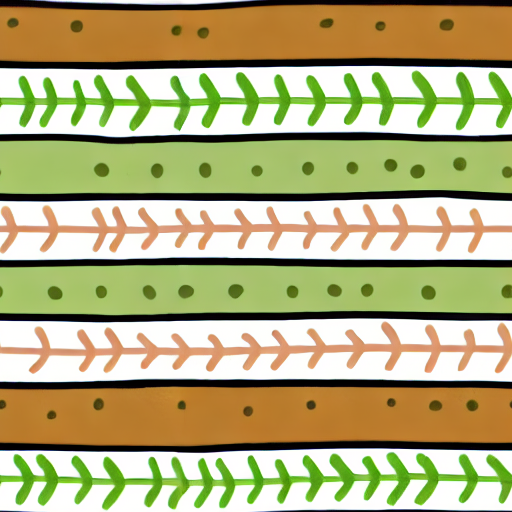} &
        \figframe[width=0.23\linewidth, height=0.23\linewidth]{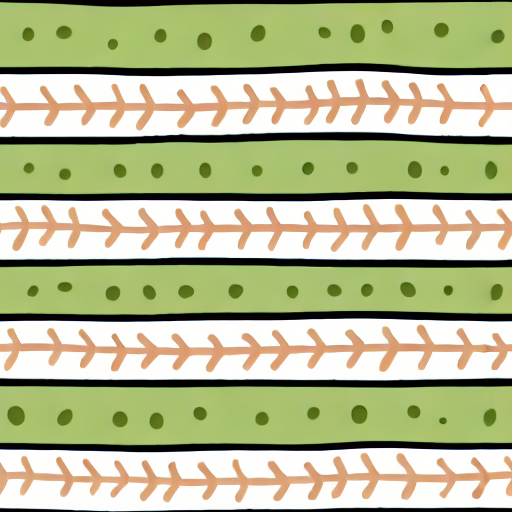} &
        \figframe[width=0.23\linewidth, height=0.23\linewidth]{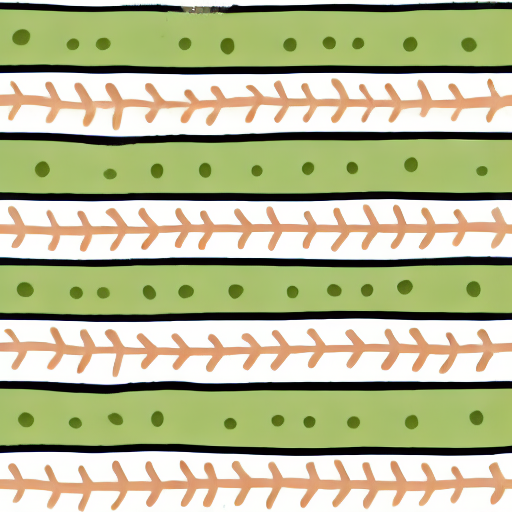} \\
        \vspace{-0.75mm}\hspace{-1mm}
        \figframe[width=0.23\linewidth, height=0.23\linewidth]{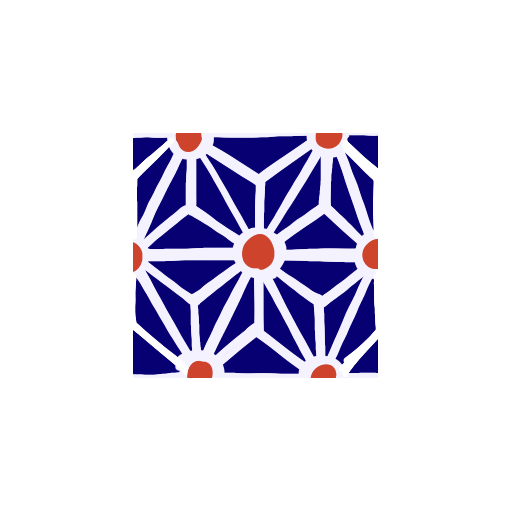} &
        \figframe[width=0.23\linewidth, height=0.23\linewidth]{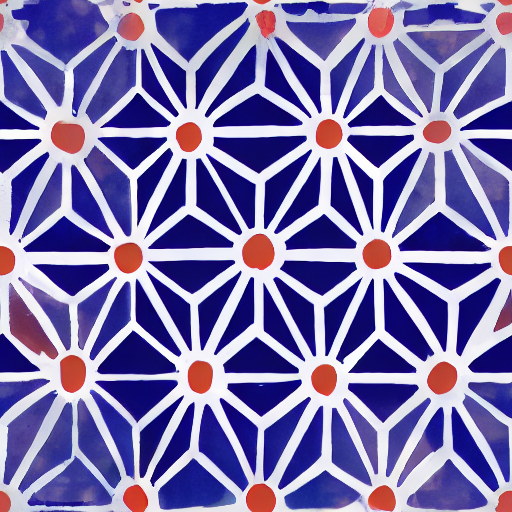} &
        \figframe[width=0.23\linewidth, height=0.23\linewidth]{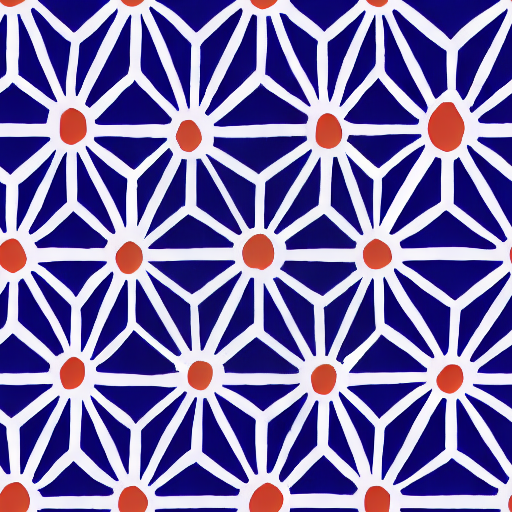} &
        \figframe[width=0.23\linewidth, height=0.23\linewidth]{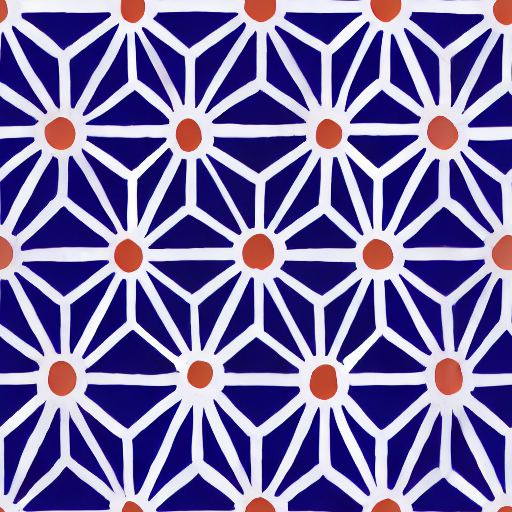} \\

    \end{tabular}
    \caption{\textbf{Text prompt ablation.} Instead of requiring the text prompt to be tailored to a particular pattern sample (c), our general-purpose text prompt (b) enhances the overall generation quality by introducing structural and coherency features that are less respected when employing image prompting alone (a).}
    \label{fig:ablation_text}
\end{figure}

\subsection{User Study}
To evaluate our method's performance, we conducted a comparative study with \citet{heitz2021sliced}, GCD~\cite{zhou2023neural}, MatFuse~\cite{vecchio2023matfuse}, and \citet{zhou2024generating}. This involved 80 MS/PhD students in computer science who were tasked with selecting their preferred expanded pattern based on the quality and consistency of the generation. We showed each one of them 20 randomly chosen pattern generations--including both the input crop and the output for each method compared, in a random sequence--from a set of 35 expanded patterns. %
As shown in Fig.\ref{fig:user-study}, our approach received a higher number of votes (\textbf{Ours}=\textbf{1471 (i.e.: 91.93\%)}, \citet{heitz2021sliced}=0, GCD=7 (i.e.: 0.44\%), MatFuse=0, \citet{zhou2024generating}=122 (i.e.: 7.63\%)), showing a significant general preference of the expansions generated from our approach compared to the other methods. These results further support our claims.

\subsection{Quantitative Evaluation}
\begin{figure}[t]
    \centering
    \setlength{\tabcolsep}{.5pt}
    \begin{tabular}{ccc}
        & & \small{Tiled Expansion}\\

        \hspace{-2mm} \raisebox{1.6\height}{    
            \begin{minipage}{0.23\linewidth}
                \centering
                \hspace{-4mm} \small{Init} \\   
                \figframe[width=1\linewidth, height=1\linewidth]{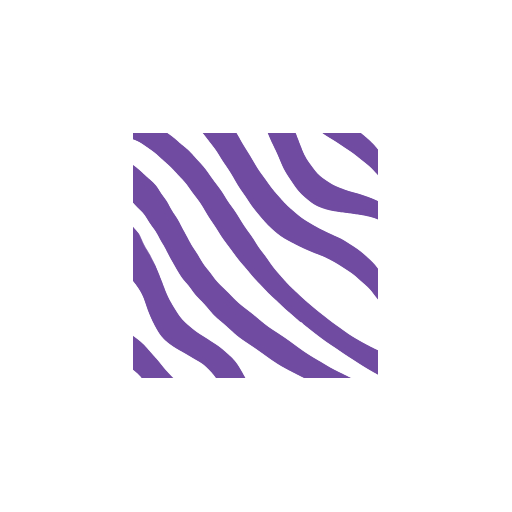}
            \end{minipage}
        } & 
        \hspace{-2mm} \raisebox{1.6\height}{
            
            \begin{minipage}{0.23\linewidth}
                \centering
                \hspace{-4mm} \small{Expansion} \\ 
                \figframe[width=1\linewidth, height=1\linewidth]{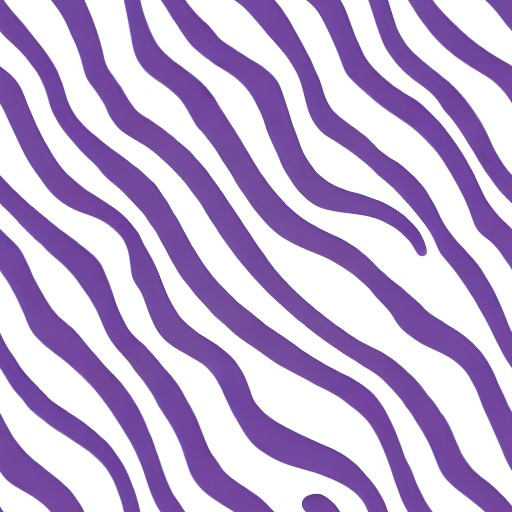}
            \end{minipage}
        } & 
        \hspace{-2.5mm} \figframe[width=0.46\linewidth, height=0.46\linewidth]{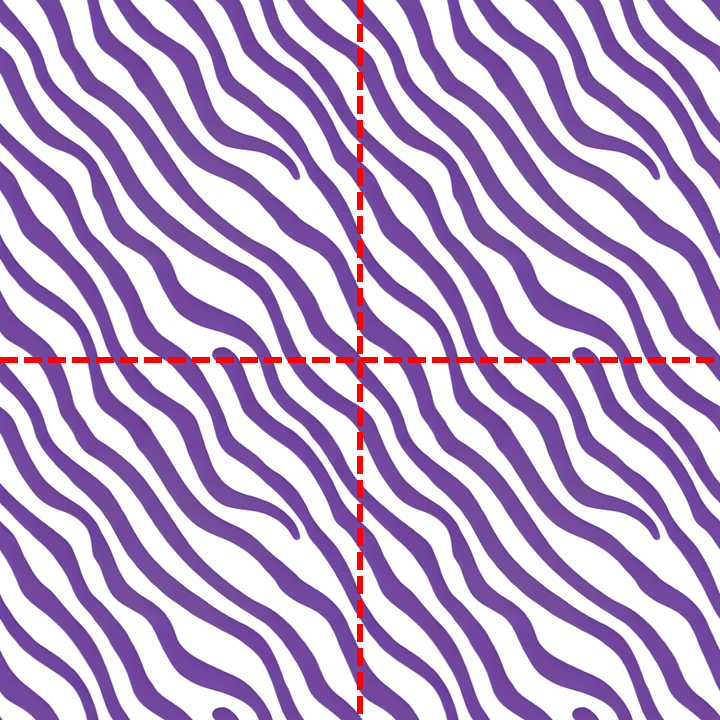}\\

    \end{tabular}
    \caption{\textbf{Tileability metric.} Although reaching $60.7\%$ score according to the TexTile metric \cite{rodriguez2024textile}, the expanded zebra pattern (middle) shows no seam while being replicated along vertical and horizontal directions (right), with concatenation edges being highlighted by the red dashed lines. }
    \label{fig:textile}
\end{figure}

\begin{figure*}
    \centering
    \setlength{\tabcolsep}{.5pt}

    \begin{tabular}{ccc}
        \vspace{-1mm}\hspace{-1mm}
        \figframe[height=0.24\textheight]{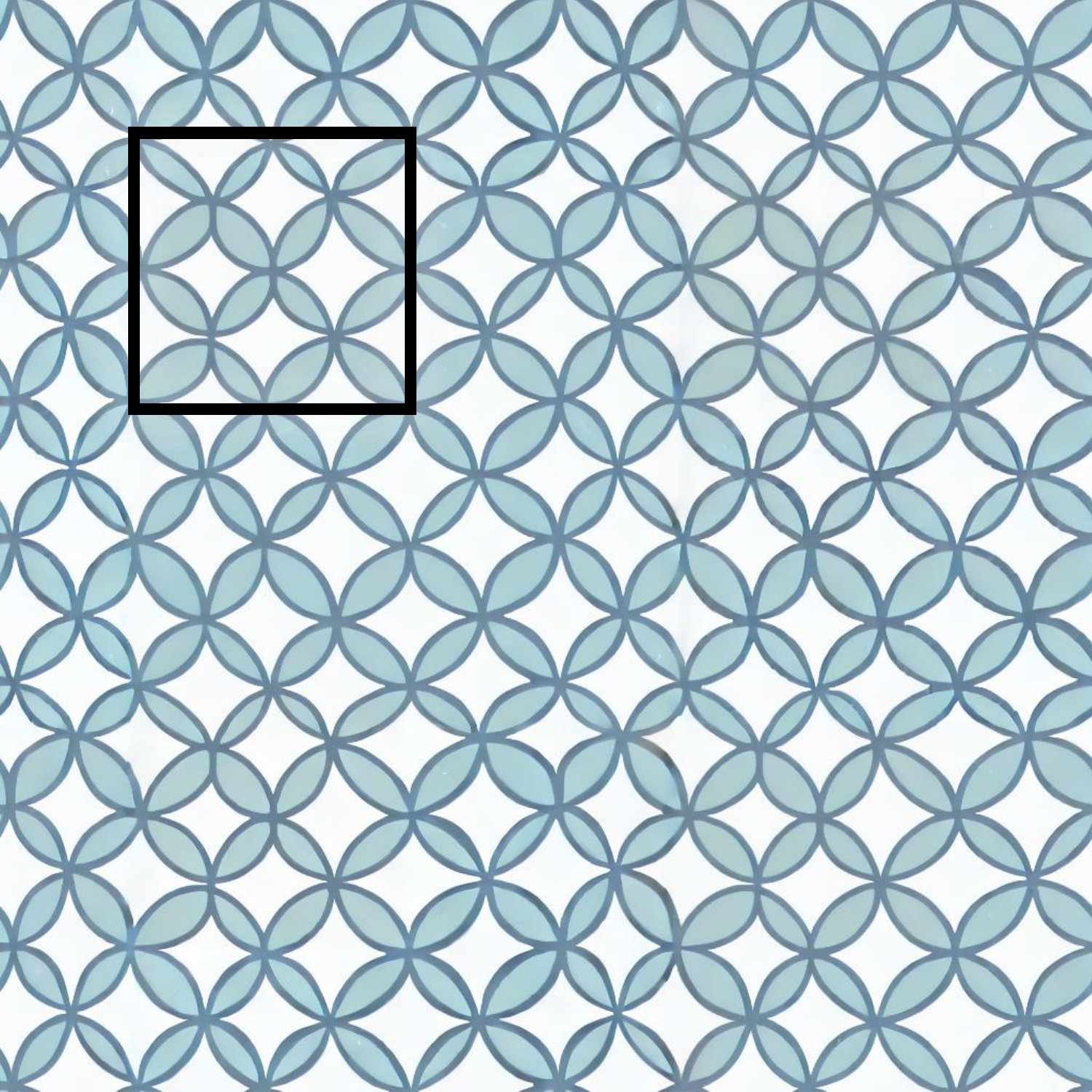} &
        \figframe[height=0.24\textheight]{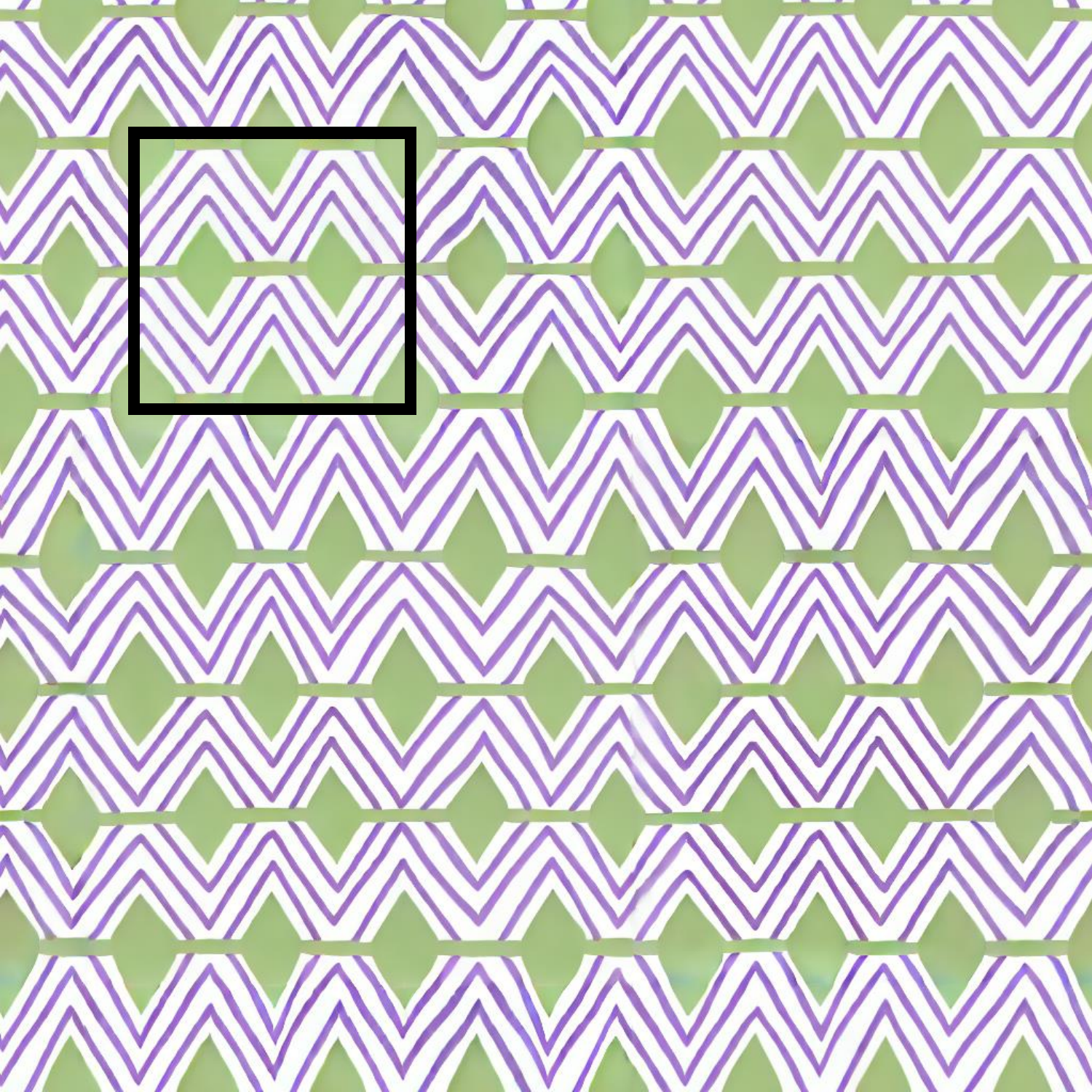 } &
        \figframe[height=0.24\textheight]{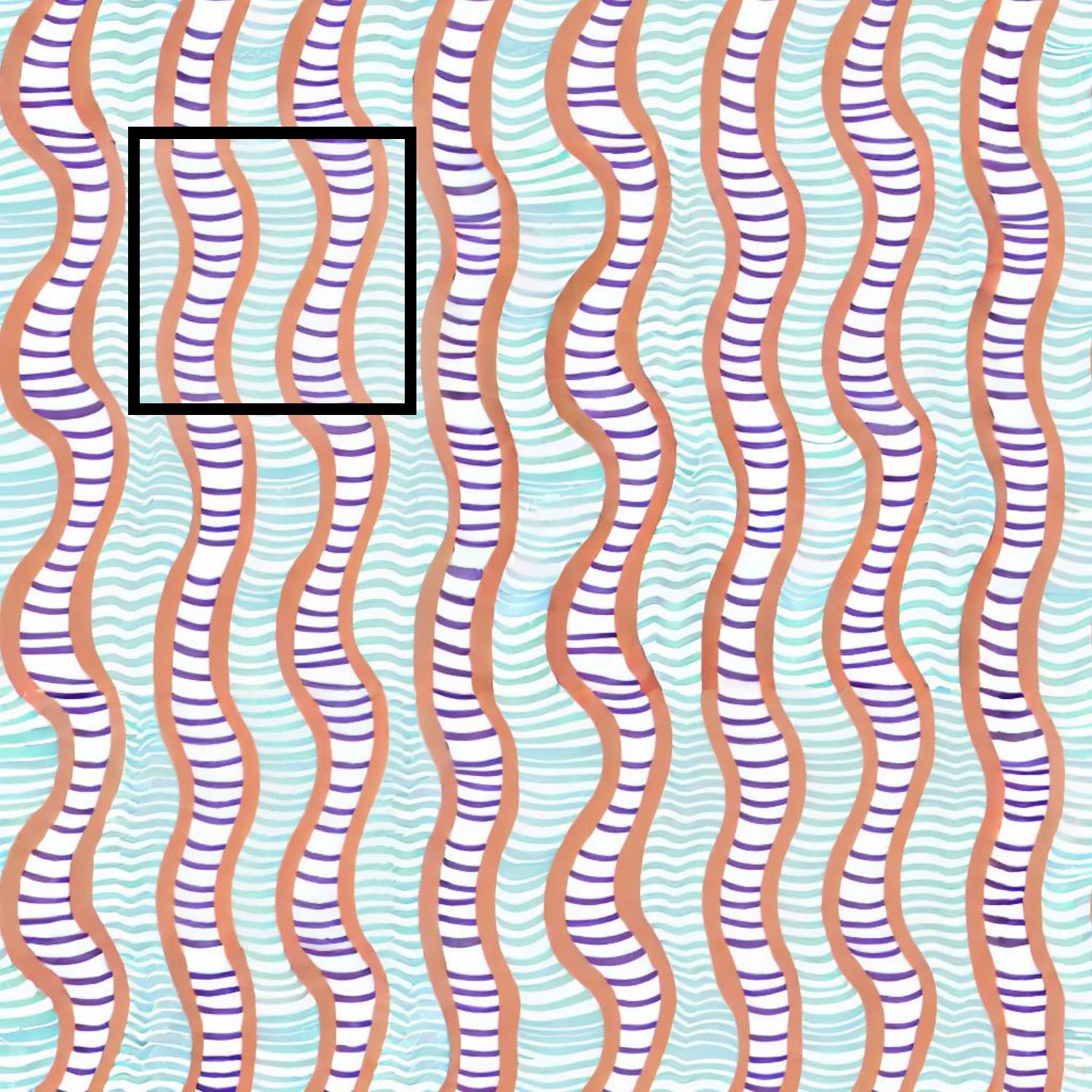}\\

        \vspace{-0.5mm}
        \figframe[height=0.24\textheight]{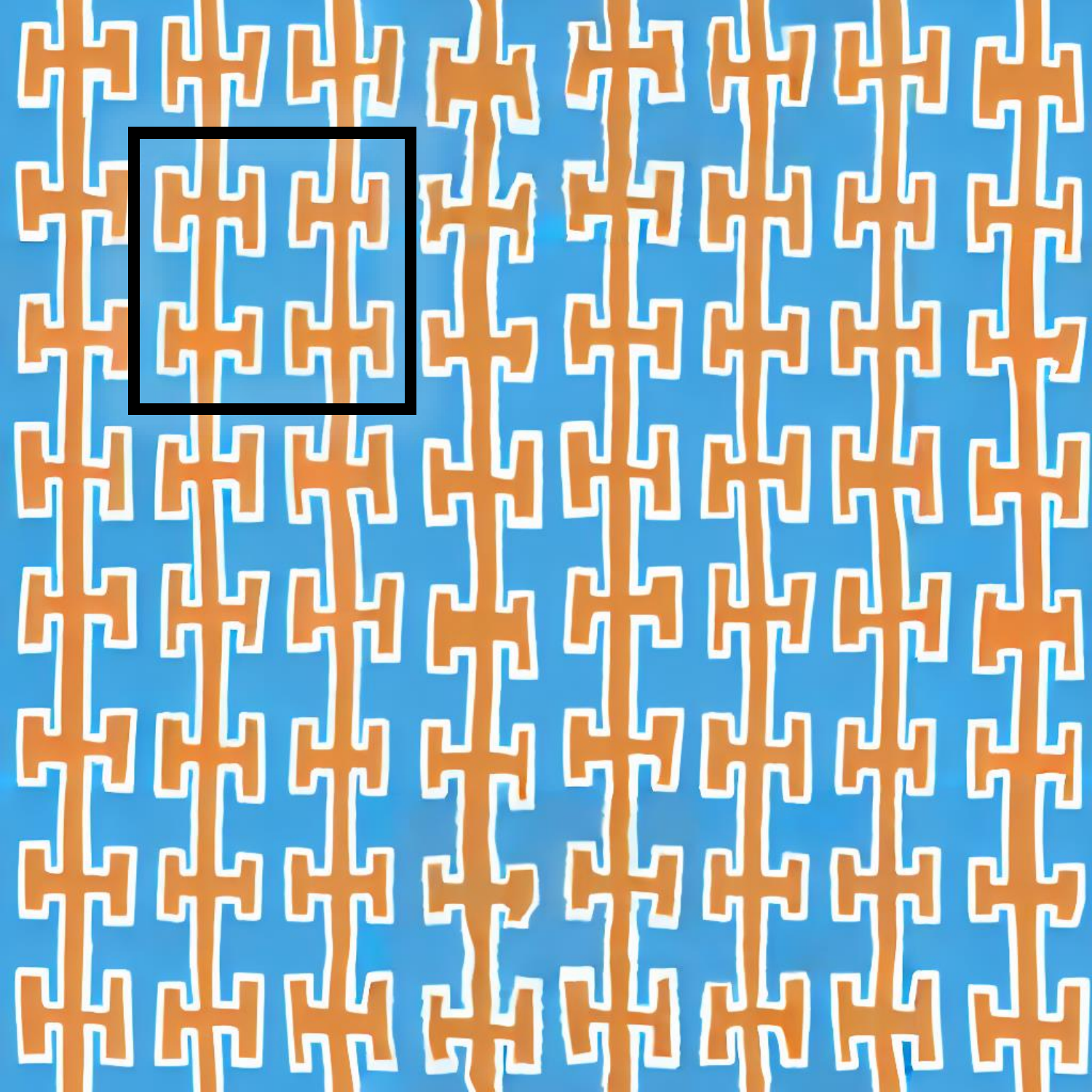} &
        \figframe[height=0.24\textheight]{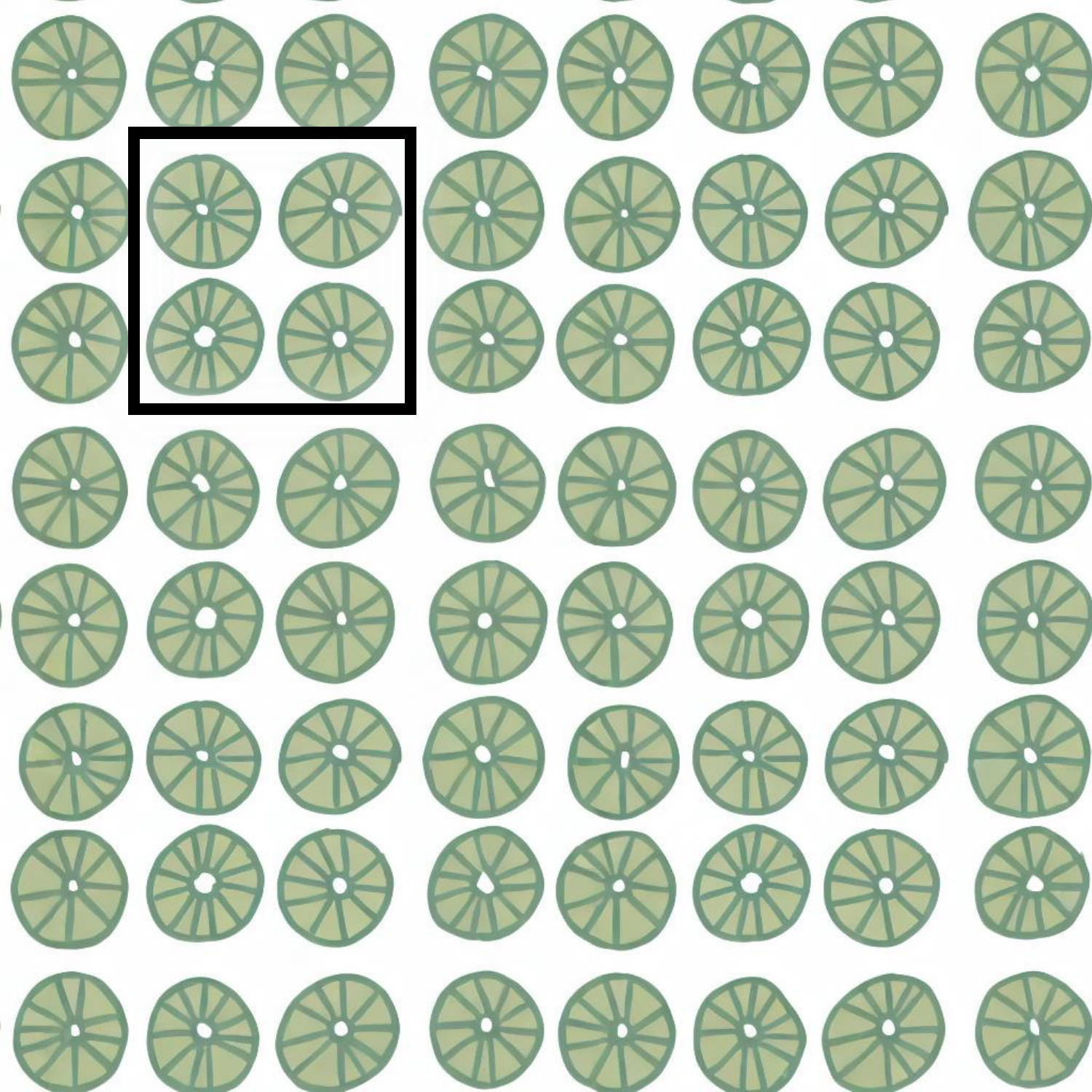} &
        \figframe[height=0.24\textheight]{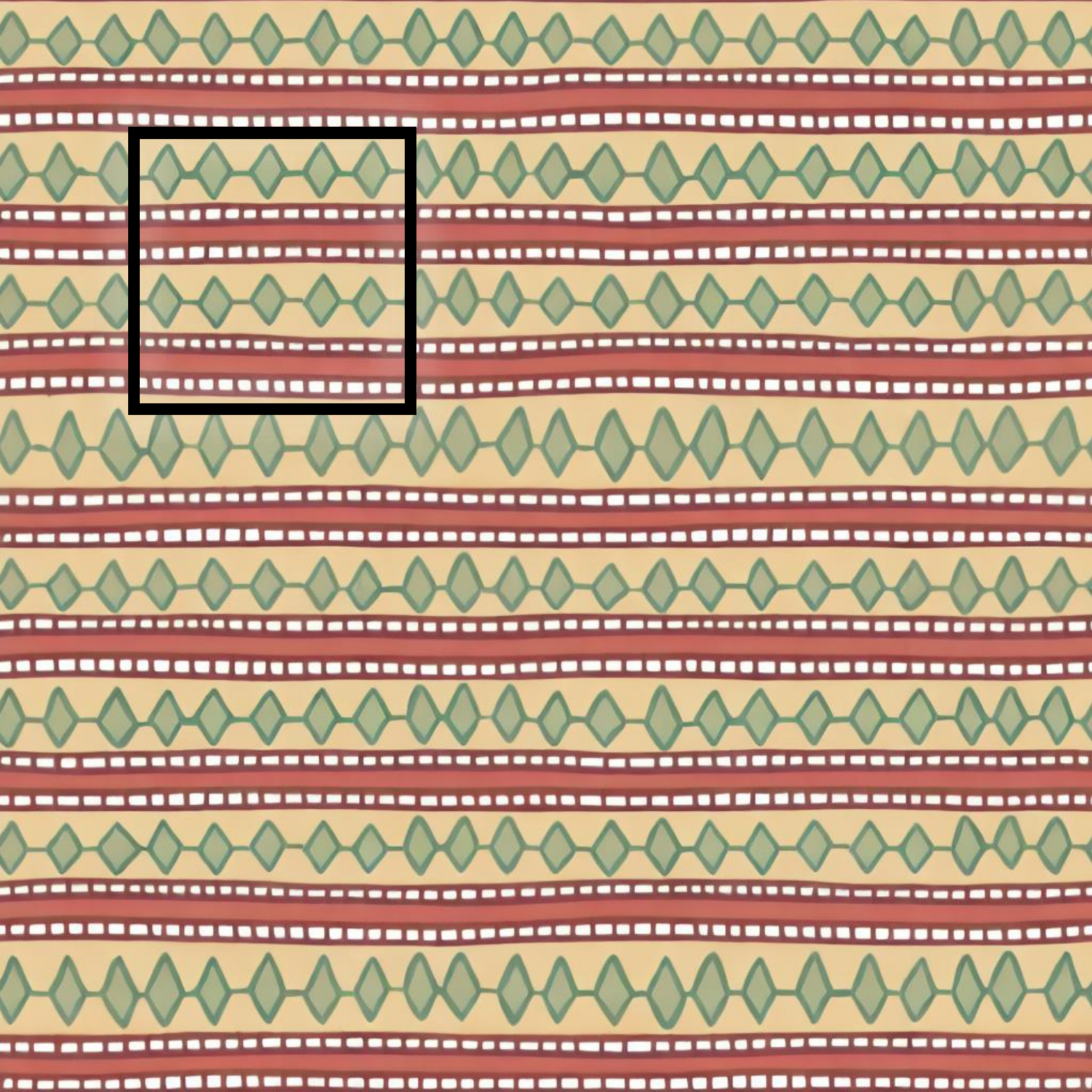}\\
        
        \vspace{-0.5mm}
        \figframe[height=0.24\textheight]{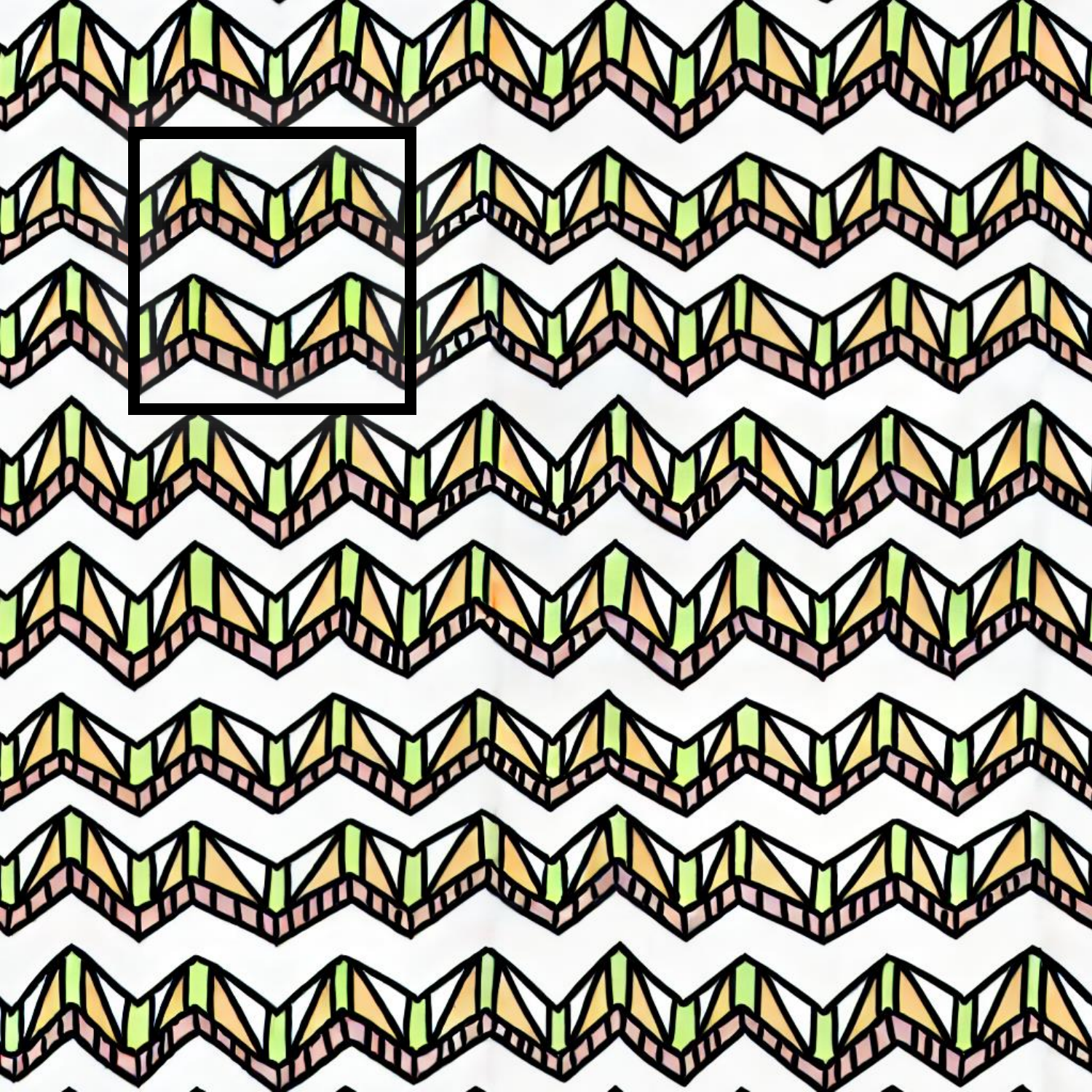} &
        \figframe[height=0.24\textheight]{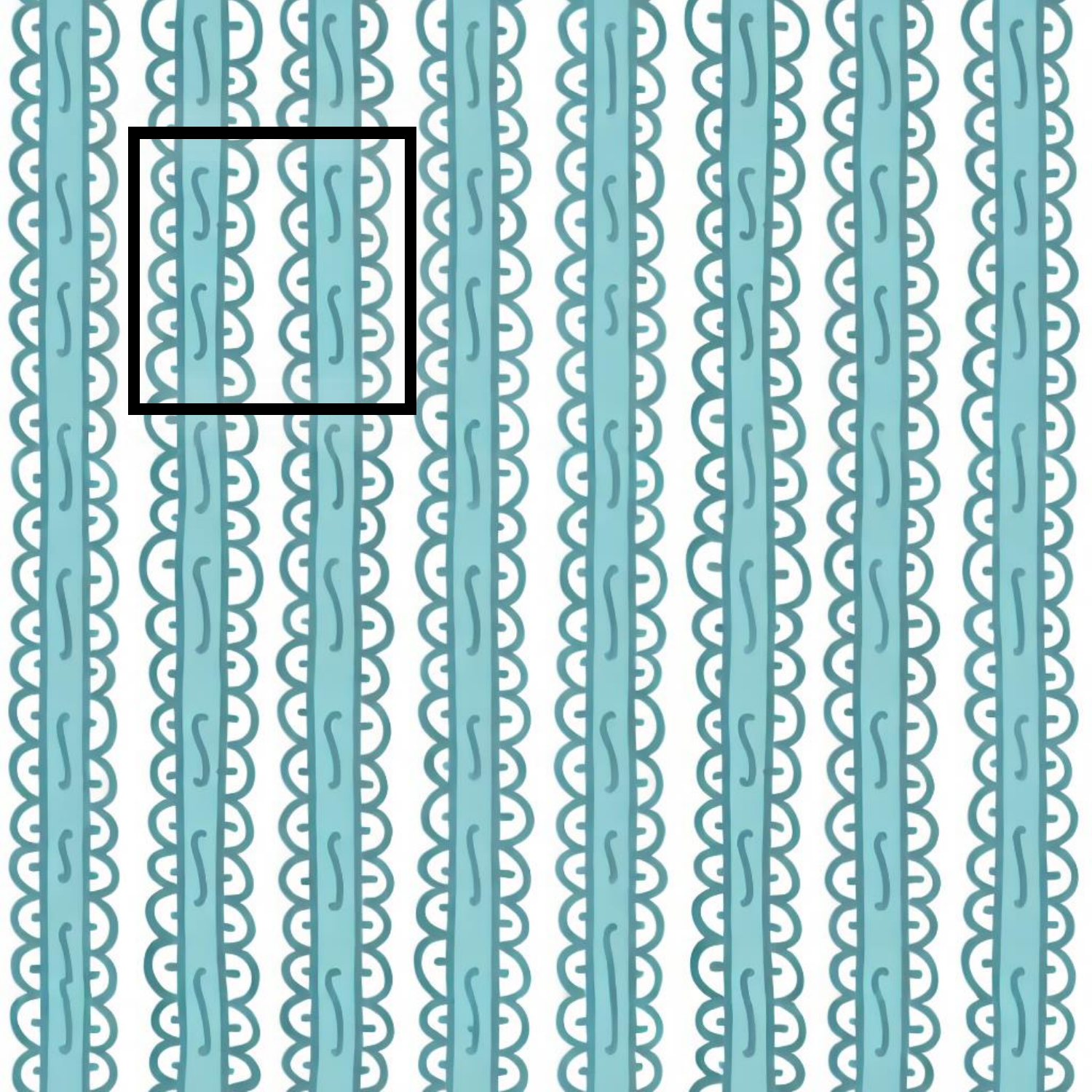} &
        \figframe[height=0.24\textheight]{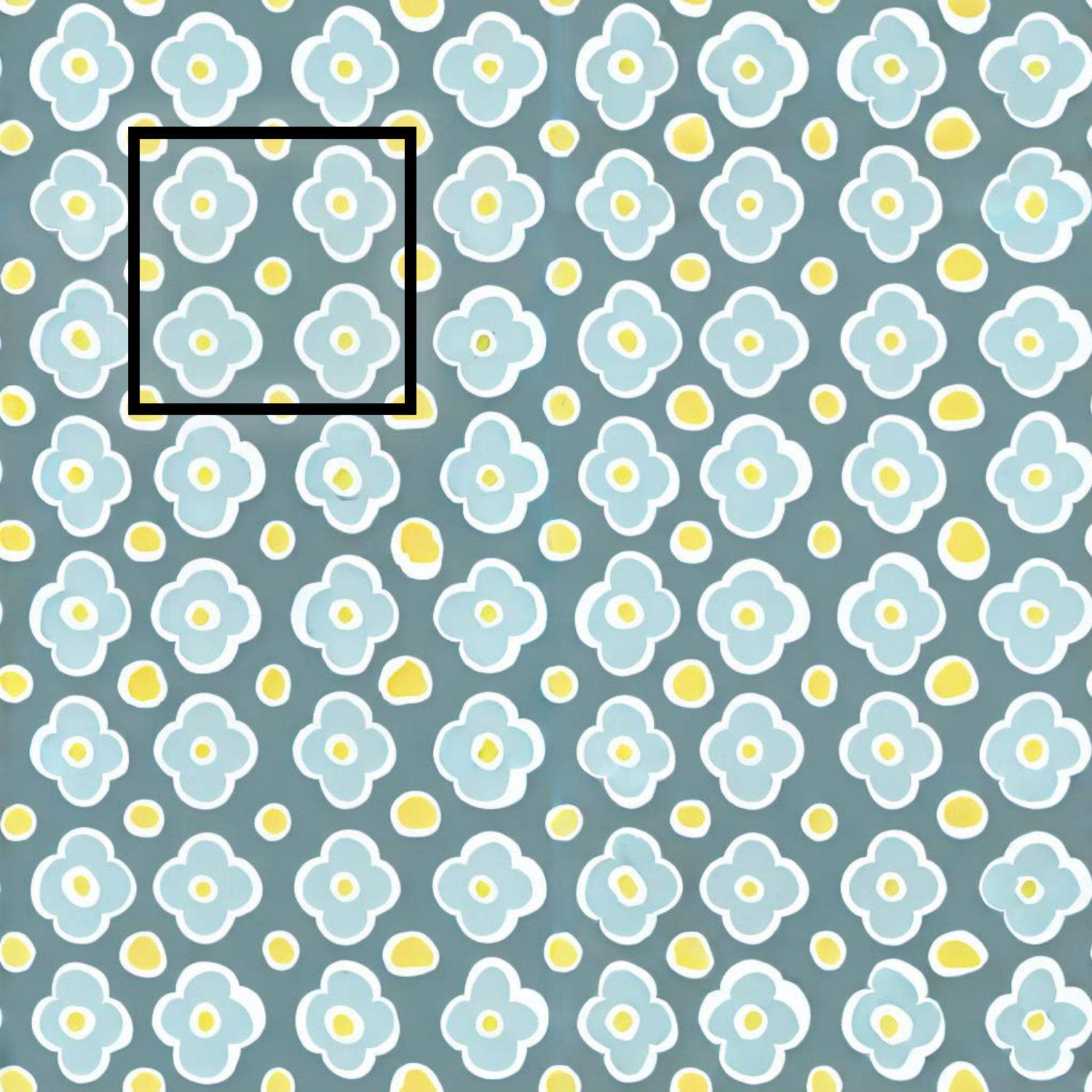}\\
        
        \vspace{-0.5mm}
        \figframe[height=0.24\textheight]{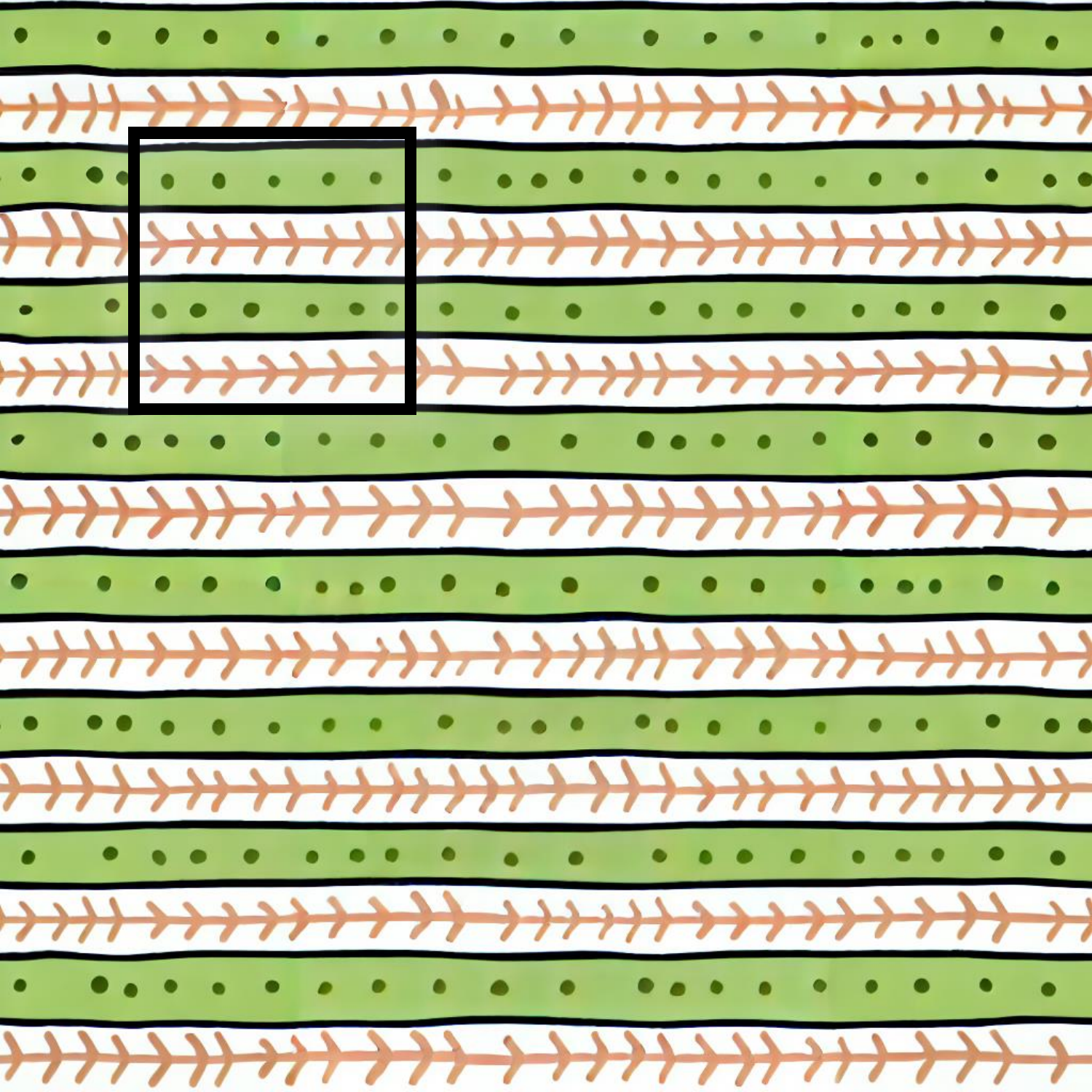} &
        \figframe[height=0.24\textheight]{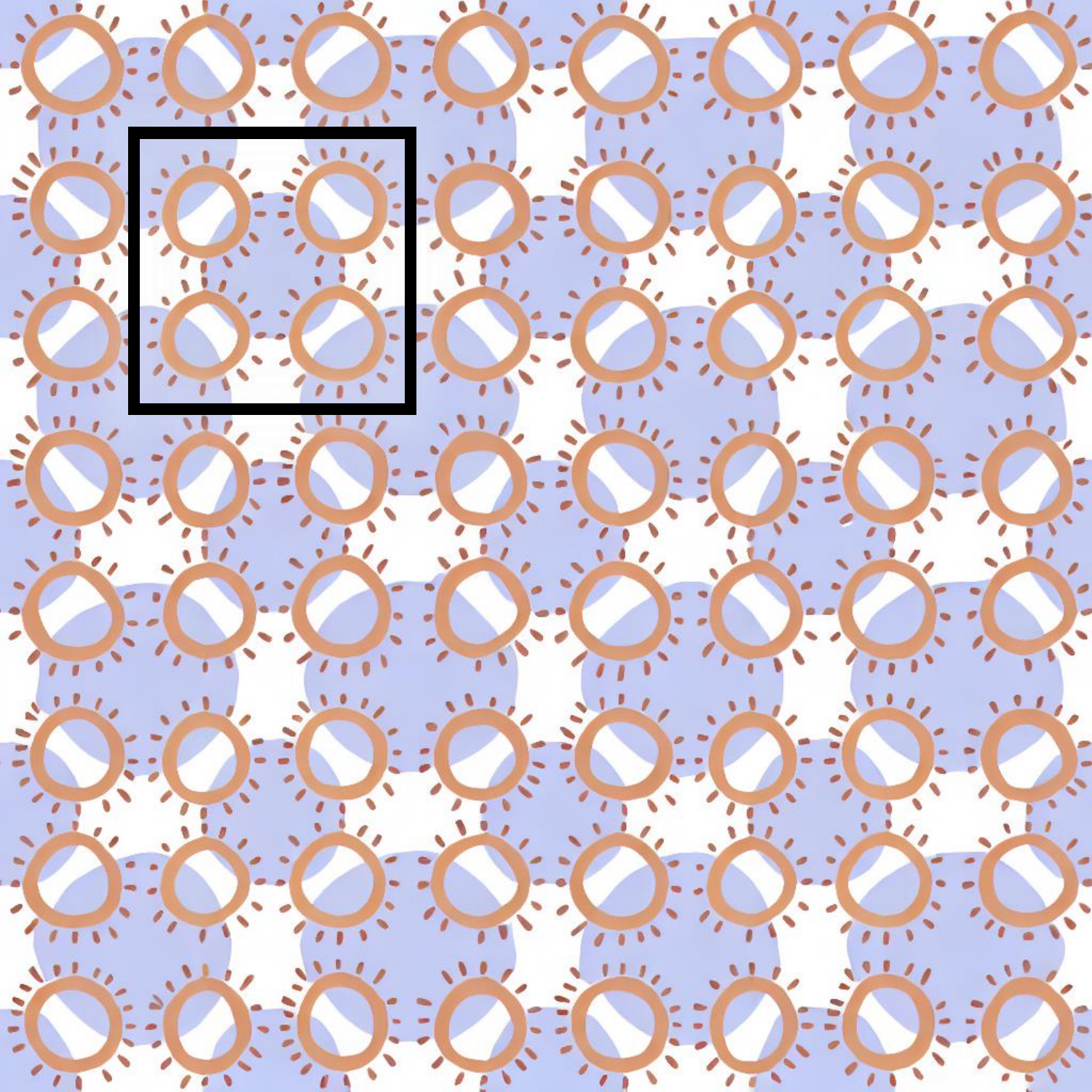} &
        \figframe[height=0.24\textheight]{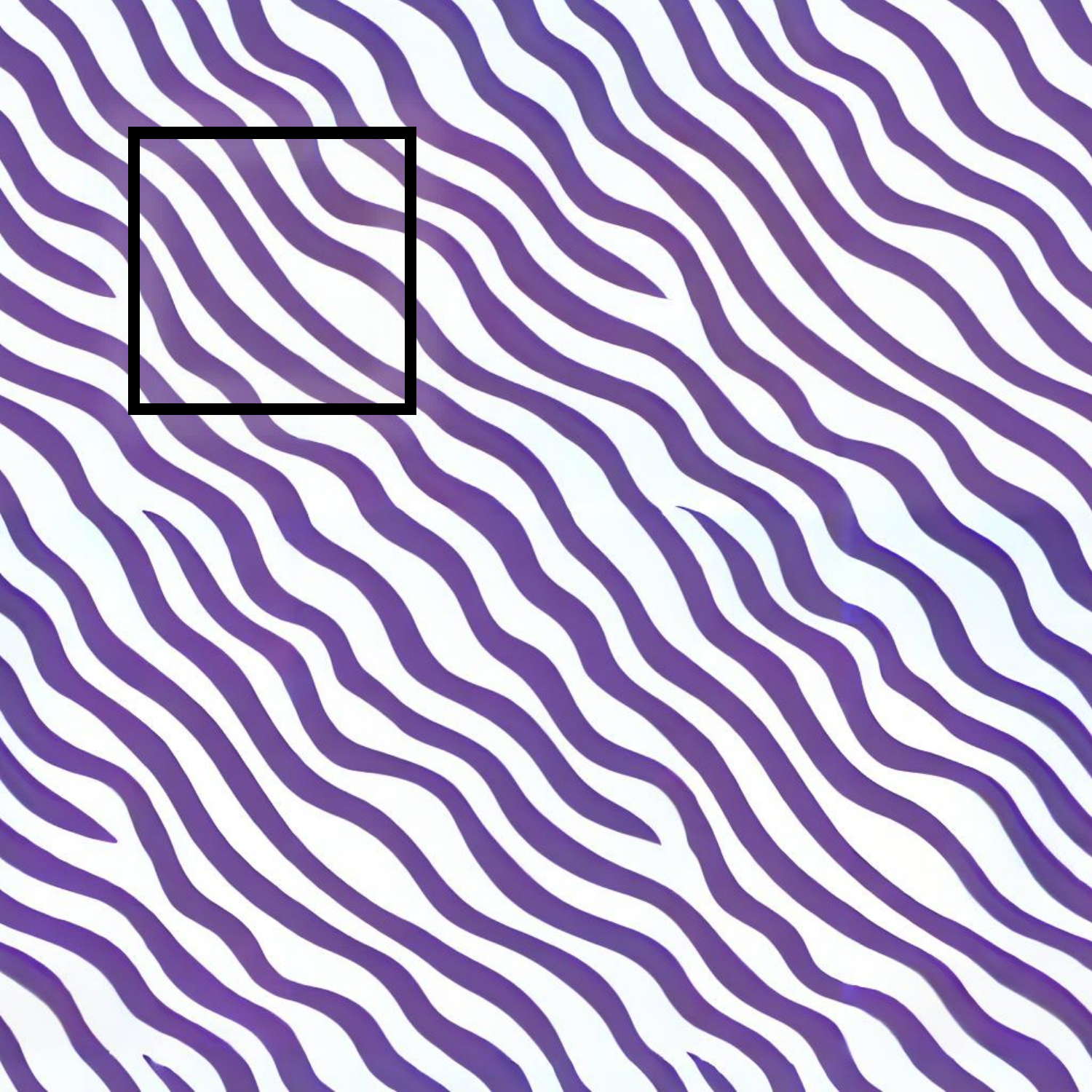}\\
    \end{tabular}
    \caption{\textbf{Results.} Our diffusion-based pattern expansion method enables the generation of large-scale, high-quality, and tileable patterns from a small user-drawn input, reported in the black boxes. By being fine-tuned on domain-specific data, it adapts to different structured arrangements of solid-colored shapes, consistently extending the input design features to a larger-scale result.}
    \label{fig:results_highres}
\end{figure*}

We quantitatively evaluate our method by computing the TexTile metric score \cite{rodriguez2024textile} for 35 expanded patterns. TexTile provides a differentiable metric that quantifies how likely a texture can be concatenated with itself without introducing artifacts. I our experiments we achieve a TexTile score of $63.93\%\pm8.24\%$, which aligns with the score obtatined on tileable textures from the dataset ($62.25\%\pm14.04\%$). We manually performed a texture concatenation in both vertical and horizontal directions to further validate the tileability of the expanded patterns. Fig.\ref{fig:textile} shows that our pattern expansion pipeline produces perfectly tileable results, due to the application of the noise rolling technique. However, we encourage larger scale expansion instead of replicating a smaller scale one to avoid detail repetition (see Fig.\ref{fig:results_highres}).

\subsection{Ablation Study}
\label{sec:ablation}

We evaluate our design choices starting from the baseline solution and gradually introducing the different proposed architectural components and diffusion elements--IP-Adapter, LoRA finetuning, and noise rolling--. To systematically assess the impact of each component, we test the different configurations on a series of example patterns. We provide qualitative results of the ablation study in Fig.~\ref{fig:ablation}.

We first evaluate the Stable Diffusion base model performing a text-guided inpainting task (\ref{fig:ablation}a). This sets a performance baseline without being influenced by any of the design choices presented in the paper. Although the model is able to fill in the missing areas, it tends to diverge from the input condition and break the overall structure. Even for simple examples, the text-guided approach is not a natural mean to express pattern structures such as shapes and arrangements, and moreover, it is not versatile enough to perfectly describe the design of the partial input pattern.  

To provide control in a more natural way, we include an IP-Adapter ~\cite{ye2023ip} that introduces an image prompt as further guidance for the inpainting process. The guidance image is constructed by simply repeating the image prompt multiple times to fill a $512\times512$ canvas. As described in Sec.~\ref{sec:inference} this helps the CLIP encoder better capture the visual features and sharpness of the guidance, due to the high sensitivity of CLIP to image resolution~\cite{wang2023exploring}. As shown in our results in ~\ref{fig:ablation}b, visual guidance allows the model to better follow the input, while still presenting some visual inconsistencies and limitations, mostly due to the training on natural images. In fact, the model is capable of better catching the style and colors provided by the guidance image, but it still fails at reconstructing its geometrical details in both shape features or pattern scale and often provides natural-looking results.

Since the model is more exposed to photorealistic, natural, and unstructured data during training, we perform a fine-tuning of the structured patterns to better adapt it to this new domain and task. To do so, we trained a LoRA module on our crafted pattern dataset. By combining the LoRA domain knowledge with the Stable Diffusion Model backbone, we noticed that the overall result quality and consistency are significantly improved, thanks to the new adaptation to the pattern domain. In particular, results preserve the same style as the provided input and reconstruct geometrical details and arrangements in a more resilient way (\ref{fig:ablation}c). 

\begin{figure*}[h]
    \centering
    \setlength{\tabcolsep}{.5pt}
    \begin{tabular}{ccc}
        \vspace{-0.5mm}\hspace{-1mm}
        \figframe[width=0.32\textwidth]{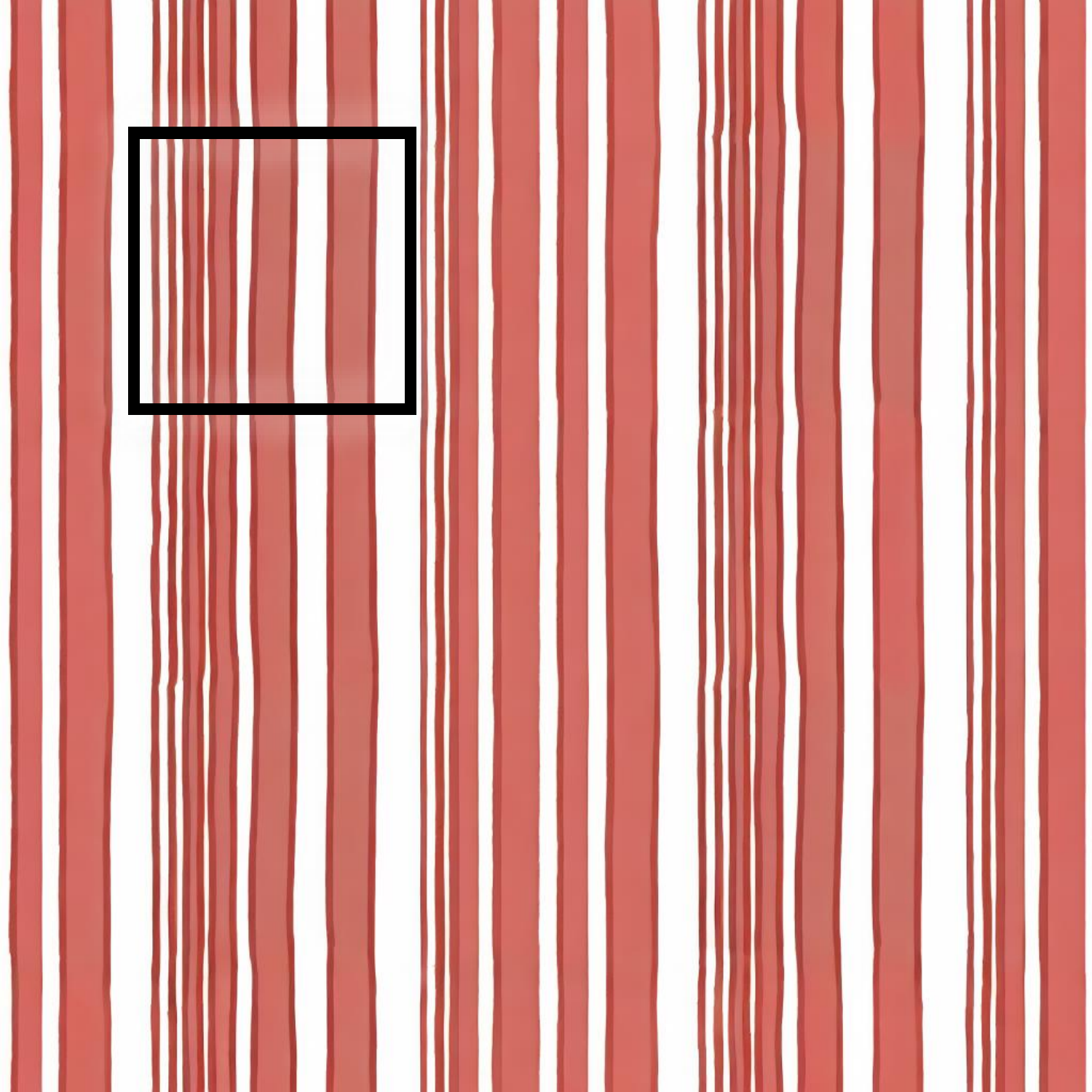} &
        \figframe[width=0.32\textwidth]{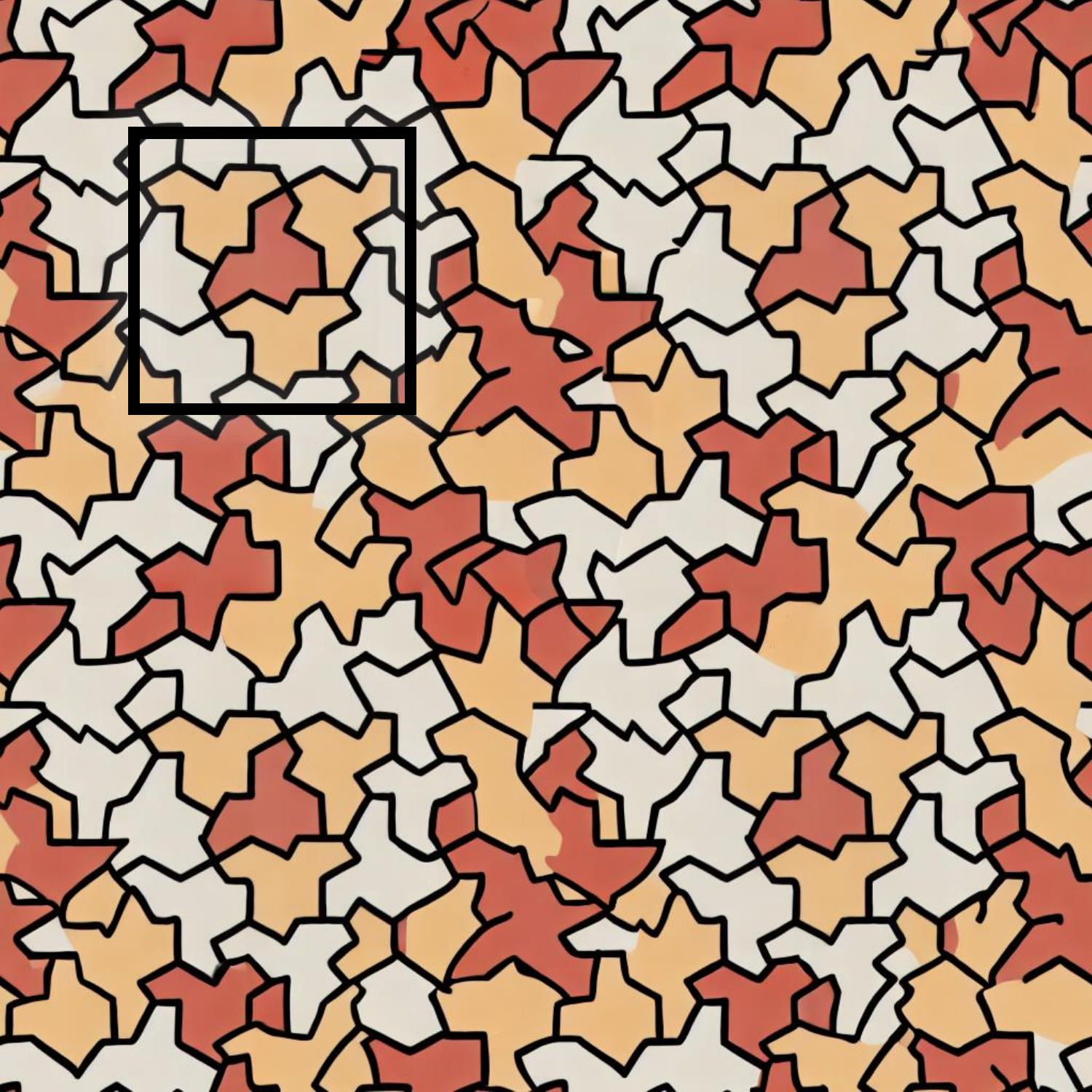} &
        \figframe[width=0.32\textwidth]{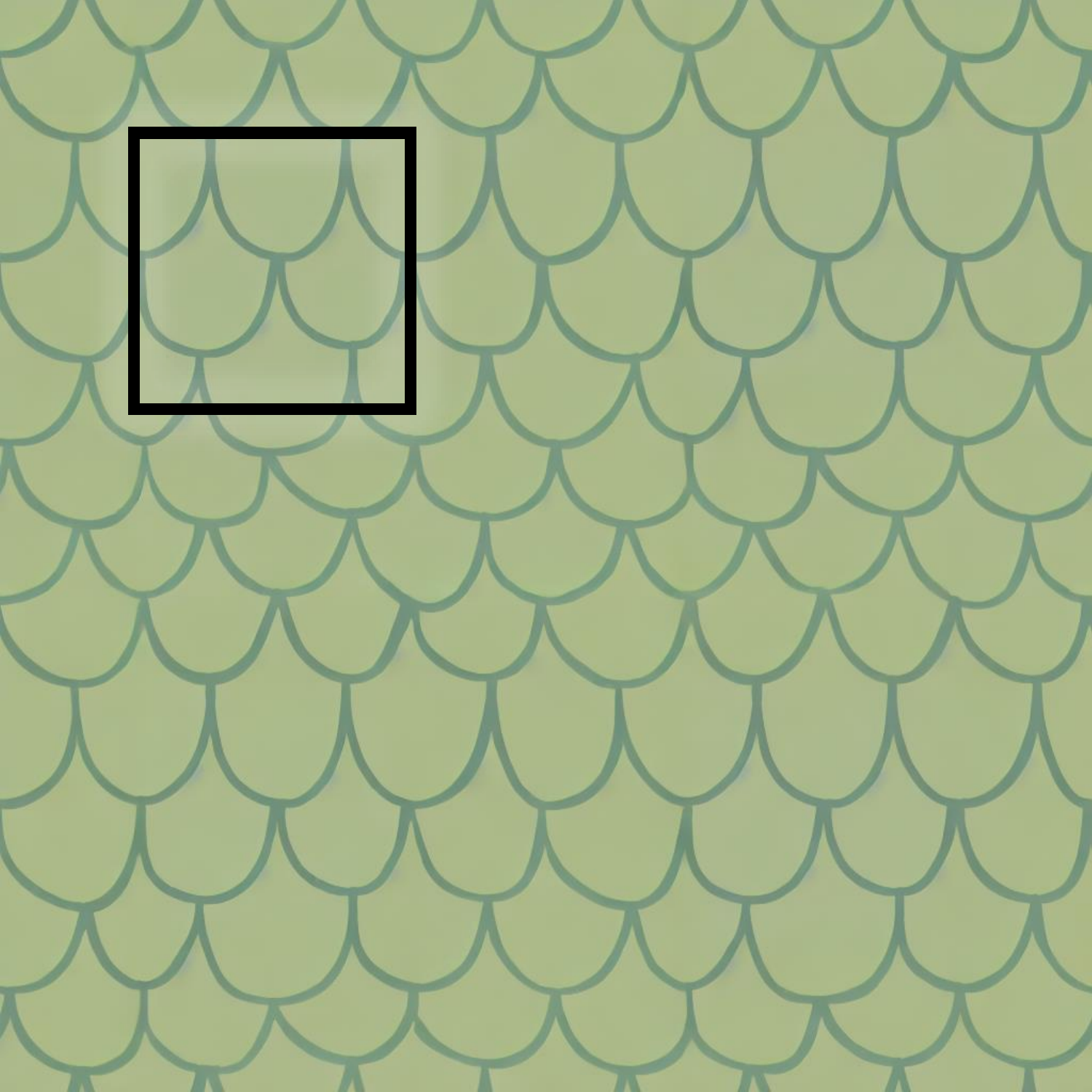} \\
        \end{tabular}
    \caption{\textbf{Limitations.} Our model presents some limitations, which can be categorized into domain-specific and architectural limitations. First, it is by design unable to expand non-repeating patterns, either non-stationary or aperiodic \cite{smith2023aperiodic} (left and center). Additionally, it can fail to generate very structured patterns at a consistent scale while ensuring tilebility, thus squeezing or distorting the pattern to make it fit the canvas (right).}
    \label{fig:limitations}
\end{figure*}

Despite good results could be achieved on small expansions, we notice a deterioration of the output for higher expansion factors. As reported in Fig.\ref{fig:ablation}d, the expansion tends to produce a degraded output that influences the style and the structure, in terms of color artifacts and discontinuities in the pattern respectively. The introduction of the noise rolling technique enables us to produce results that correctly integrate the provided image by maintaining both the visual and geometrical aspects Fig.\ref{fig:ablation}e. In particular, this addition has a twofold effect: it makes the generated pattern tileable by removing edge discontinuities, as already assessed in \cite{vecchio2023controlmat}, and it helps in better capturing long-range dependencies inside the image, thus allowing us to increase the expansion factor without losing quality. 

In Fig.~\ref{fig:ablation_text} we assess the design choice of having a general purpose text prompt as a support to the guidance image prompt. Our fixed text prompt drives the diffusion to maintain properties like high regularity in terms of both structure and colors, which are common features in the pattern domain we focus on. The absence of a text prompt (\ref{fig:ablation_text}a) tends to include color variations and shape misalignments during the diffusion steps, deteriorating the overall coherency with the input pattern sample. In contrast, using a text prompt that is tailored to the actual input sample (\ref{fig:ablation_text}c) does not significantly improve the generation quality while requiring an additional, non trivial, effort by the user. The use of the proposed general purpose text prompt represents a good tradeoff between expansion quality and pipeline generalizability (\ref{fig:ablation_text}b).

\section{Limitations and Future Work}
\label{sec:limitation}

Our method comes with some limitations that we can divide between architectural limitations and domain limitations. Examples of failure cases or unexpected behavior are presented in Fig.~\ref{fig:limitations}. As discussed in the paper, our method cannot faithfully expand non-repeating patterns, either non-stationary (Fig.~\ref{fig:limitations} left) or aperiodic.
\cite{smith2023aperiodic} (Fig.~\ref{fig:limitations} center). This limitation comes from the design choices of our approach, which focus on repeating patterns.
While both expansions present plausible patterns, they don't necessarily follow the expected behavior, where the lines in the first figure should keep growing while the tiles should not present a predictable pattern.
Future work could focus on tackling non-repeating patterns by injecting, into the generation, additional information in the form of conditioning about the patterns' repetitiveness.
The last failure case (Fig.~\ref{fig:limitations} right), on the other hand, shows a design limitation of our approach, which can fail to generate very structured patterns at a consistent scale in the presence of tilebility. This is related to the noise rolling, which enforces tileability on the border of the image, thus squeezing the border shingles to make them fit the canvas. Possible improvements could involve an automated solution to find the optimal crop of the pattern~\cite{rodriguez2024textile} before beginning the expansion.

\section{Conclusion}
\label{sec:conclusion}

In this paper, we presented a diffusion-based architecture for structured pattern expansion, with a focus on the controllability of the generated pattern. We demonstrated the expansion of several hand-drawn patterns samples with distinctly different structures, symmetries, and appearance. Our results show the robustness of the proposed architecture and its controllability, while the comparison with prior work shows that our method is significant with respect to the state of the art.

\bibliographystyle{ACM-Reference-Format}
\bibliography{bibliography} 

\end{document}